\documentclass[sts]{imsart}

\RequirePackage[OT1]{fontenc}
\usepackage{amsthm,amsmath,amsfonts,natbib}
\RequirePackage[colorlinks,citecolor=blue,urlcolor=blue]{hyperref}

\usepackage{dsfont, bbm}

\usepackage{graphicx} 
\usepackage{subfigure}

\startlocaldefs
\endlocaldefs

\begin{document}

\begin{frontmatter}
\title{Mapping Energy Landscapes of Non-Convex Learning Problems}
\runtitle{Mapping Energy Landscapes}

\begin{aug}
\author{\fnms{Maria} \snm{Pavlovskaia}\ead[label=e1]{mariapavl@ucla.edu}},
\author{\fnms{Kewei} \snm{Tu}\ead[label=e2]{tukw@shanghaitech.edu.cn}}
\and
\author{\fnms{Song-Chun} \snm{Zhu}\ead[label=e3]{sczhu@stat.ucla.edu}}

\thankstext{t1}{The authors  thank Dr. Qing Zhou for his tutorial on the algorithm and many helpful suggestions, thank  Drs. Yingnian Wu and Adrian Barbu for their discussions, and acknowledge the support of a DARPA MSEE project FA 8650-11-1-7149.}
\runauthor{Pavlovskaia, Tu and Zhu}

\affiliation{University of California, Los Angeles and ShanghaiTech University}

\address{Maria Pavlovskaia is Ph.D. Student, Department of Statistics, University of California, Los Angeles, 8125 Math Science Bldg, Los Angeles, CA 90095, USA \printead{e1}.}

\address{Kewei Tu is Assistant Professor, School of Information Science and Technology, ShanghaiTech University, No. 8 Building, 319 Yueyang Road, Shanghai 200031, China \printead{e2}.}

\address{Song-Chun Zhu is Professor, Department of Statistics, University of California, Los Angeles, 8125 Math Science Bldg, Los Angeles, CA 90095, USA \printead{e3}.}

\end{aug}

\begin{abstract}
In many statistical learning problems, the target functions to be optimized are highly non-convex in various model spaces and thus are difficult to analyze. In this paper, we compute \emph{Energy Landscape Maps} (ELMs) which characterize and visualize an energy function with a tree structure, in which each leaf node represents a local minimum and each non-leaf node represents the barrier between adjacent energy basins. The ELM also associates each node with the estimated probability mass and volume for the corresponding energy basin. We construct ELMs by adopting the generalized Wang-Landau algorithm and multi-domain sampler that simulates a Markov chain traversing the model space by dynamically reweighting the energy function. We construct ELMs in the model space for two classic statistical learning problems: i) clustering with Gaussian mixture models or Bernoulli templates; and ii) bi-clustering. We propose a way to measure the difficulties (or complexity) of these learning problems and study how various conditions affect the landscape complexity, such as separability of the clusters, the number of examples, and the level of supervision; and we also visualize the behaviors of different algorithms, such as K-mean, EM, two-step EM and Swendsen-Wang cuts, in the energy landscapes. 
\end{abstract}


\end{frontmatter}

\section{Introduction}

In many statistical learning problems, the energy functions to be optimized are highly non-convex. A large body of research has been devoted to either approximating the target function by convex optimization, such as replacing $L_0$ norm by $L_1$ norm in regression,  or designing algorithms to find a good local optimum, such as EM algorithm for clustering.  Much less work has been done in analyzing the properties of such non-convex energy landscapes. 

In this paper, inspired by the success of visualizing the landscapes of Ising and Spin-glass models by \cite{becker} and \cite{zhou}, we compute \emph{Energy Landscape Maps} (ELMs) in the high-dimensional model spaces (i.e. the hypothesis spaces in the machine learning literature) for some classic statistical learning problems --- clustering and bi-clustering. 

\begin{figure}
    \center
    \includegraphics[scale=.40]{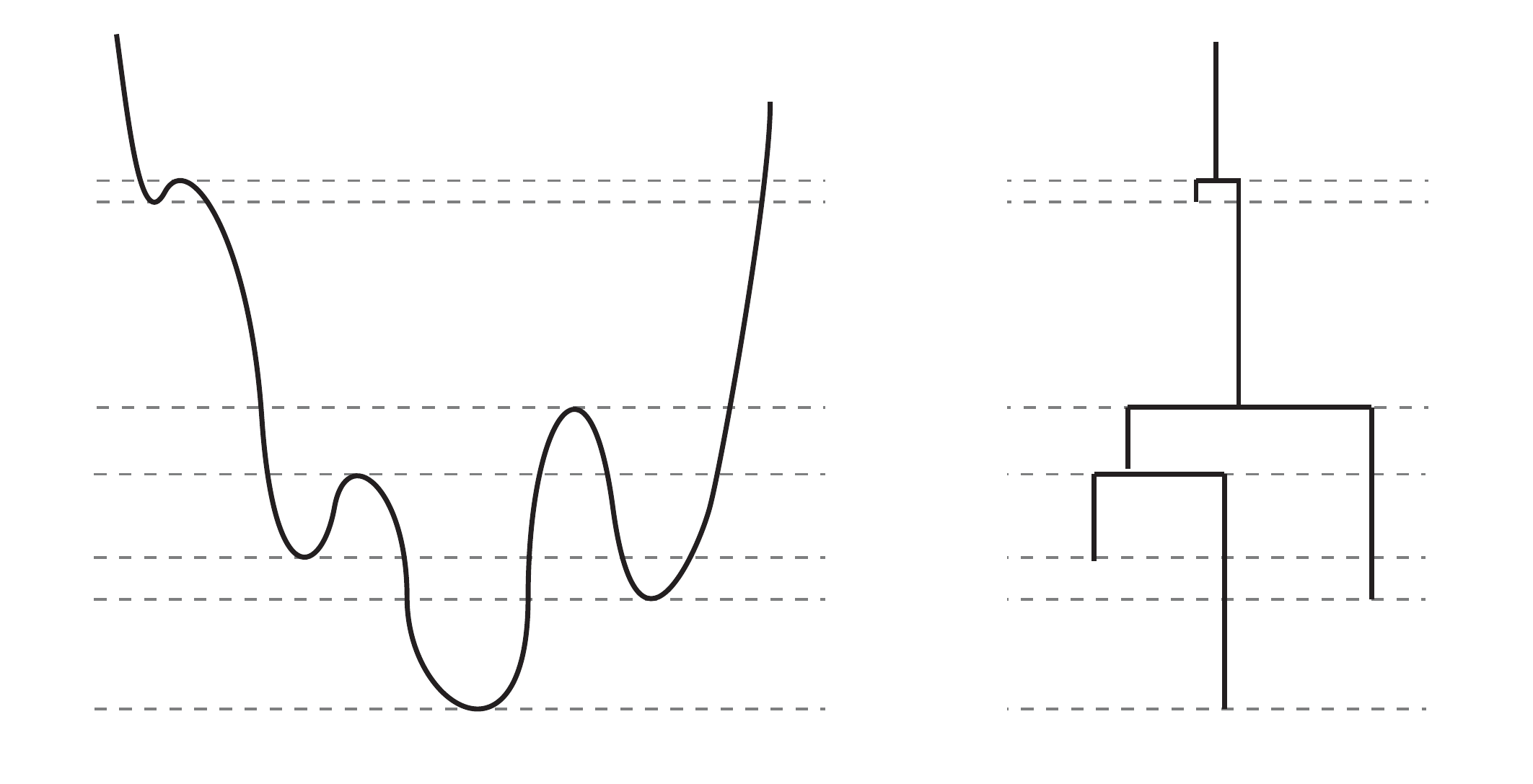}
    \caption{An energy function and the corresponding Energy Landscape Map (ELM). The y-axis of the ELM is the energy level, each leaf node is a local minimum and the leaf nodes are connected at the ridges of their energy basins.  
    } \label{fig:mapping-to-elm} 
\end{figure}

 The ELM is a tree structure, as  Figure~\ref{fig:mapping-to-elm} illustrates,  in which each leaf node represents a local minimum and each non-leaf node represents the barrier between adjacent energy basins. The ELM characterizes the energy landscape with the following information.
\begin{itemize}
\item The number of local minima and their energy levels;
\item The energy barriers between adjacent local minima and their energy levels; and
\item The probability mass and volume of each local minimum (See Figure~\ref{fig:2d-mass-vol}).
\end{itemize} 
Such information is useful in the following tasks.
\begin{enumerate}
\item  Analyzing the intrinsic difficulty (or complexity) of the optimization problems, for either inference or learning tasks. 
For example, in bi-clustering, we divide the problem into the {\em easy}, {\em hard}, and {\em impossible} regimes under different conditions. 
 
\item Anlyzing the effects of various conditions on the ELM complexity, for example, the separability in clustering, the number of training examples, the level of supervision (i.e. how many percent the examples are labeled), and the strength of regularization (i.e. prior model).

\item Analyzing the behavior of various algorithms by showing their frequencies of visiting the various minima. For example, in the muilti-Gaussian clustering problem, we find that when the Gaussian components are highly separable, K-means clustering works better than the EM algorithm \cite{rubin}, and the opposite is true when the components are less separable. In contrast to the frequent visits of local minimum by K-means and EM, the Swendsen-Wang cut method \cite{zhu-sw-cut} converges to the global minimum in all separability conditions.
\end{enumerate}

\begin{figure}[htbp!]
    \center
    \subfigure[] {
        \includegraphics[width=0.6\columnwidth]{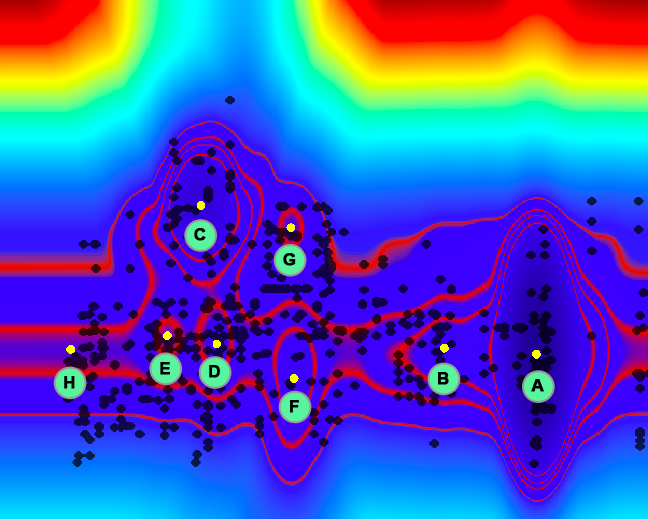}
    }
    \subfigure[]{
        \includegraphics[width=0.2\columnwidth]{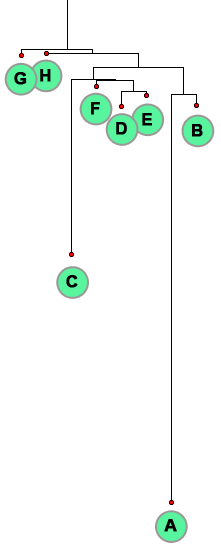}
    }
    \caption{ (a) Energy Landscape for a 4-component 1-d GMM with all parameters fixed except two means. Level sets are highlighted in red. The local minima are shown in yellow dots and the first 200 MCMC samples are shown in black dots. (b) The resulting ELM and the correspondence between the leaves and the local minima from the energy landscape.} \label{fig:2vars}
\end{figure}


\begin{figure} [htbp!]
    \center
    \subfigure[probability mass] {
        \includegraphics[width=0.25 \columnwidth]{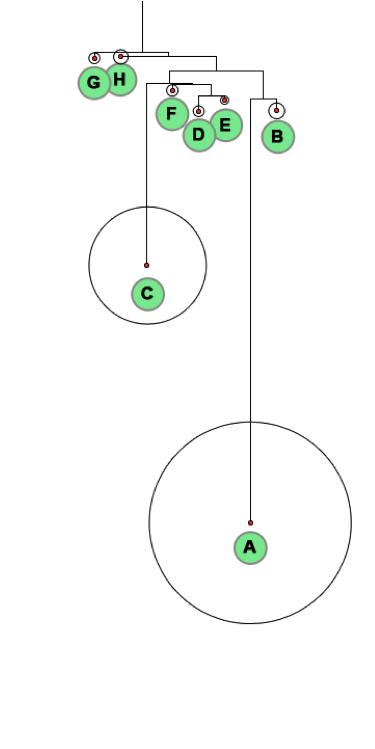}
    } 
    \hspace{2cm}
    \subfigure[volume]{
        \includegraphics[width=0.25 \columnwidth]{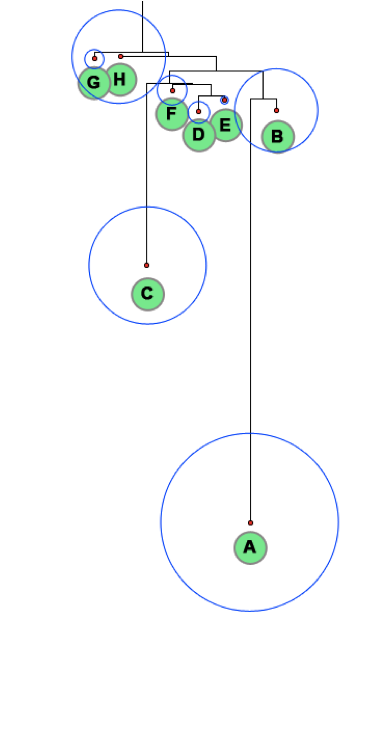}
    }
 
    \caption{The probability mass and volume of the energy basins for the 2-d landscape shown in Figure \ref{fig:2vars}. } \label{fig:2d-mass-vol} 
\end{figure}

We start with a simple illustrative example in Figures~\ref{fig:2vars} and \ref{fig:2d-mass-vol}.  Suppose the underlying probability distribution is a 4-component Gaussian mixture model (GMM) in 1D space, and the components are well separated. The model space is 11-dimensional 
with parameters $\{ (\mu_i, \sigma_i, \alpha_i): i=1,2,3,4\}$ denoting the means, variance and weights for each components.  We sampled $70$ data points from the GMM and  construct the ELM in the model space. We bound the model space to a finite range defined by the samples.
 
As we can only visualize 2D maps, we set all parameters to equal the truth value except keeping $\mu_1$ and $\mu_2$ as the unknowns.
Figure \ref{fig:2vars}(a) shows the energy map  on a range of $0 \leq \mu_1, \mu_2 \leq 5$.  The asymmetry in the landscape is caused by the fact that the true model has different weights between the first and second component. Some shallow local minima, like E, F, G,H, are little ``dents'' caused by the finite data samples. 

Figure \ref{fig:2vars} (a) shows that all the local minima are identified. Additionally, it shows the first 200 MCMC samples that were accepted by the algorithm that we will discuss late. The samples are clustered around the local minima, and cover all energy basins. They are not present in the high energy areas away from the local minima, as would be desired.  Figure \ref{fig:2vars} (b) shows the resulting ELM and the correspondence between the leaves and the local minima in the energy landscape. Furthermore, Figures \ref{fig:2d-mass-vol} (a) and (b) show the probability mass and the volume of these energy basins.

In the literature, \cite{becker} presents the first work for visualizing multidimensional energy landscapes for the spin-glass model. Since then statisticians have developed a series of MCMC methods for improving the efficiency of the sampling algorithms traversing the state spaces. Most notably, \cite{WL_Liang,GWL} generalize the Wang-Landau algorithm~\cite{WL} for random walks in the state space.  \cite{zhou}
uses the generalized Wang-Landau algorithm to plot the disconnectivity graph for Ising model with 100s of local minimum and proposes an effective way for estimating the energy barrier. Furthermore, \cite{Zhou2} construct the energy landscape for Bayesian inference of DNA sequence segmentation by clustering Monte Carlo samples.

In contrast to the above work that compute the landscapes in ``state'' spaces for inference problems, our work is focused on the landscapes in ``model'' spaces (the sets of all models; also called hypothesis spaces in the machine learning community) for statistical learning and model estimation problems. There are some new issues in plotting the model space landscapes. i) Many of the basins have a flat bottom, for example, basin A in Figure~\ref{fig:2vars}.(a). This may result in a large number of false local minima. ii) The are constraints among some parameters, for example the weights have to sum to one --- $\sum_i \alpha_i =1$. Thus we may need to run our algorithm on a manifold.


\section{ELM construction}

In this section, we introduce the basic ideas for constructing the ELM and estimating its properties \-- mass, volume and complexity.


\subsection{Space partition}

Let $\Omega$ be the  model space over which a probability distribution $\pi(x)$ and energy $E(x)$ are defined.  In this paper, we assume $\Omega$ is bounded using properties of the samples.   $\Omega$ is  partitioned into $K$ disjoint subspaces which represent the energy basins
\begin{equation}
\Omega = \cup^{K}_{i=1} D_i, \quad
  \cap_{i=1}^K D_i =  \emptyset ~~\forall i\neq j.
\end{equation}
That is, any point $x \in D_i$ will converge to the same minimum through gradient descent.


\begin{figure}
\center
 \includegraphics[width= 0.4 \columnwidth] {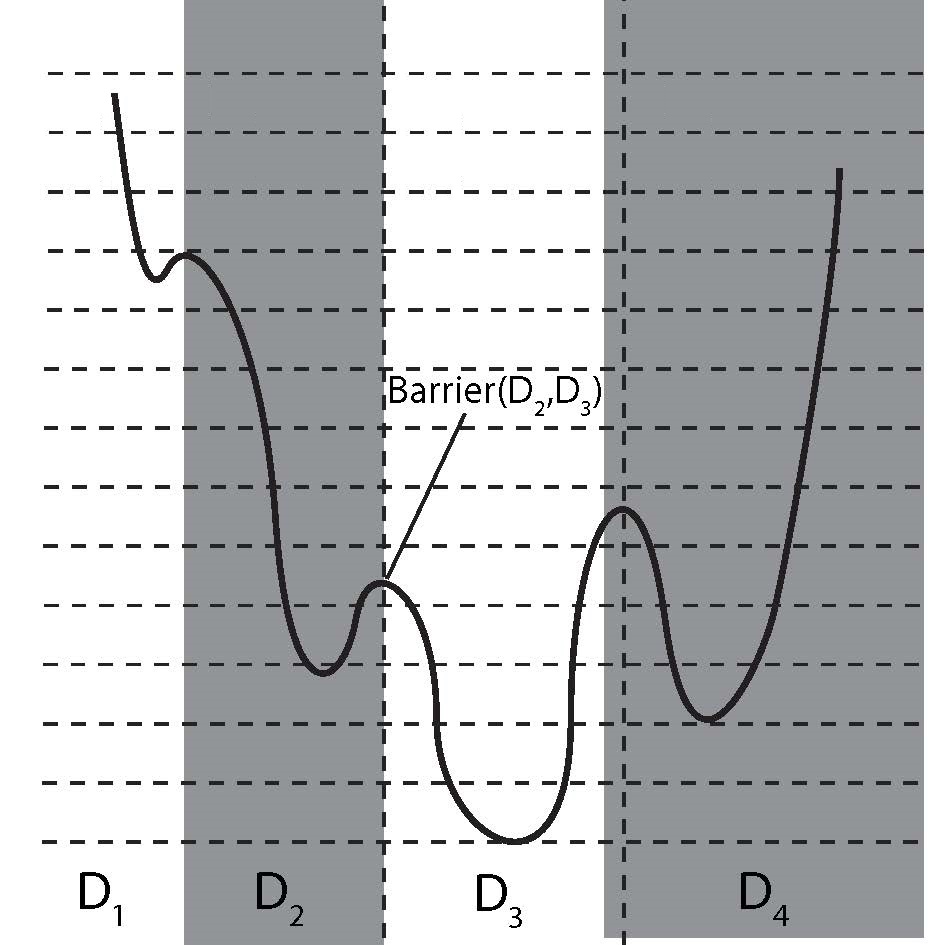}
    \caption{ The model space $\Omega$ is partitioned into energy basins $D_i$ (along the x-axis), and the energy $\mathbb{R}$ (the y-axis) is partitioned into uniform intervals $ [u_{j+1}, u_j)$.    } \label{fig:binning} 
\end{figure}

As Figure~\ref{fig:binning} shows, the energy is also partitioned into intervals $ [u_{j+1}, u_j), j=1,2,...,L$.   Thus we obtain a set of bins as the quantized atomic elements in the product space  $\Omega \times \mathbb{R}$, 
 \begin{equation}
 B_{ij} = \{ x: x \in D_i, ~ E(x) \in [u_{j+1}, u_j)\}.
 \end{equation}
The number of basins $K$ and the number of intervals $L$ are unknown and have to be estimated during the computing process in an adaptive and iterative manner.
 
\subsection{Generalized Wang-Landau algorithm}

The objective of the generalized Wang-Landau (GWL) algorithm is to simulate a Markov chain that visits all the bins $\{ B_{ij}, \forall i,j\}$ with equal probability, and thus effectively reveal the structure of the landscape. 

Let $\phi: \Omega \rightarrow \{1,\dots,K\} \times \{1, ...,L\}$ be the mapping between the model space and bin indices: $\phi(x) =(i,j)$ if $x\in B_{ij}$.
Given any $x$, by gradient descent or its variants, we can find and record the basin $D_i$ that it belongs to, compute its energy level $E(x)$, and thus find the index $\phi(x)$.

We define $\beta(i,j)$ to be the probability mass of a bin 
\begin{equation}
 \beta(i,j) = \int_{B_{i,j}} \pi(x) ~ dx.
\end{equation}
Then, we can define a new probability distribution which has equal probability among all the bins,
\begin{equation}
\pi'(x) = \frac{1}{Z} \pi(x) / \beta(\phi(x)),
\end{equation}
with $Z$ being a scaling constant. 

To sample from $\pi'(x)$, one can estimate $\beta(i,j)$ by a variable $\gamma_{ij}$. We define the probability function $\pi_\gamma: \Omega \rightarrow \mathbb{R}$ to be 
\begin{align*}
\pi_\gamma(x) \propto \frac{\pi(x) }{\gamma_{\phi(x)}} = \sum_{i,j} \frac{\pi(x)}{\gamma_{ij}} \mathbbm{1}(x \in B_{ij}) ~\text{ st. } \int_{\Omega} \pi_\gamma(x)  dx = 1.
\end{align*}
We start with an initial $\gamma^0$, and update $\gamma^t =\{\gamma_{ij}^t, \forall i,j\}$ iteratively using stochastic approximation \cite{Stoch_grad}. Suppose $x_t$ is the MCMC state at time $t$, then $\gamma^t$ is updated in an exponential rate,
\begin{equation}
   \label{eq:update}
   \log \gamma_{ij}^{t+1} = \log \gamma_{ij}^{t} + \eta_t \mathbbm{1}(x_t \in B_{ij}), \quad \forall i,j.
\end{equation}
$\eta_t$ is the step size at time $t$. The step size is decreased over time and the decreasing schedule is either pre-determined as in \cite{Stoch_grad} or determined adaptively as in \cite{zhou2011multi}.

Each iteration with given $\gamma^t$ uses a Metropolis step. Let $Q(x,y)$ be the proposal probability for moving from $x$ to $y$, then the acceptance probability is 

\begin{eqnarray}\label{eq:acceptance}
\alpha(x,y) &= \min\left(1, \frac{Q(y,x) \pi_{\gamma}(y)}{Q(x,y) \pi_{\gamma}(x)}\right)\\ \nonumber
& = \min\left( 1, \frac{Q(y,x)}{Q(x,y)} \frac{\pi(y)} {\pi(x)} \frac{\gamma^t_{\phi(x)}}{\gamma^t_{\phi(y)}}\right).
\end{eqnarray} 
Intuitively, if $\gamma^t_{\phi(x)} < \gamma^t_{\phi(y)}$, then the probability of visiting $y$ is reduced. For the purpose of exploring the energy landscape,  the GWL algorithm improves upon conventional methods, such as the simulated annealing \cite{geyer} and tempering \cite{marinari} process. The latter sample from $\pi(x)^{\frac{1}{T}}$ and do not  visit the bins with equal probability even at high temperature.

\begin{figure}
    \center
    {
        \includegraphics[width = 0.4 \columnwidth ]{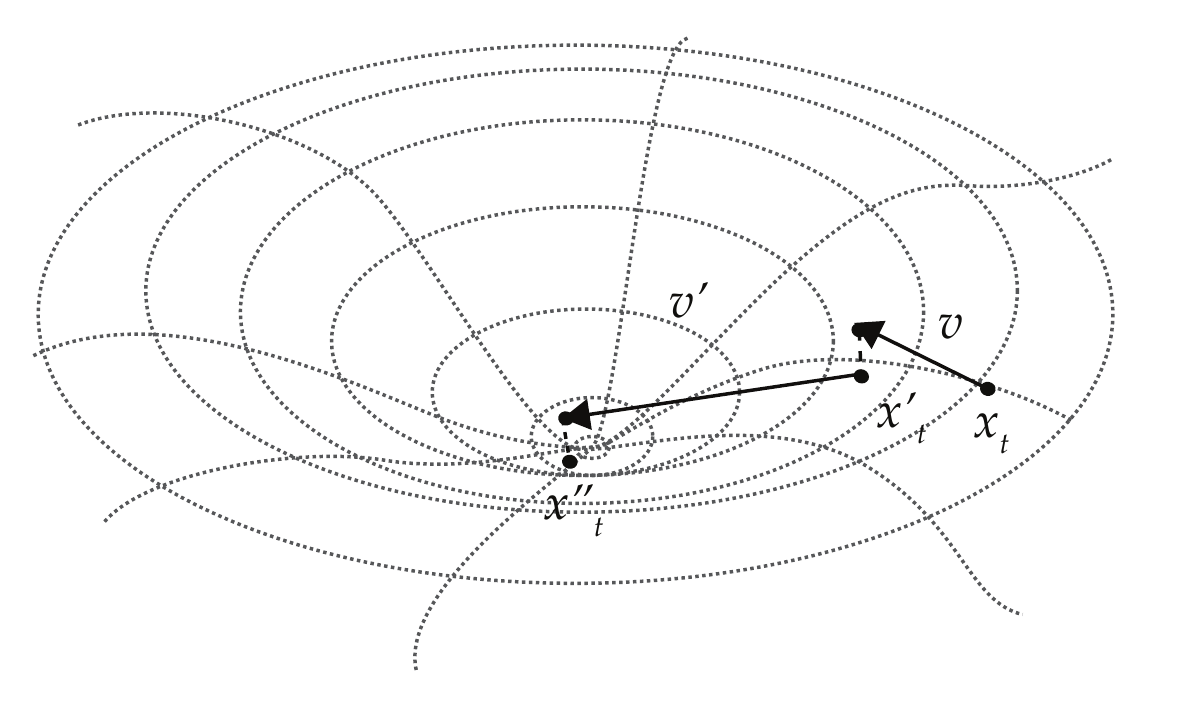}
    }
    \caption{First two steps of projected gradient descent. The algorithm is initialized with MCMC sample $x_t$. $v$ is the gradient of $E(x)$ at the point $x_t$. Armijo line search is used to determine the step size $\alpha$ along the vector $v$.  $x'_t$ is the projection $T(x_t+\alpha v)$ onto the subspace $\Gamma$. Then $x''_t$ is the projection $T(x_t+\alpha' v')$, and so on.}
    \label{fig:projection}
\end{figure}

In performing gradient descent, we employ Armijo line search to determine the step size; if the model space $\Omega$ is a manifold in $\mathbb{R}^n$, we perform projected gradient descent, as shown in Figure \ref{fig:projection}.
To avoid erroneously identifying multiple local minima within the same basin (especially when there is large flat regions), we merge local minima identified by gradient descent based on the following criteria: (1) the distance between two local minima is smaller than a constant $\epsilon$; or (2) there is no barrier along the straight line between two local minima.

\begin{figure}[t]
\center
\includegraphics[scale=.45]{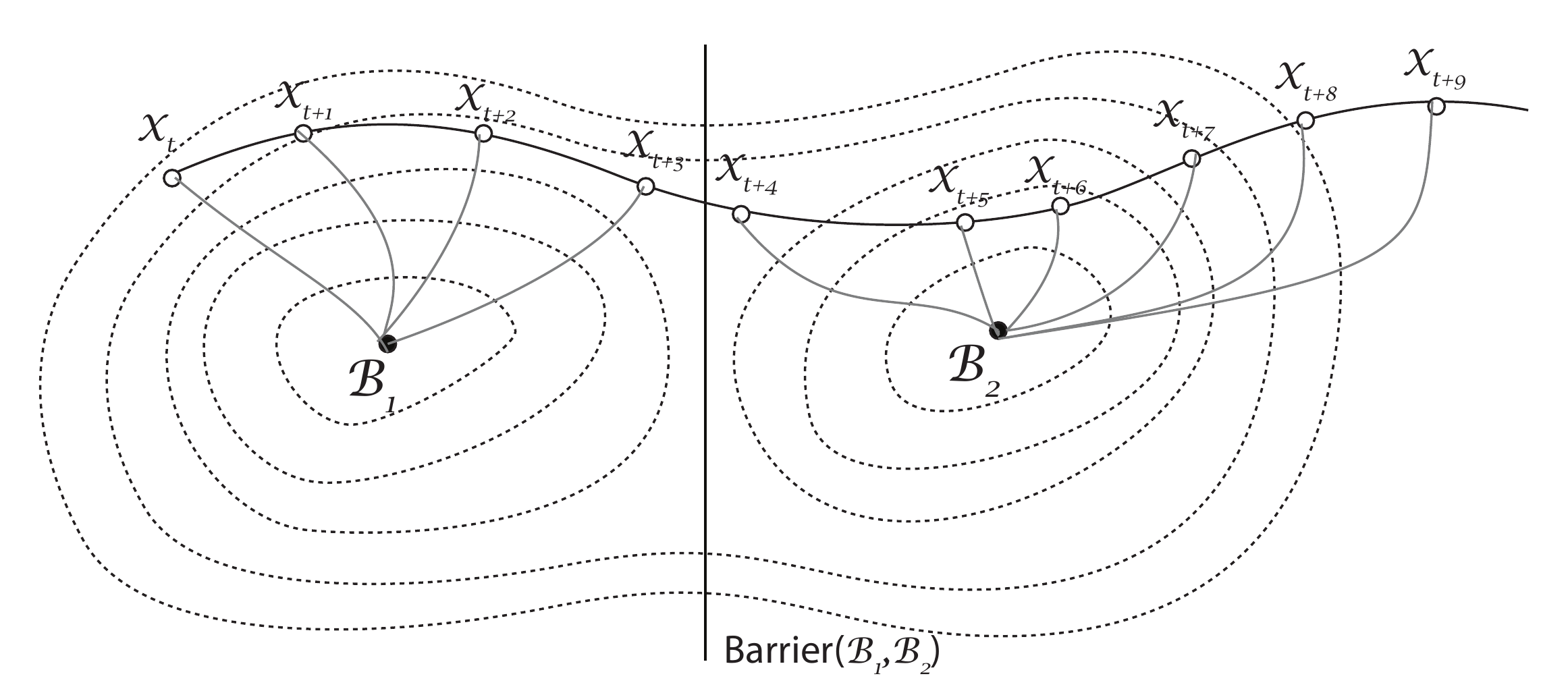}
   \caption{Sequential MCMC samples $x_t, x_{t+1}, \dots, x_{t+9}$. For each sample, we perform gradient descent to determine which energy basin the sample belongs to. If two sequential samples fall into different basins ($x_{t+3}$ and $x_{t+4}$ in this example), we estimate or update the upper-bound of the energy barrier between their respective basins ($B_1$ and $B_2$ in this example).} \label{fig:paths}
\end{figure}

Figure \ref{fig:paths} (a) illustrates a sequence of  Markov chain states $x_t, ..., x_{t+9}$ over two energy basins. The dotted curves are the level sets of the energy function.

\subsection{Constructing the ELM}

Suppose we have collected a chain of samples $x_1, \dots, x_N$ from the GWL algorithm.
The ELM construction consists of the following two processes.


\begin{figure}
\center
\includegraphics[width = 0.5\columnwidth]{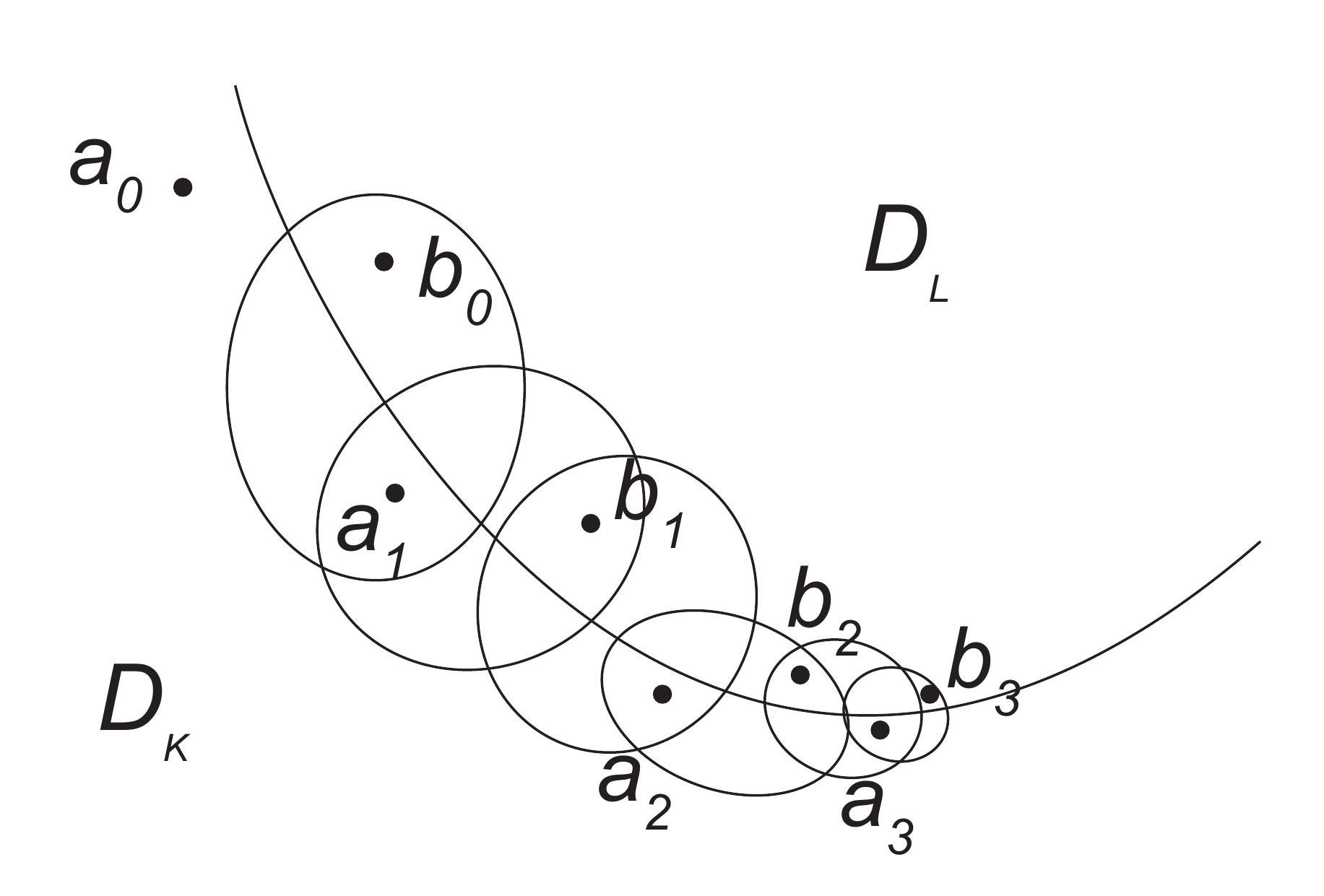}
   \caption{The ridge descent algorithm is used for estimating the energy barrier between basins $D_k$ and $D_l$ initialized at consecutive MCMC samples $a_0 = x_t, b_0 = x_{t+1}$ where $a_0 \in D_k$ and $b_0 \in D_l$. } \label{fig:ridge}
\end{figure}

{\em 1, Finding the energy barriers between adjacent basins}.   We collect all consecutive MCMC states that move across two basins $D_k$ and $D_l$,
\begin{equation}
X_{kl} = \{ (x_t, x_{t+1}): x_t \in D_k, x_{t+1} \in D_l\}
\end{equation} 
 we choose $(a_0,b_0) \in X_{kl}$ with the lowest energy 
 \[(a_0,b_0) = \text{argmin}_{(a,b) \in \Omega_{kl}}\left[\min(E(a),E(b))\right]. \]
Next we iterate the following step as Figure~\ref{fig:ridge} illustrates 
\begin{align*}
a_i &= \text{argmin}_a \left\{ E(a): a\in\text{Neighborhood}(b_{i-1})  \cap D_k \right\} \\
b_i &= \text{argmin}_b \left\{ E(b): b\in\text{Neighborhood}(a_{i})  \cap D_l \right\}
\end{align*}
until $b_{i-1} = b_i$.  The neighborhood is defined by an adaptive radius. 
Then $b_i$ is the energy barrier and $E(b_i)$ is the energy level of the barrier. A discrete version of this ridge descent method was used in \cite{zhou}. 

{\em 2, Constructing the tree structure}. The tree structure of the ELM is constructed from the set of energy basins and the energy barriers between them via an iterative algorithm modified from the hierarchical agglomerative clustering algorithm. Initially, the energy basins are represented by leaf nodes that are not connected, whose y-coordinates are determined by the local minima of the basins. In each iteration, the two nodes representing the energy basins $D_1, D_2$ with the lowest barrier are connected by a new parent node, whose y-coordinates is the energy level of the barrier; $D_1$ and $D_2$ are then regarded as merged, and the energy barrier between the merged basin and any other basin $D_i$ is simply the lower one of the energy barriers between $D_1/D_2$ and $D_i$. When all the energy basins are merged, we obtain the complete tree structure. For clarity, we can remove from the tree basins of depth less than a constant $\epsilon$. 

%
%
%


\subsection{Estimating the mass and volume of nodes in the ELM}\label{sec:alg}
In the ELM, we can estimate the probability mass and the volume of each energy basin. When the algorithm converges, the normalized value of $\gamma_{ij}$ approaches the probability mass of bin $B_{ij}$:
\[
	\hat{P}(B_{ij}) = \frac{\gamma_{ij}}{\sum_{kl} \gamma_{kl}} \rightarrow \beta(i,j), \quad {\rm almost \; surely}.
\]
Therefore the probability mass of a basin $D_i$ can be estimated by
\begin{equation}
  \hat{P}(D_i) = \sum_j \hat{P}(B_{ij}) = \frac{\sum_j \gamma_{ij}}{\sum_{kl} \gamma_{kl}}
\end{equation}


Suppose the energy $E(x)$ is partitioned into sufficiently small intervals of size $du$.
Based on the probability mass, we can then estimate the size\footnote{Note that the size of a bin/basin in the model space is called its volume by \cite{Zhou2}, but here we will use the term ``volume'' to denote the capacity of a basin in the energy landscape.} of the bins and basins in the model space $\Omega$. 
A bin $B_{ij}$ with energy interval $[u_j, u_j+du)$ can be seen as having energy $u_j$ and probability density $\alpha e^{-u_j}$ ($\alpha$ is a normalization factor). The size of bin $B_{ij}$ can be estimated by
\[
  \hat{A}(B_{ij}) = \frac{\hat{P}(B_{ij})}{\alpha e^{-u_j}} = \frac{\gamma_{ij}}{\alpha e^{-u_j} \sum_{kl} \gamma_{kl}}
\]
The size of basin $D_i$ can be estimated by
\begin{equation}
  \hat{A}(D_i) = \sum_j \hat{A}(B_{ij}) = \frac{1}{\sum_{kl} \gamma_{kl}} \sum_j \frac{\gamma_{ij}}{\alpha e^{-u_j}}
\end{equation}

\begin{figure}[t]
\center
 \includegraphics[width= 0.8 \columnwidth] {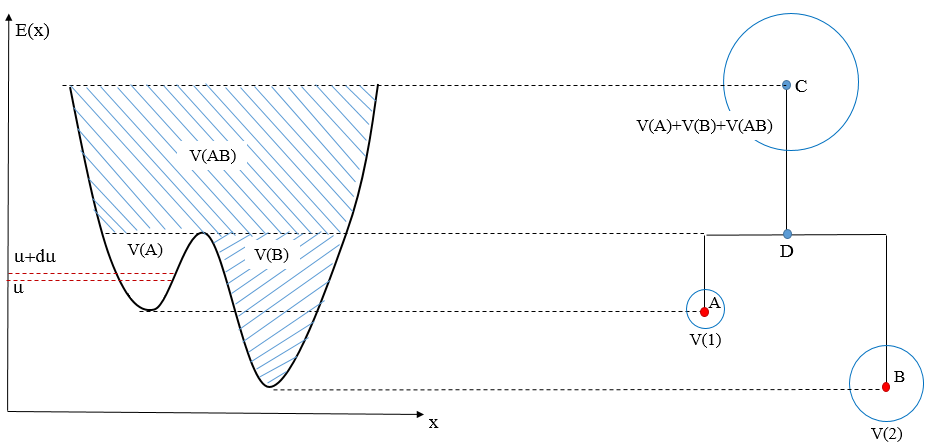}
    \caption{The volume of basins. Assuming that $du$ is sufficiently small, the volume of an energy basin can be approximated by the summation of the estimated volume at each energy interval. } \label{fig:mass-volume-calc} 
\end{figure}

Further, we can estimate the volume of a basin in the energy landscape which is defined as the amount of space contained in a basin in the space of $\Omega\times \mathbb{R}$.
\begin{equation}
   \hat{V}(D_i) = \sum_{j} \sum_{k: u_k \leq u_j} \hat{A}(B_{ik}) \times du = \frac{du}{\sum_{lm} \gamma_{lm}} \sum_{j} \sum_{k: u_k \leq u_j} \frac{\gamma_{ik}}{\alpha e^{-u_k}}
\end{equation} 
where the range of $j$ depends on the definition of the basin. In a restricted definition, the basin only includes the volume under the closest barrier, as Figure~\ref{fig:mass-volume-calc} illustrates. The volume above the basins $1$ and $2$ is shared by the two basins, and is between the two energy barriers $C$ and $D$. Thus we define the volume for a non-leaf node in the ELM to be the sum of its childen plus the volume between the barriers. For example,  node $C$ has volume $V(A) + V(B) + V(AB)$.

If our goal is to develop a scale-space representation of the ELM by repeatedly smoothing the landscape, then basins $A$ and $B$ will be merges into one basin at certain scale, and volume above the two basins will be also added to this new merged basin.

Note that the partition of the space into bins, rather than basins, facilitates the computation of energy barriers, the mass and volume of the basins.

\begin{figure}[t]
\center
 \includegraphics[width= 1.0 \columnwidth] {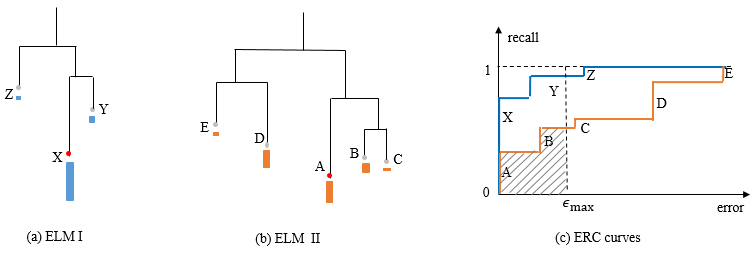}
    \caption{Characterizing the difficulty of learning in the ELM. For two learning tasks with ELM I and ELM II, the colored bar show the frequency that a learning algorithm converges to the basins, from which two Error-recall curves are plotted. The difficulty of learning task, with respect to this algorithm, can be measured by the area under the curve within an acceptable maximum error.} \label{fig:difficulty_measure} 
\end{figure}

\subsection{Characterizing the difficulty (or complexity) of learning tasks \label{sect:difficulty}}

It is often desirable to measure the difficulty of the learning task by a single number. For example, we compare two ELMs in Figure~\ref{fig:difficulty_measure}. Learning in the landscape of ELM I looks easier than that of ELM II. However, the difficulty also depends on the learning algorithms.   Thus we can run the learning algorithm   many times and record the frequency that it converges to each basin or minimum.
The frequency is shown by the lengths of the colored bars under the leaf nodes.
 
Suppose that $\Theta^\ast$ is the true model to be learned. In Figure~\ref{fig:difficulty_measure}, $\Theta^\ast$ corresponds to nodes $X$ in ELM I and node $A$ in ELM II. In general, $\Theta^\ast$ may not be the global minimum or not even a minimum. 
We then measure the distance (or error) between $\Theta^\ast$ and any other local minima. As the  error  increases, we accumulate the frequency to plot a curve.
We call it the Error-Recall curve (ERC), as the horizontal axis is the error and the vertical axis is the frequency of recall the solutions. This is like the 
ROC (receptor-operator characteristics) curves in Bayesian decision theory, pattern recognition and machine learning. By sliding the threshold 
$\epsilon_{\rm max}$ which is maximum  error tolerable, the curve characterizes the difficulty of the ELM with respect to the algorithm. 

A single numeric number that characterizes the difficulty can be the area under the curve (AUC) for a given $\epsilon_{\rm max}$.  this is illustarted by the shadowed area in  \ref{fig:difficulty_measure}.(c) for ELM II. When $AUC$ is close to $1$, the task is easy, and when $AUC$ is close to $0$, learning is impossible.  

In a learning problem, we can set different conditions which correspond to a range of ELMs. The difficulty measures of these ELMs can be visualized in the space of the parameters as a difficulty map.   We will show such maps in experiment III.

\subsection{MCMC moves in the model space}
To design the Markov chain moves in the model space $\mathbb{R}$, we use two types of proposals in the metropolis-Hastings design in equation~(\ref{eq:acceptance}).

1, A random proposal probability $Q(x,y)$ in the neighborhood of the current model $x$. 

2, Data augmentation. A significant portion of non-convex optimization problems involve latent variables. For example, in the clustering problem, the class label of each data point is latent. For such problems, we use data augmentation \cite{tanner1987calculation} to improve the efficiency of sampling. In order to propose a new model $y=x_{t+1}$, we first sample the values of the latent variables $Z_t$ based on $p(Z_t|x_t)$ and then sample the new model $x_{t+1}$ based on $p(x_{t+1}|Z_t)$. The proposal $y=x_{t+1}$ is then either accepted or rejected based on the same acceptance probability in Equation~\ref{eq:acceptance}. 

Note that, however, our goal in ELM construction is to traverse the model space instead of sampling from the probability distribution. When enough samples are collected and therefore the weights $\gamma_{ij}$ become large, the reweighted probability distribution would be significantly different from the original distribution $\pi(x)$ and the rejection rate of the models proposed via data augmentation would become high.
Therefore, we use the proposal probability based on data augmentation more often at the beginning and increasingly rely on random proposal when the weights become large.

\begin{figure}
    \center
 \includegraphics[width=0.6\columnwidth]{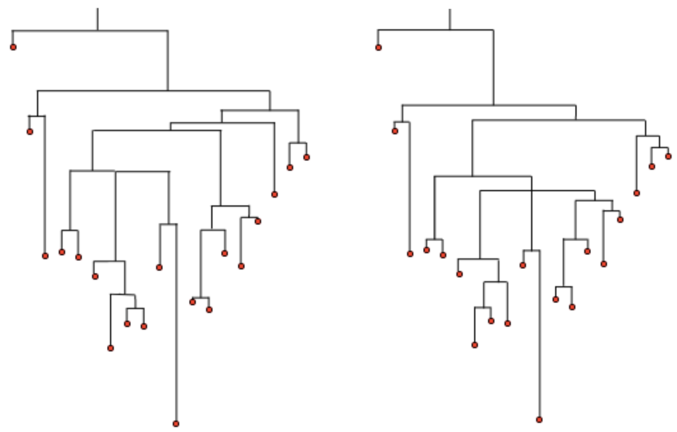}
\caption{ Two ELMs generated from two MCMC chains $C_1$ and $C_2$ initialized at different starting points after convergence in $24,000$ iterations.  }
\label{fig:ELM_variations}
\end{figure}

\subsection{ELM convergence analysis}

The convergence of the GWL algorithm to a stationary distribution is a necessary but not sufficient condition for the convergence of the ELMs.
As shown in Figure~\ref{fig:ELM_variations}, the constructed ELMs may have minor variations due to two factors: (i) the left-right ambiguity when we plot the branches under a barrier; and (ii) the precision of the energy barriers will affect the internal structure of the tree.

In experiments, firstly we monitor the convergence of the GWL in the model space. We run multiple MCMC initialized with  random starting values.  After a burn-in period, we collect samples and project in a 2-3 dimensional space using Multi-dimensional scaling. We  check whether the chains have converged to a stationary distribution using the multivariate extension of the Gelman and Rubin criterion~\cite{Gelman1}\cite{Gelman2}. 

\begin{figure}
    \center
 \subfigure[] {
   \includegraphics[width=0.38\columnwidth]{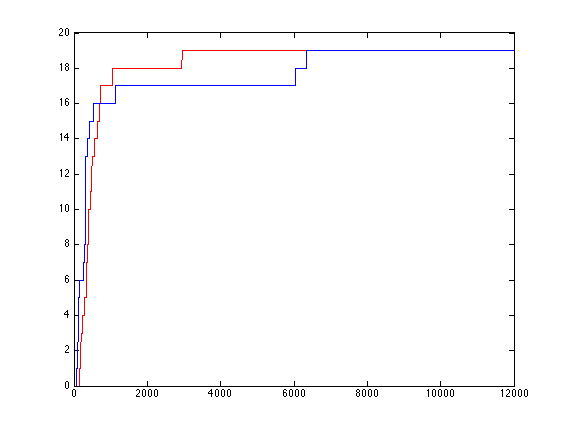}
  }
 \subfigure[] {
 \includegraphics[width=0.5\columnwidth]{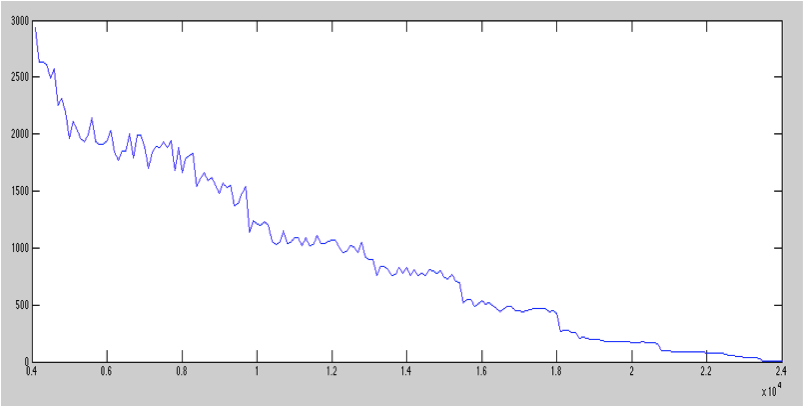}
  }
\caption{ Monitoring the convergence of ELMs generated from two MCMC chains $C_1$ and $C_2$ initialized at different starting points.  (a) The number of local minima found vs number of iterations for $C_1$ and $C_2$.  (b) the distance between the two ELMs vs. number of iterations.}
\label{fig:convergence}
\end{figure}

Once the GWL is believed to have converged, we can monitor the convergence of the ELM by checking the convergence of the following two sets over time $t$.
\begin{enumerate}
\item The set of leaf notes of the tree $S_L^t$ in which each point $x$ is a local
 minimum with energy $E(x)$. As $t$ increase, $S_L^t$ grows monotonically until no
 more local minimum is found, as is shown in Figure~\ref{fig:convergence}.(a). 

\item The set of internal nodes of the tree $S_N^t$ in which each point $y$ is an
 energy barrier at level $E(y)$. As $t$ increases, we may find lower barrier as the 
 Markov chain crosses different ridge between the basins. Thus $E(y)$ decreases
 monotonically until no barrier in $S_N^t$ is updated during a certain time period.
\end{enumerate}
 
 We further calculate a distance measure between two ELMs constructed by two MCMCs with different initialization. To do so, we compute a best node matching between the two trees and then the distance is defined on the differences of the matched leaf nodes and barriers, and penalties on unmatched nodes. We omit the details of this definition as it is not important for this work.  Figure~\ref{fig:convergence}.(b) shows the distance decreases as more samples are generated. 

\section{Experiment I: ELMs of Gaussian Mixture Models}\label{sec:gmm}

In this section, we compute the ELMs for learning Gaussian mixture models for two purpuses:  (i) study the influences of different conditions, such as separability and level of supervision; and ii) compare the behaviors and performances of popular algorithms including K-mean clustering, EM (Expectation-Maximization), two-step EM, and Swendson-Wang cut. We will use both synthetic data and real data in the experiments.

\subsection{Energy and Gradient Computations}

A Gaussian mixture model $\Theta$ with $n$ components in $d$ dimensions have weights $\alpha_i$, means $\mu_i$ and covariance matrices $\Sigma_i$ for $i=1,\dots,n$. Given a set of observed data points $\{z_i, i=1,...,m\}$,  we write the energy function as   
\begin{eqnarray}\label{eq:gmm_energy}
 E(\Theta) &=& - \log P( z_i : i=1\dots m | \Theta ) - \log P(\Theta) \\
      &=& - \sum^m_{i=1} \log f(z_i | \Theta)  - \log P(\Theta).
\end{eqnarray}
 $P(\Theta)$  is the product of a Dirichlet prior and a NIW prior. Its partial derivatives are trivial to compute. $f(z_i | \Theta) = \sum^n_{j=1} \alpha_j G(z_i; \mu_j, \Sigma_j)$ is the likelihood for data $z_i$, where $G(z_i; \mu_j, \Sigma_j) = \frac{1}{{ \sqrt {\det(2\pi \Sigma_j )} }} \exp \left[-\frac{1}{2}\left( {z_i - \mu_j } \right)^T \Sigma_j^{-1} \left( {z_i - \mu_j } \right)\right]$ is a Gaussian model.
 
For a sample $z_i$, we have the following partial derivatives of the log likelihood for calculating the gradient in the energy landscape.

a), Partial derivative with respect to each weight $\alpha_j$:
\begin{align*}
\frac{\delta \log f(z_i) }{\delta \alpha_j} = \frac{ G(z_i; \mu_j, \Sigma_j)} {\sum^K_{k=1} \alpha_k G(z_i, \mu_k, \Sigma_k)}.
\end{align*}

b), Partial derivative with respect to each mean $\mu_j$:
\begin{align*}
\frac{\delta \log f(z_i) }{\delta \mu_j} =  \frac{\alpha_j G(z_i; \mu_j, \Sigma_j)}{ \sum^K_{k=1} \alpha_k G(z_i; \mu_k, \Sigma_k)} \Sigma_j^{-1}(\mu_j - z_i ).
\end{align*}

c), Partial derivative with respect to each covariance $\Sigma_j$:
\begin{align*}
\frac{\delta \log f_\text{mm}(z_i) }{\delta \Sigma_j} & =  \frac{\alpha_j G(z_i; \mu_j, \Sigma_j)} { \sum^K_{k=1} \alpha_k G(z_i; \mu_k,\Sigma_k)} 
\frac{1}{2} \left[ 
\frac{\delta}{\delta \Sigma_j}\log \alpha_j G(z_i; \mu_j, \Sigma_j) \right] \\
& = \frac{\alpha_j G(z_i; \mu_j, \Sigma_j)} { \sum^K_{k=1} \alpha_k G(z_i; \mu_k,\Sigma_k)}  \frac{1}{2} 
\left[ -\Sigma_j^{-T} + \Sigma_j^{-T} \left( {z_i - \mu_j } \right) \left( {z_i - \mu_j } \right)^T \Sigma_j^{-T}
\right]
\end{align*}

During the computation, we need to restrict the $\Sigma_j$ matrices so that each inverse  $\Sigma_j^{-1}$ exists in order to have a defined gradient.
Each $\Sigma_j$ is semi-positive definite, so each eigenvalue is greater than or equal to zero. Consequently we only need the minor restriction
that for each eigenvalue $\lambda_i$ of $\Sigma_j$, $\lambda_i > \epsilon$ for some $\epsilon > 0$. However, it is possible that after one gradient descent step, the new GMM parameters will be outside of the valid GMM space, i.e. the new $\Sigma^{t+1}_j$ matrices at step $t+1$ will not be symmetric positive definite. Therefore, we need to project each $\Sigma^{t+1}_j$ into the symmetric positive definite space with the projection 
\begin{align*} 
P_{\text{symm}}(P_{\text{pos}}(\Sigma^{t+1}_j)).
\end{align*}
 The function $P_{\text{symm}}(\Sigma)$ projects the matrix into the space of symmetric matrices by 
 \begin{align*} 
 P_{\text{symm}}(\Sigma) = \frac{1}{2} (\Sigma+ \left(\Sigma\right)^T).
 \end{align*}
Assuming that $\Sigma$ is symmetric, the function $P_{\text{pos}}(\Sigma)$ projects $\Sigma$ into the space of symmetric matrices with eigenvalues greater than $\epsilon$. Because $\Sigma$ is symmetric, it can be decomposed into $\Sigma = Q \Lambda Q^T$ where $\Lambda$ is the diagonal eigenvalue matrix $\Lambda = diag\{\lambda_{1}, \dots, \lambda_{n}\}$, and $Q$ is an orthonormal eigenvector matrix. Then the function
 \begin{align*} 
P_{\text{pos}}(\Sigma) = Q \begin{pmatrix}
\max(\lambda_{1},\epsilon) &  0  & \ldots & 0\\
0  &  \max(\lambda_{2},\epsilon) & \ldots & 0\\
\vdots & \vdots & \ddots & \vdots\\
0  &   0       &\ldots & \max(\lambda_{n},\epsilon)
\end{pmatrix} Q^T
\end{align*}
ensures that $P_{\text{pos}}(\Sigma)$ is symmetric positive definite. 
 
\subsection{Bounding the GMM space}

From the $m$ data points $\{z_i, i = 1,\dots, m\}$, we can estimate a boundary of the space of possible parameter $\Theta$ if $m$ is sufficiently large. 


Let $\mu_o$ and $\Sigma_o$  be the sample mean and sample covariance matrix  of all $m$ points. We set a range for the means $\mu_j$ of the Gaussian components,
 \begin{align*} 
||\mu_j -  \mu_o ||_2 < \max_i || z_i - \mu_o||_2 + \epsilon_m.
\end{align*} 
$\epsilon_m$ is a constant that we will select in experiments.
To bound the covariance matrices $\Sigma_j$, let $\Sigma_o = Q \Lambda Q^T$ be the eigenvalue decomposition of $\Sigma_o$ with $\Lambda =diag\{\lambda_1, \cdots, \lambda_n\}$.
We denote by $L = \max(\lambda_1, \dots, \lambda_n) + \epsilon_m$ the upper bound of the eigen-values, and bound all the eigenvalues of $\Sigma_j$ by $L$.

\begin{figure}
    \center
    \subfigure[unbounded GMM space] {
     \includegraphics[width=0.45\columnwidth]{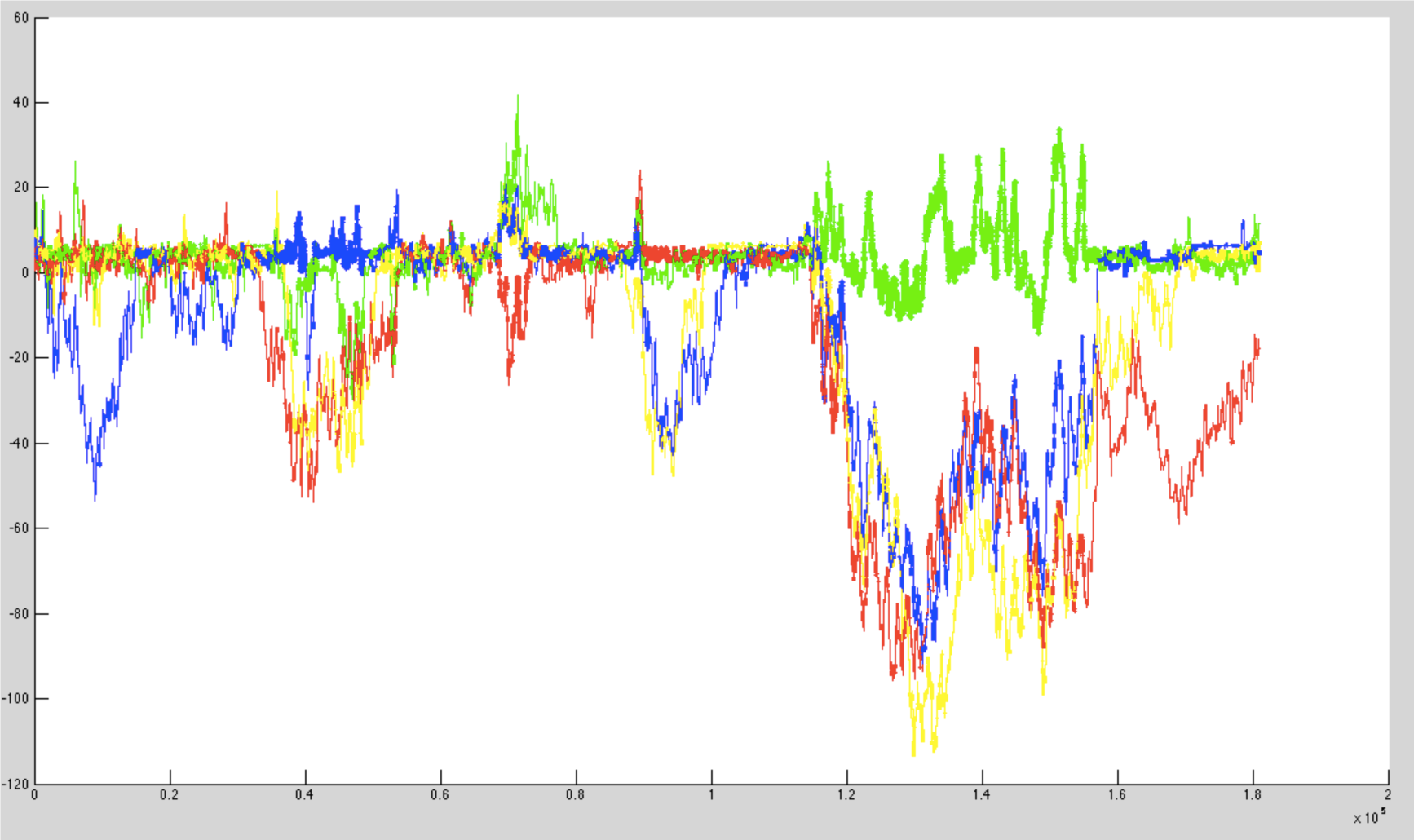}
    }
    \subfigure[bounded GMM space]{
     \includegraphics[width=0.38\columnwidth]{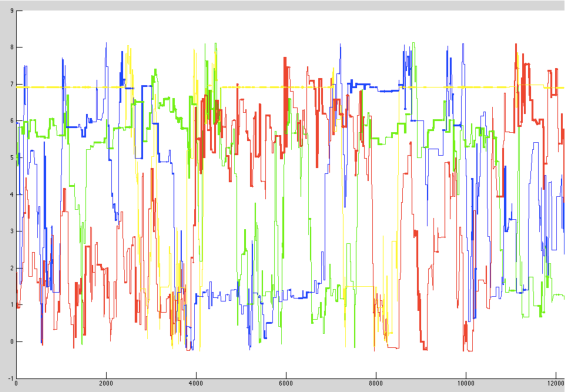}
    }
    \caption{We sampled 70 data points from a 1-dimensional 4-component GMM and ran the MCMC random walk for ELM construction algorithm in the (a) unbounded (b) bounded GMM space. The plots show the evolution of the location of the centers of the 4 components over time. The width of the line represents the weight of the corresponding component. } \label{fig:cluster-centers}
\end{figure}


Figure \ref{fig:cluster-centers} (a,b) compare the MCMCs in unbounded and bounded spaces repsectively.  We sampled $m=70$ data points from a 1-dimensional, 4-component GMM and ran the MCMC random walk for ELM construction algorithm. The plots show the evolution of the locations of $\mu_1,...,\mu_4$ over time. Notice that in Figure \ref{fig:cluster-centers} (a), the MCMC chain can move far from the center and spends the majority of the time outside of the bounded subspace. In Figure \ref{fig:cluster-centers} (b), by forcing the chain to stay within the boundary, we are able to explore the relevant subspace more efficiently. 

\subsection{Experiments on Synthetic Data}

We start with synthetic data with $k=3$ component GMM on $2$ dimensional space, draw $m$ samples and run our algorithm to plot the ELM under different settings.

\begin{figure}
    \center
        \includegraphics[width= \columnwidth]{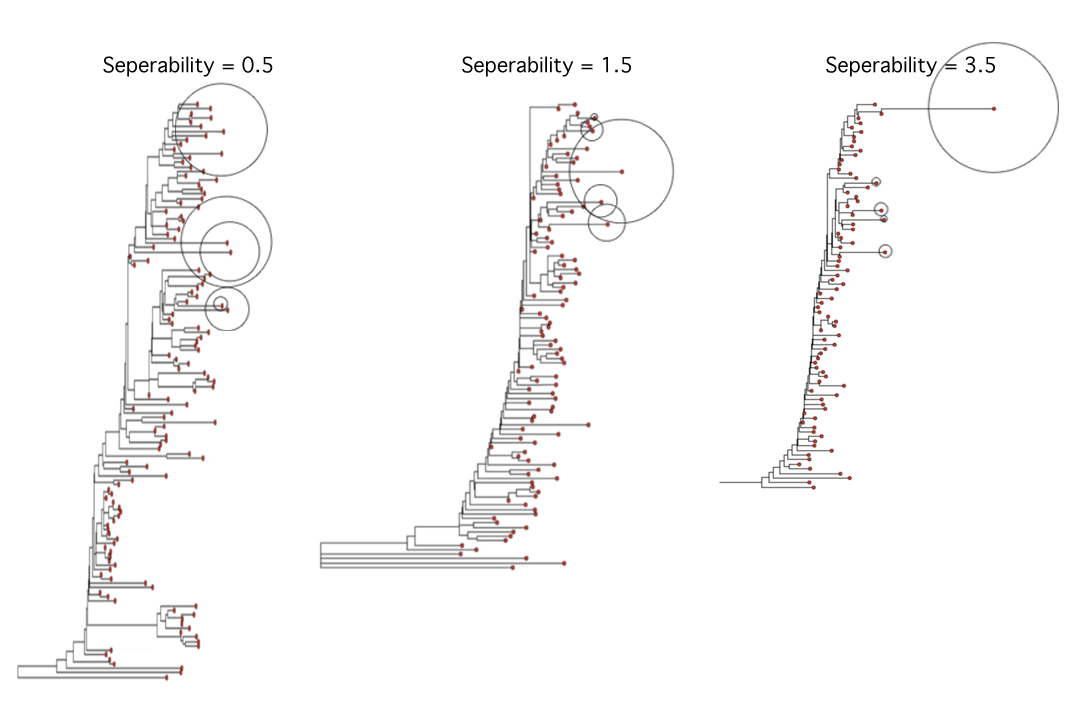}
    \caption{ELMs for $m=100$ samples drawn from GMMs with low, medium and high separability $c = 0.5, 1.5, 3.5$ respectively. The circle represents the probability mass of the basins. } \label{fig:trees-by-sep} 
\end{figure}

{\bf 1) The effects of separability}. The  separability of the GMM represents the overlap between components of the  model and is defined as $c = \min \left( \frac{||\mu_i - \mu_j|| }{\sqrt{n}\max(\sigma_1,\sigma_2)} \right)$.  This is often used in the literature to measure the difficulty of learning the true GMM model. 

Figure~\ref{fig:trees-by-sep} shows three representative ELMs with the separability $c=0.5, 1.5, 3.5$ respectively for $m = 100$ data points. This clearly shows that at $c=0.5$, the model is hardly identifiable with many local minima reaching similar energy levels. The energy landscape becomes increasingly simple as the separability increases. When $c = 3.5$, the prominent global minimum dominates the landscape.   

\begin{figure}
\center
 \includegraphics[width=\textwidth] {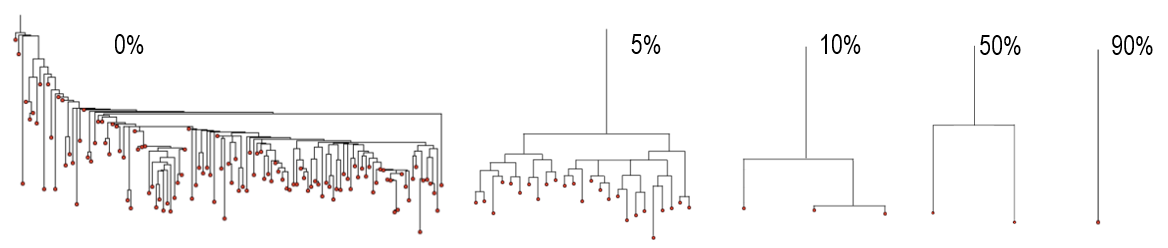}
    \caption{ ELMs with of synthesized GMMs (separability $c=1.0$, nSamples = 100) with $\{0\%, 5\%, 10\%, 50\%, 90\% \}$ labelled data points.} \label{fig:partial-lab-trees} 
\end{figure}
\begin{figure}
\center
 \includegraphics[width= 0.5\columnwidth] {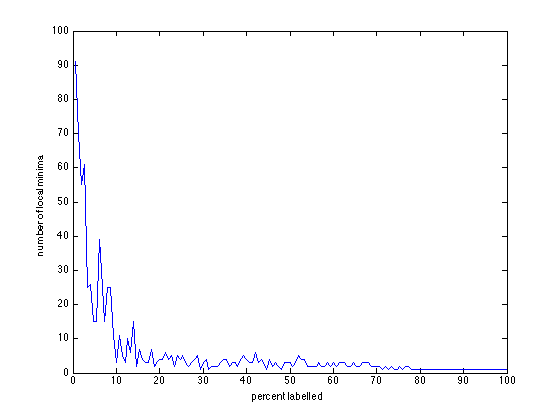}
    \caption{ Number of local minima versus the percentage of labelled data points for a GMM with separability $c=1.0$.  } \label{fig:lm_by_labels}
\end{figure}

{\bf 2) The effects of partial supervision}. We assign ground truth labels to a portion of  the $m$ data points. For $z_i$, its label $\ell_i$ indicates which component it belongs to. We set $m=100$, separability $c=1.0$.  Figure \ref{fig:partial-lab-trees} shows the ELMs with $0\%, 5\%, 10\%, 50\%, 90\%$ data points labels. While unsupervised learning ($0\%$) is very challenging, it becomes much simpler when $5\%$ or $10\%$ data are labeled. When $90\%$ data are labeled, the ELM has only one minimum.
Figure \ref{fig:lm_by_labels} shows the number of local minima in the ELM when labeling $1, \dots, 100$ samples. This shows a significant decrease in landscape complexity for the first 10\$ labels, and diminishing returns from supervised input after the initial 10\%. 

{\bf 3) Behavior of Learning Algorithms.} We compare the behaviors of the following algorithms under different separability conditions.
\begin{itemize}
\item Expectation-maximization (EM) is the most popular algorithms for learning GMM in statistics.
\item K-means clustering is a popular algorithm in machine learning and pattern recognition.

\item Two-step EM is a variant of EM proposed in \cite{dasgupta} who have proved a performance guarantee under certain separability conditions. It starts with an excessive number of components and then prune them. 

\item The Swedsen-Wang Cut (SW-cut) algorithm proposed in \cite{zhu-sw-cut} and \cite{zhu-sw-cut-pami}. This generalizes the SW method \cite{sw-cut} from Ising/Potts models to arbitrary probabilities. 

\end{itemize}
We modified EM, two-step EM and SW-cut in our experiments so that they minimize the energy function defined in Equation \ref{eq:gmm_energy}. K-means does not optimize our energy function, but it is frequently used as an approximate algorithm for learning GMM and therefore we include it in our comparison.

\begin{figure}
    \center
        \includegraphics[width=\columnwidth]{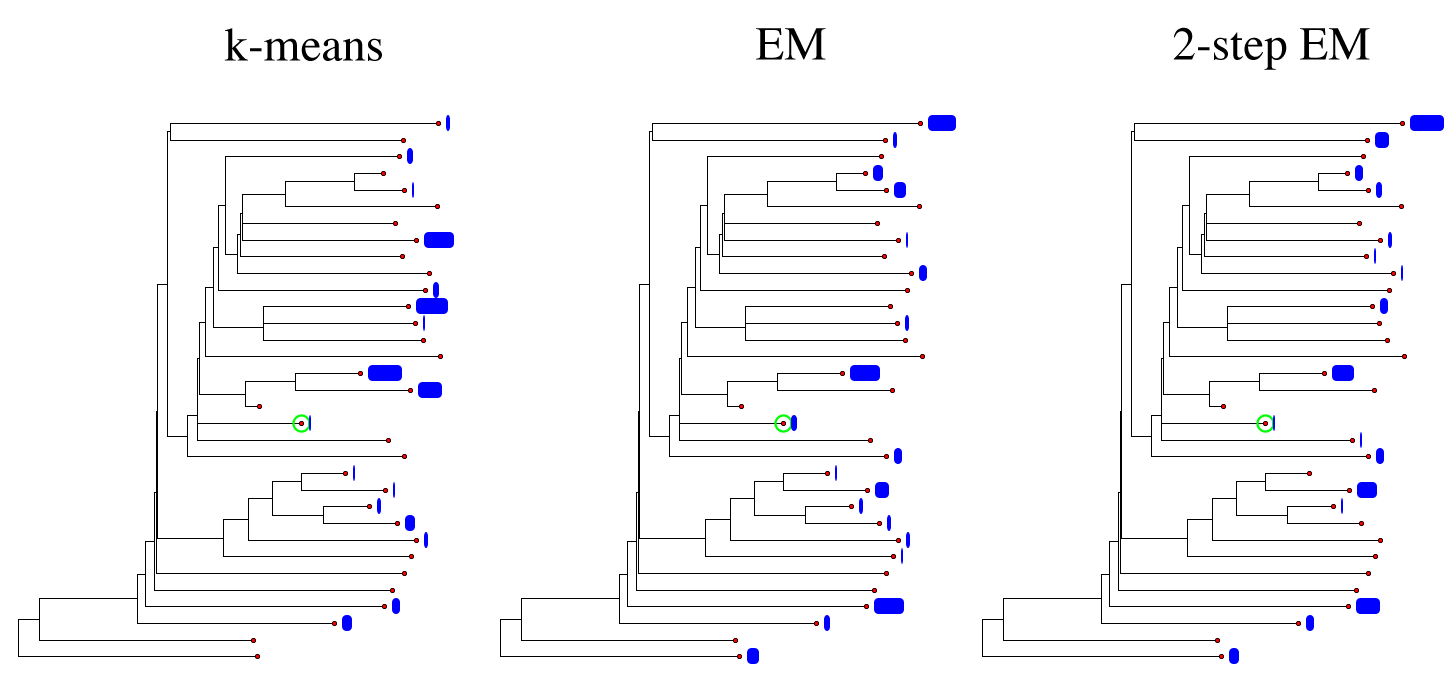}
    \caption{ The performance of the k-means, EM and 2-step EM algorithms on the ELMs with 10 samples drawn from a GMM with low separability ($c = 0.5$) } \label{fig:scatteredResults}
\end{figure}

\begin{figure}
    \center
    \subfigure[EM] {
        \includegraphics[width=0.30\textwidth,height=150pt]{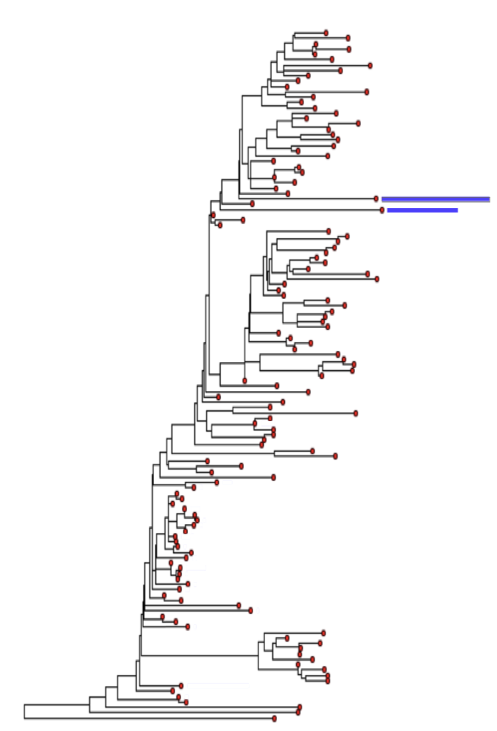}
    } 
    \subfigure[k-means]{
        \includegraphics[width=0.30\textwidth,height=150pt]{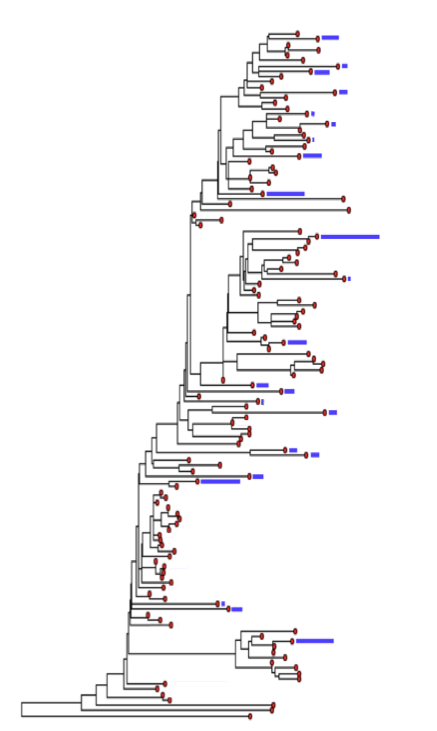}
    }
    \subfigure[SW-cut]{
        \includegraphics[width=0.33\textwidth,height=150pt]{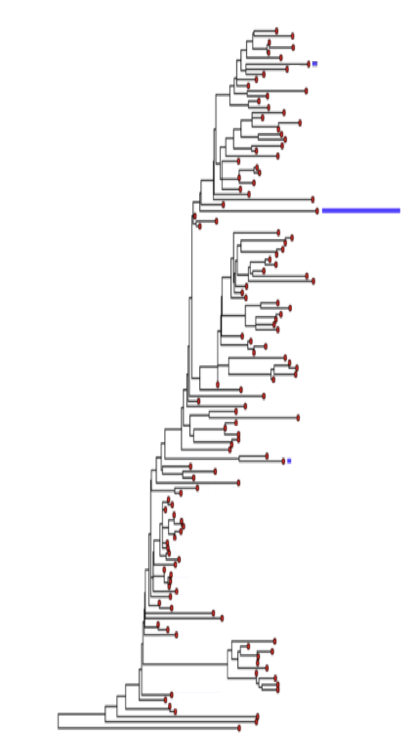}
    }
    
    \subfigure[EM] {
        \includegraphics[width=0.27\textwidth,height=120pt]{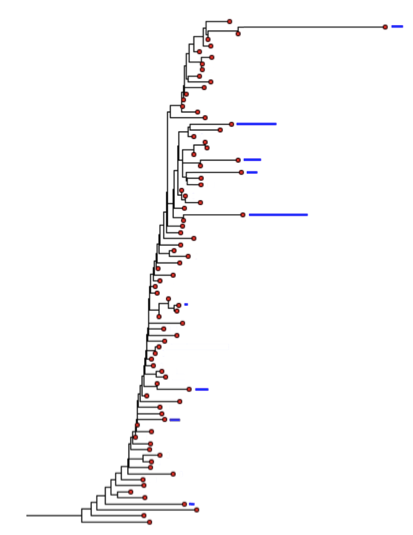}
    } 
    \subfigure[k-means]{
        \includegraphics[width=0.33\textwidth,height=120pt]{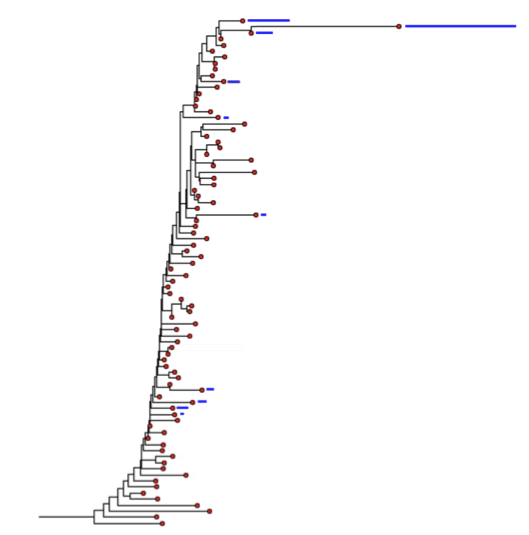}
    }
    \subfigure[SW-cut]{
        \includegraphics[width=0.33\textwidth,height=120pt]{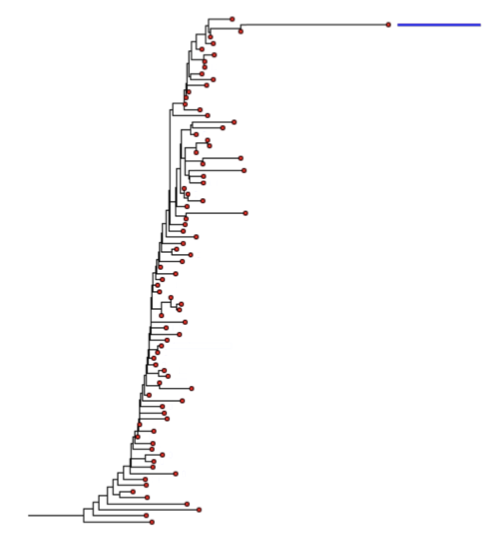}
    }
    \caption{ The performance of the EM, k-means, and SW-cut algorithm on the ELM. (a-c) Low separability $c = 0.5$. (d-f) High separability $c = 3.5$.  } \label{fig:behavior} 
\end{figure}

For each synthetic dataset in the experiment, we first construct the ELM, and then ran each of the algorithms for $200$ times and record which of the energy basins the algorithm lands to. Hence we obtain the visiting frequency of the basins by each algorithm, which are shown as bars of varying length at the leaf nodes in Figures~\ref{fig:scatteredResults} and \ref{fig:behavior}.

Figure~\ref{fig:scatteredResults} shows a comparison between the K-means, EM and two-step EM algorithms for $n=10$ samples drawn from a low  ($c=0.5$) separability GMM. The results are scattered across different local minima regardless of the algorithm. This illustrates the difficulty in learning a model from a landscape with many local minima separated by large energy barriers.


Figure~\ref{fig:behavior} show a comparison of the EM, k-means, and SW-cut algorithms for $m=100$ samples drawn from low ($c = 0.5$) and high ($c = 3.5$) separability GMMs. The SW-cut algorithm performs best in each situation, always converging to the global optimal solution.  In the low separability case, the k-means algorithm is quite random, while the EM algorithm almost always finds the global minimum and thus outperforms k-means.  However, in the high separability case, the k-means algorithm converges to the true model the majority of the time, while the EM almost always converges to a local minimum with higher energy than the true model. This result confirms a recent theoretical result showing that the objective function of hard-EM (with k-means as a special case) contains an inductive bias in favor of high-separability models \cite{Tu12,samdani2012unified}. Specifically, we can show that the actual energy function of hard-EM is:
\[
	E(\Theta) = -\log P(\Theta|Z) + \min_q \left( \textbf{KL}(q(L)||P(L|Z,\Theta)) + H_q(L) \right)
\]
where $\Theta$ is the model parameters, $Z={z_1,\ldots,z_m}$ is the set of observable data points, $L$ is the set of latent variables (the data point labels in a GMM), $q$ is an auxiliary distribution of $L$, and $H_q$ is the entropy of $L$ measured with $q(L)$. The first term in the above formula is the standard energy function of clustering with GMM. The second term is called a posterior regularization term \cite{Ganchev10}, which essentially encourages the distribution $P(L|Z,\Theta)$ to have a low entropy. In the case of GMM, it is easy to see that a low entropy in $P(L|Z,\Theta)$ implies high separability between Gaussian components.


\subsection{Experiments on Real Data}

We ran our algorithm to plot the ELM for the well-known Iris data set from the UCI repository \cite{iris}.  The Iris data set contains $150$ points in 4 dimensions and can be modeled by a 3-components GMM. The three components each represent a type of iris plant and the true component labels are known. The points corresponding to the first component are linearly separable from the others, but the points corresponding to the remaining two components are not linearly separable. 

\begin{figure}
    \center
 \includegraphics[width=\textwidth]{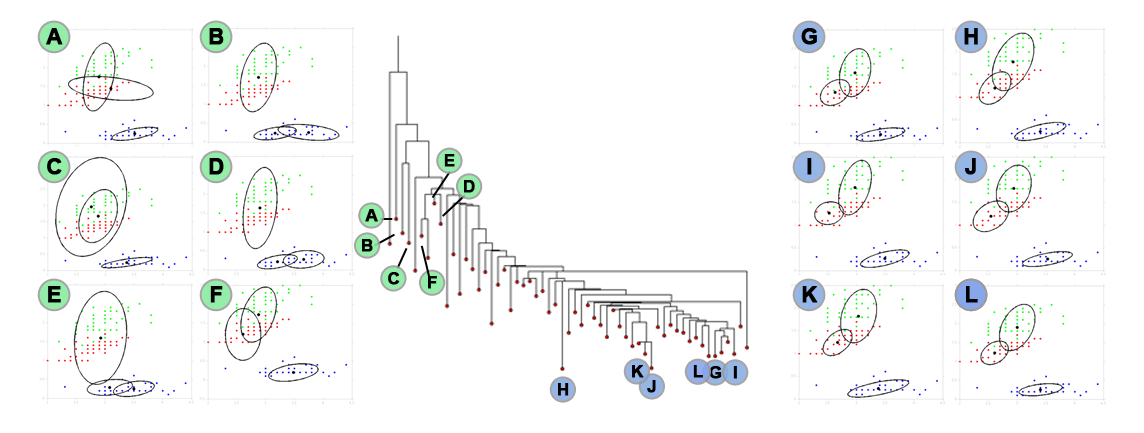}
    \caption{ ELM and some of the local minima of the Iris dataset. } \label{fig:iris} 
\end{figure}

Figure \ref{fig:iris} shows the ELM of the Iris dataset. We visualize the local minima by plotting the ellipsoids of the covariance matrices centered at the means of each component in 2 of the 4 dimensions.

The 6 lowest energy local minima are shown on the right and the 6 highest energy local minima are shown on the left. The high energy local minima are less accurate models than the low energy local minima. The local minima (E) (B) and (D) have the first component split into two and the remaining two (non-separable) components merged into one. The local minima (A) and (F) have significant overlap between the 2nd and 3rd components and (C) has the components overlapping completely. The low-energy local minima (G-L) all have the same 1st components and slightly different positions of the 2nd and 3rd components. 

\begin{figure}
\center
 \includegraphics[width=\textwidth] {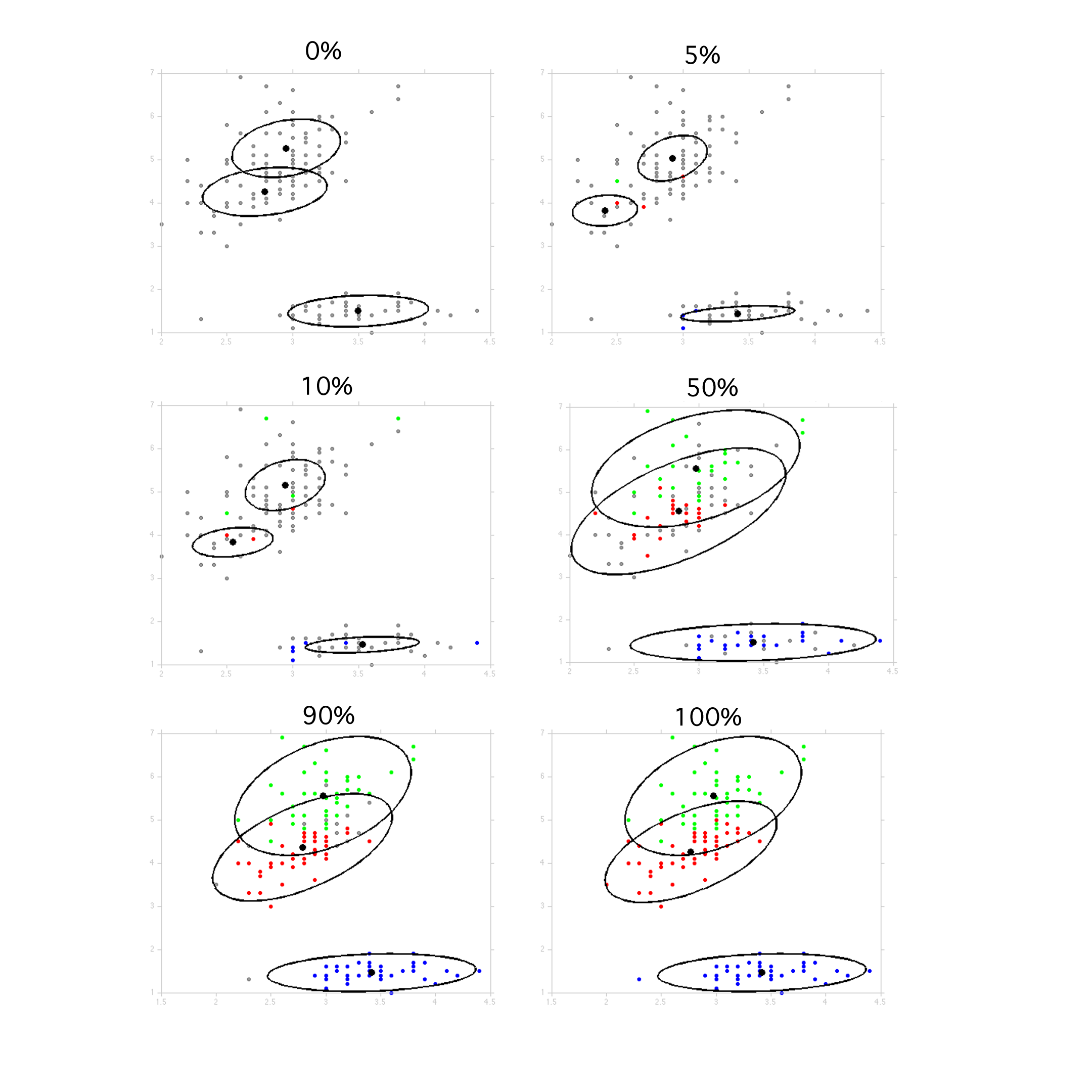}
    \caption{ Global minima for learning from the Iris dataset with 0, 5, 10, 50, 90, and 100\% of the data labeled with the ground truth values. Unlabeled points are drawn in grey and labelled points are colorized in red, green or blue.} \label{fig:partial-lab-iris} 
\end{figure}

We ran the algorithm with $0, 5, 10, 50, 90, 100$ percent of the points with the ground truth labels assigned. Figure \ref{fig:partial-lab-iris} shows the global minimum of the energy landscape for these cases.

\section{Experiment II: ELM of Bernoulli Templates} 

The synthetic data and Iris data in experiment I are in low dimensional spaces.  In this section, we experiment with very high dimensional data for a learning task in computer vision and pattern recognition.

\begin{figure*}
    \center
    \hspace{-5mm}\subfigure[cat] {
        \includegraphics[width=0.2\textwidth]{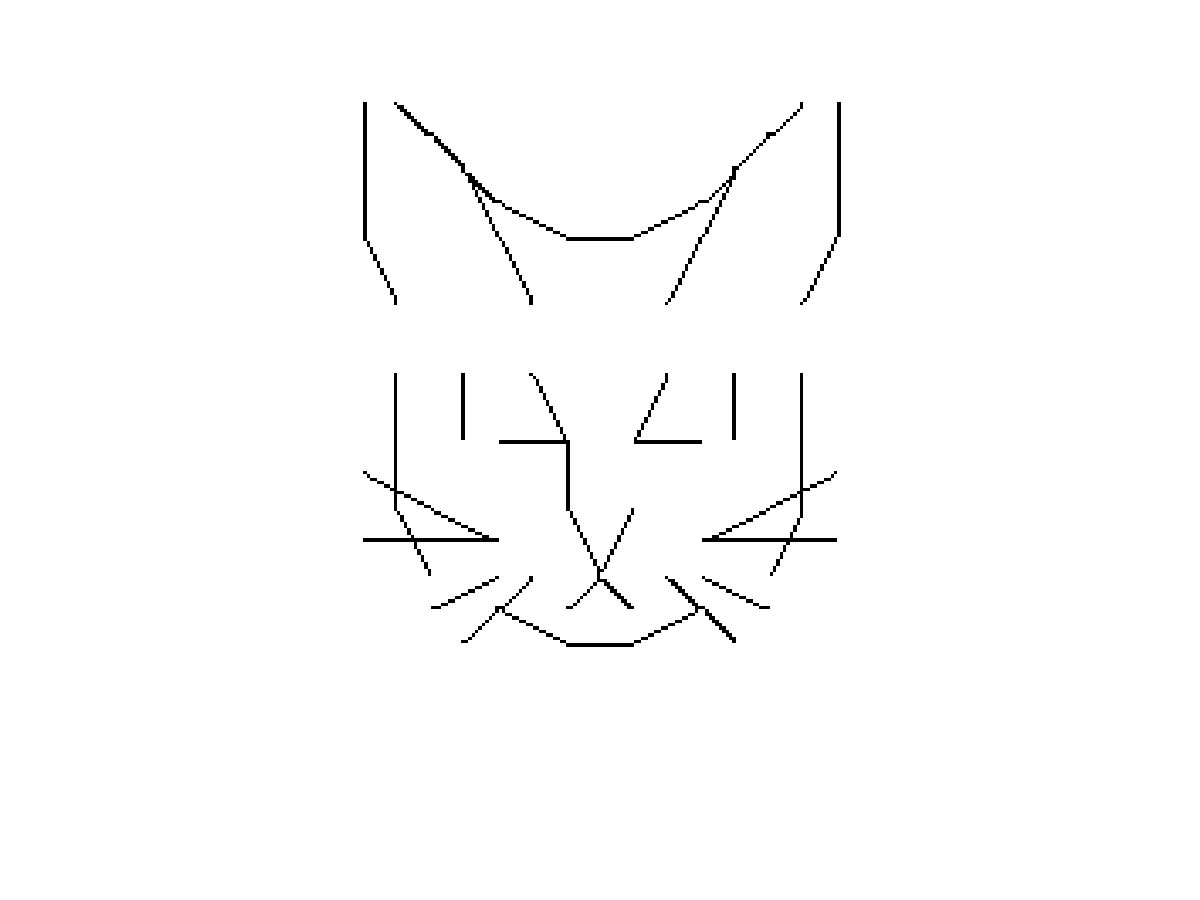}
    } 
    \hspace{-5mm} \subfigure[chilchilla]{
        \includegraphics[width=0.2\textwidth]{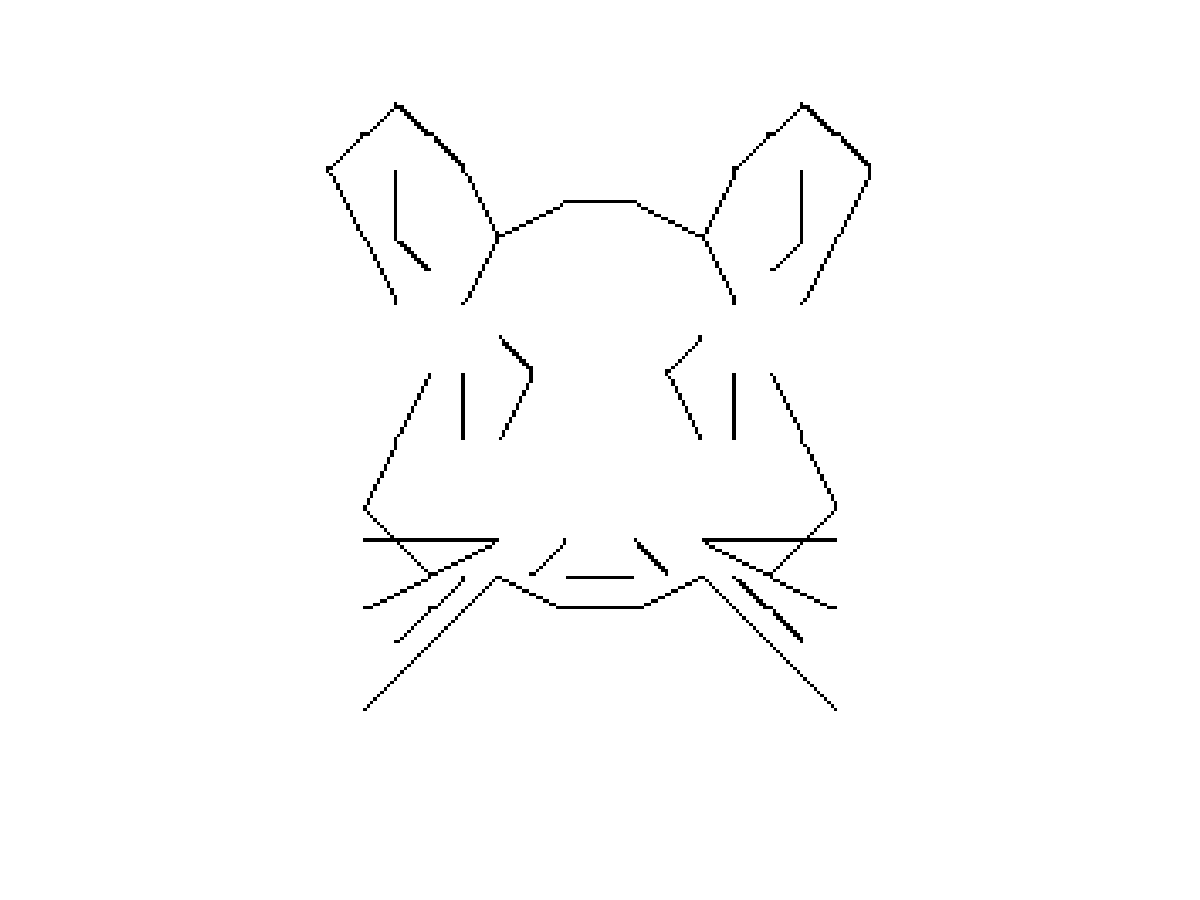}
    }
    \hspace{-5mm}\subfigure[dog]{
        \includegraphics[width=0.2\textwidth]{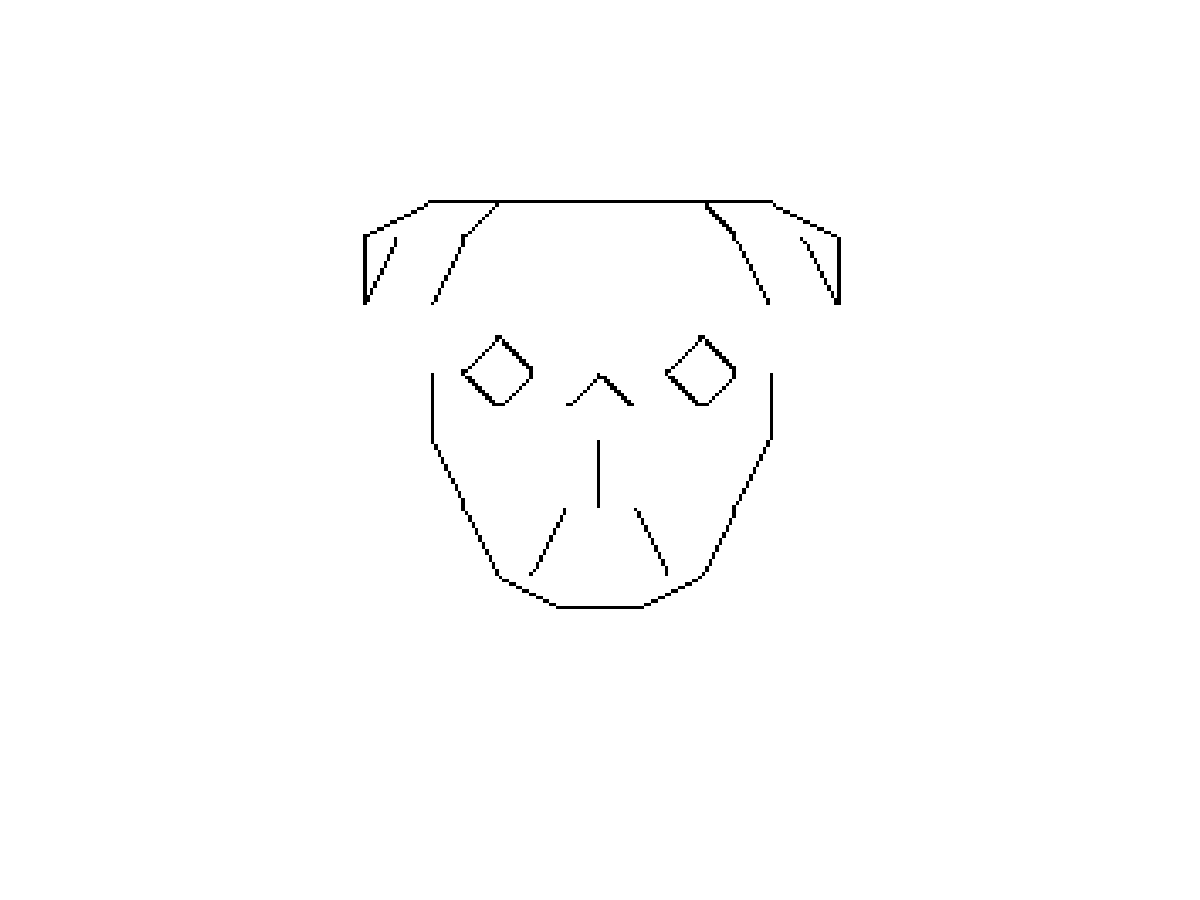}
    }
   \hspace{-5mm} \subfigure[elephant]{
        \includegraphics[width=0.2\textwidth]{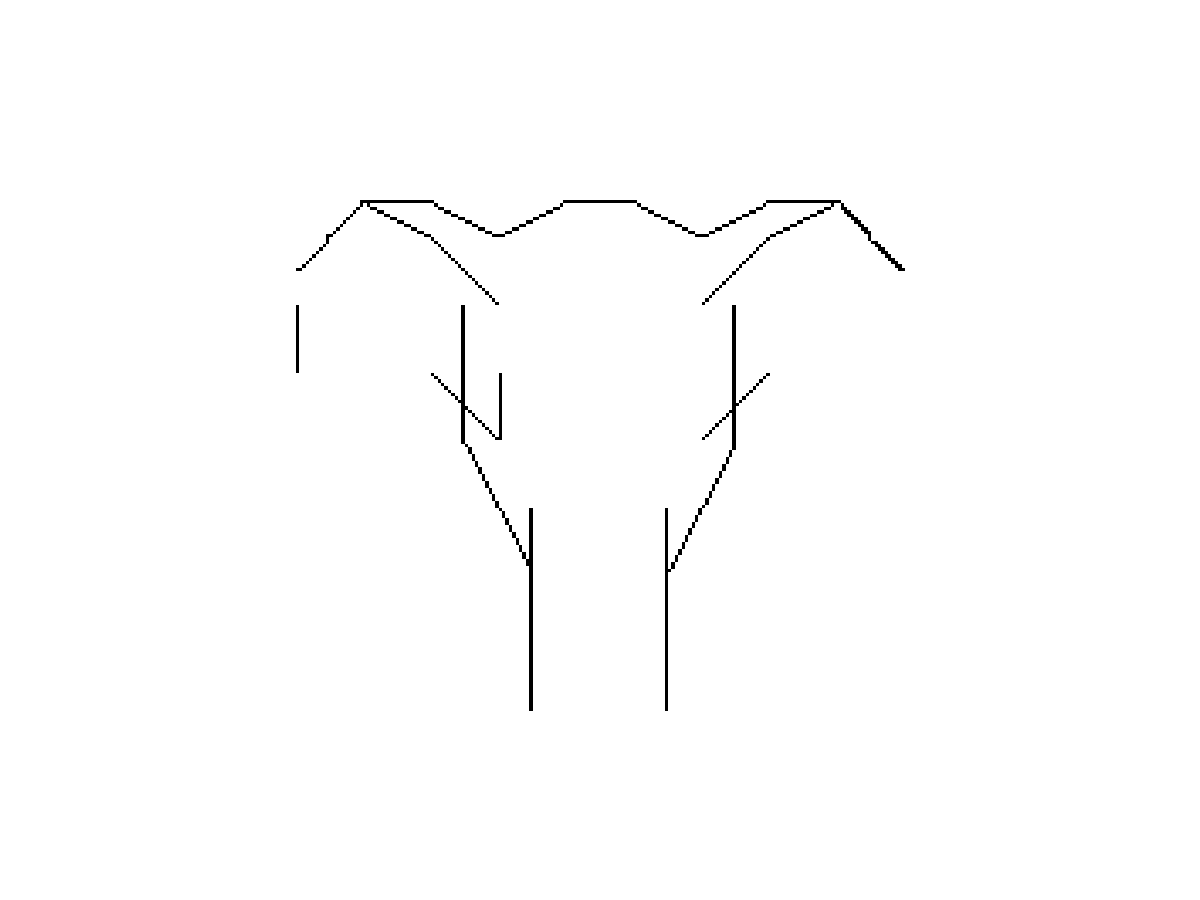}
    }
   \hspace{-5mm} \subfigure[goat]{
        \includegraphics[width=0.2\textwidth]{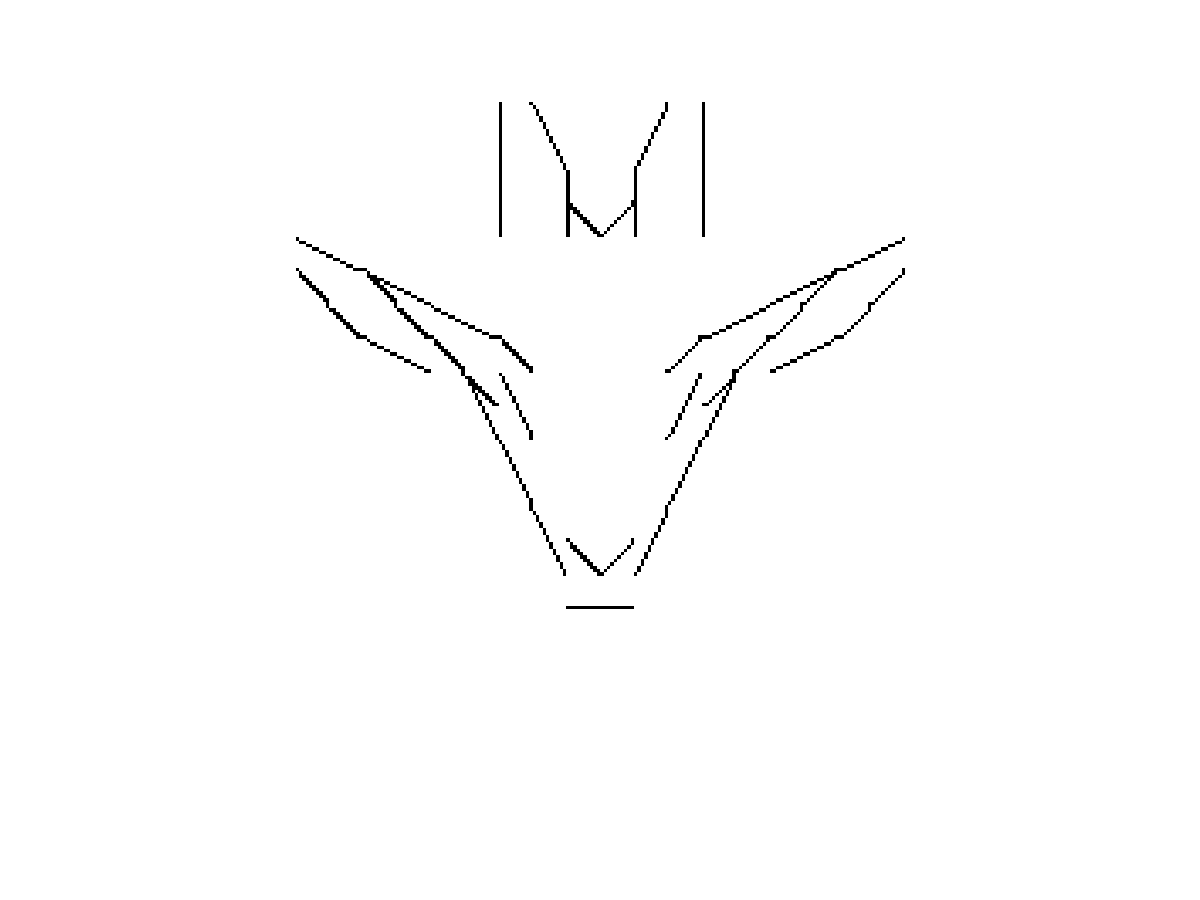}
    }\\
   \hspace{-5mm} \subfigure[lion]{
        \includegraphics[width=0.2\textwidth]{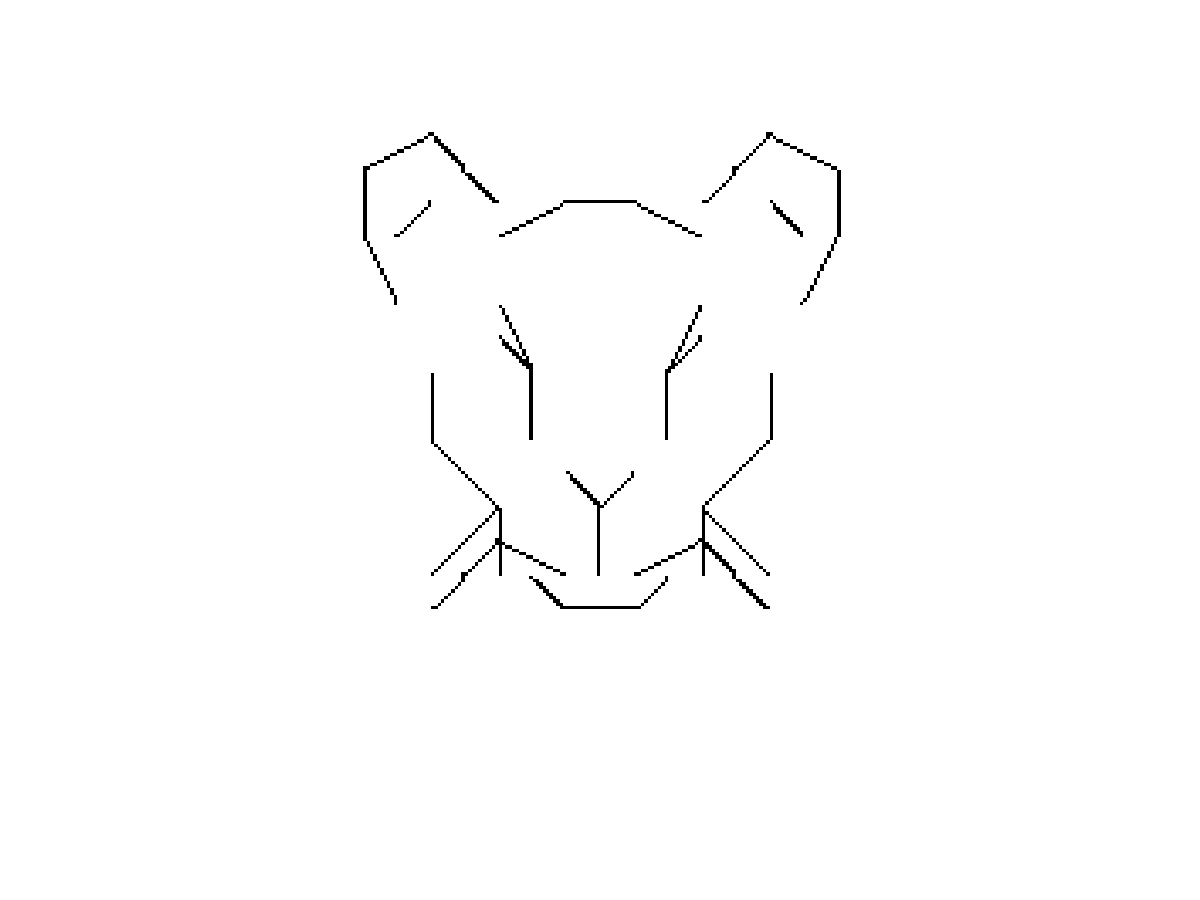}
    }
   \hspace{-5mm} \subfigure[monkey]{
        \includegraphics[width=0.2\textwidth]{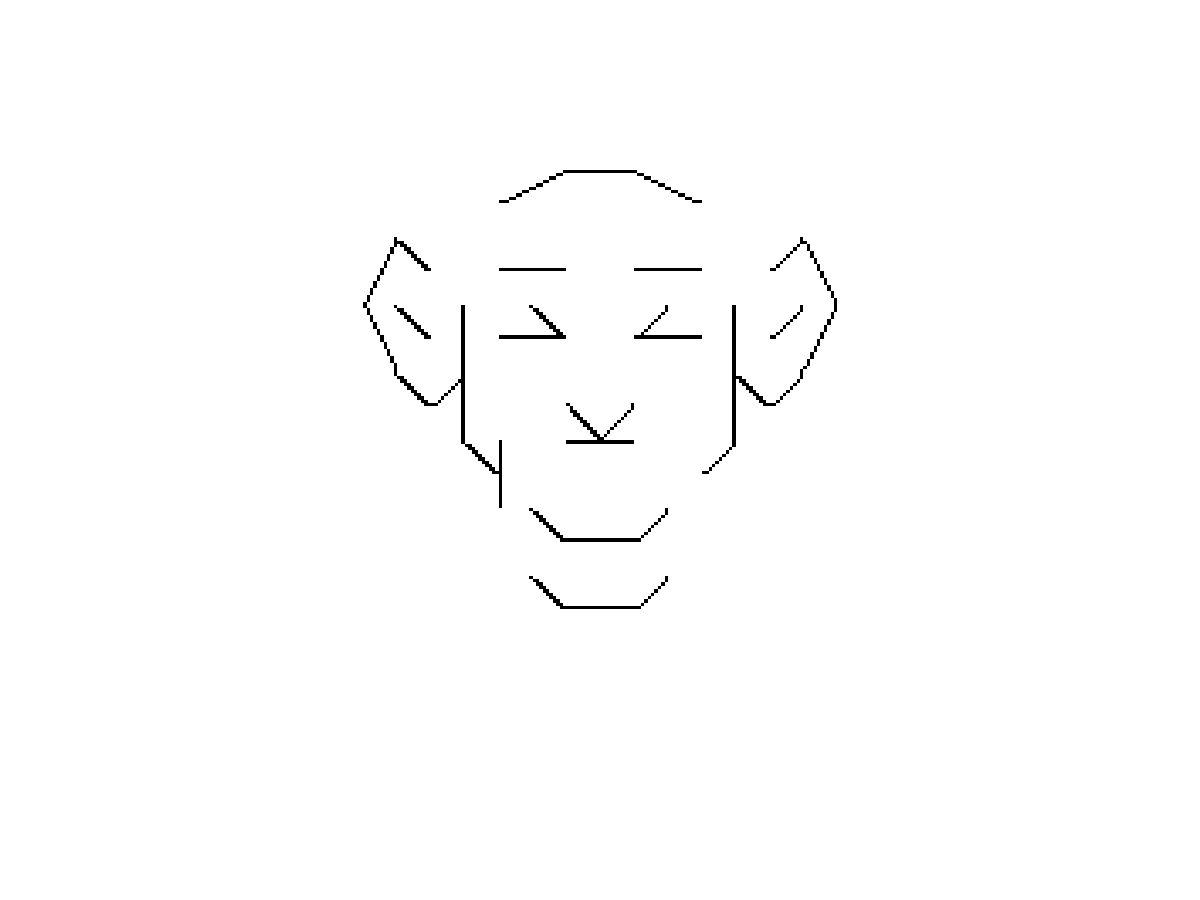}
    }
   \hspace{-5mm} \subfigure[mouse]{
        \includegraphics[width=0.2\textwidth]{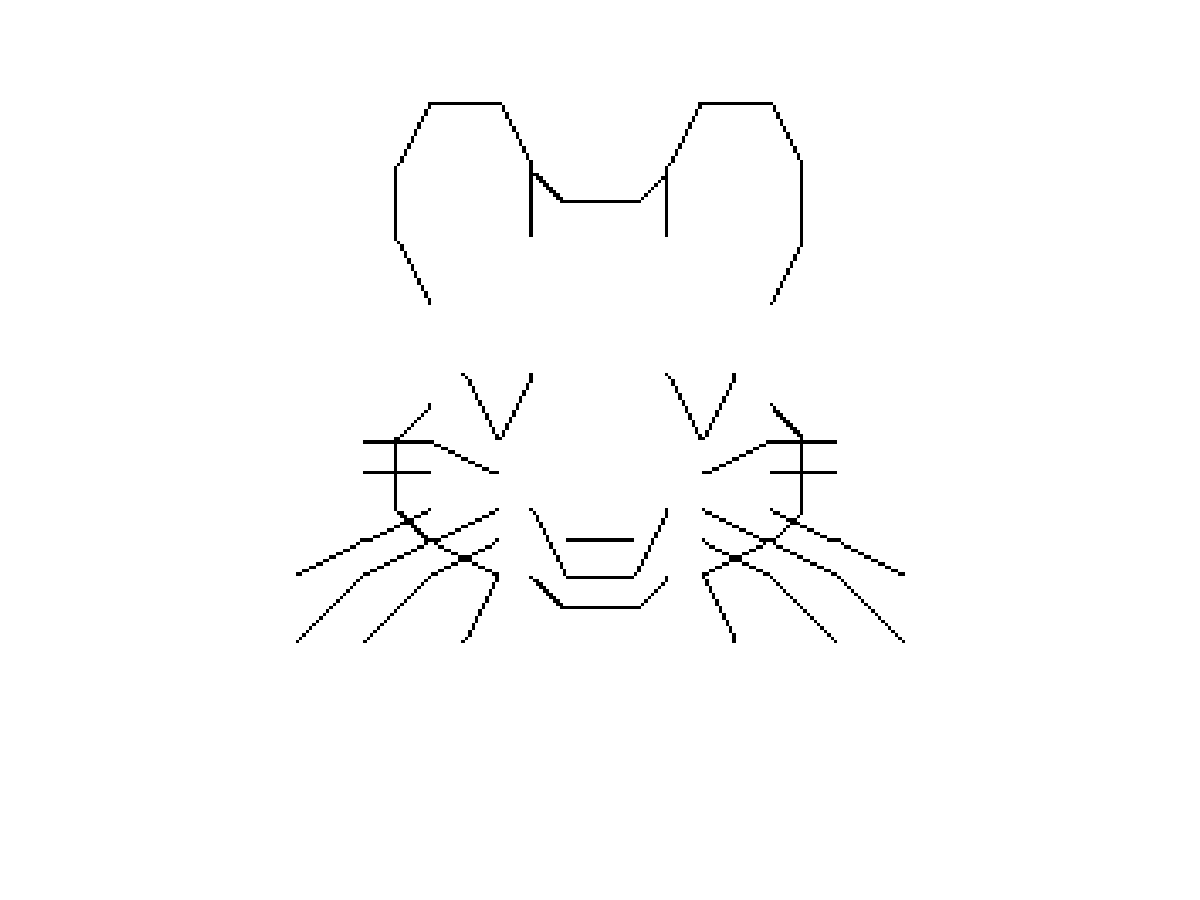}
    }
   \hspace{-5mm} \subfigure[owl]{
        \includegraphics[width=0.2\textwidth]{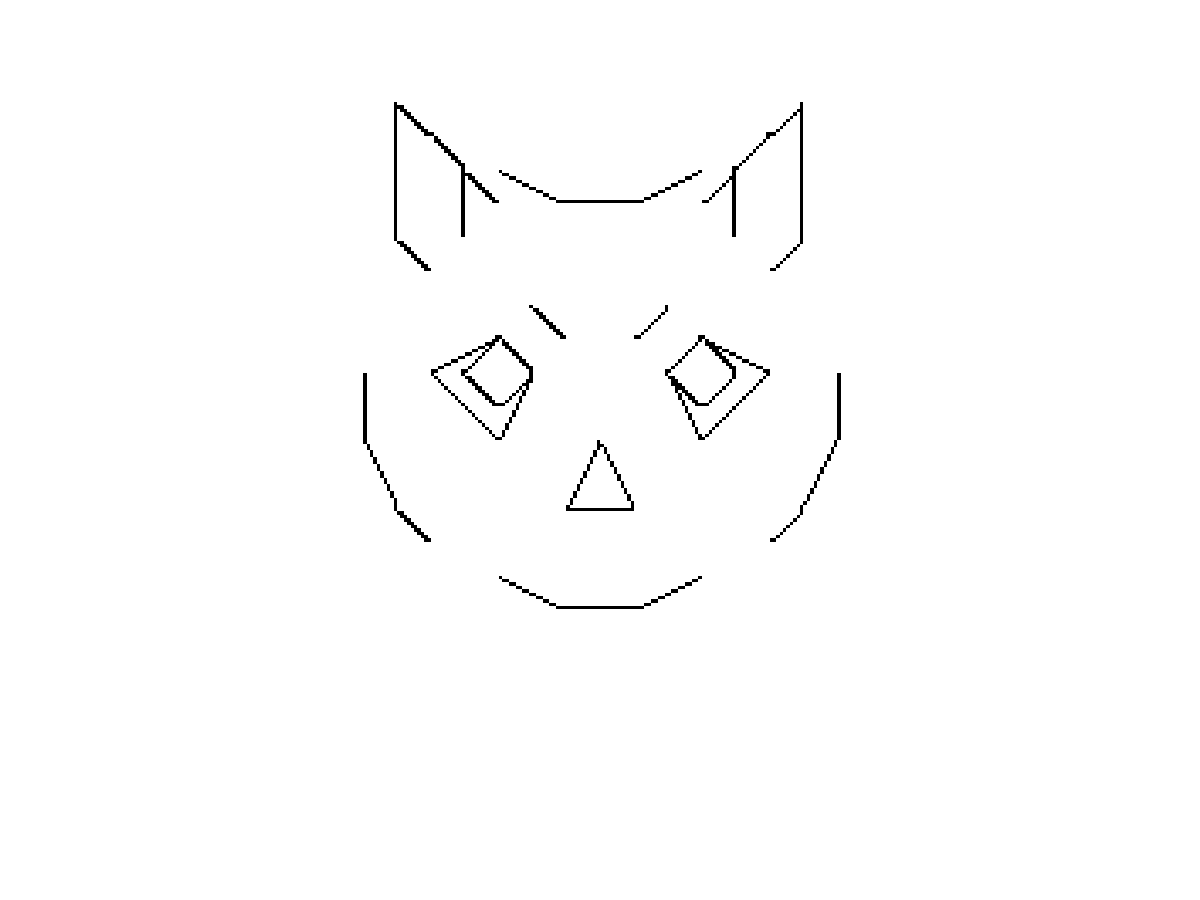}
    }
   \hspace{-5mm} \subfigure[rabbit]{
        \includegraphics[width=0.2\textwidth]{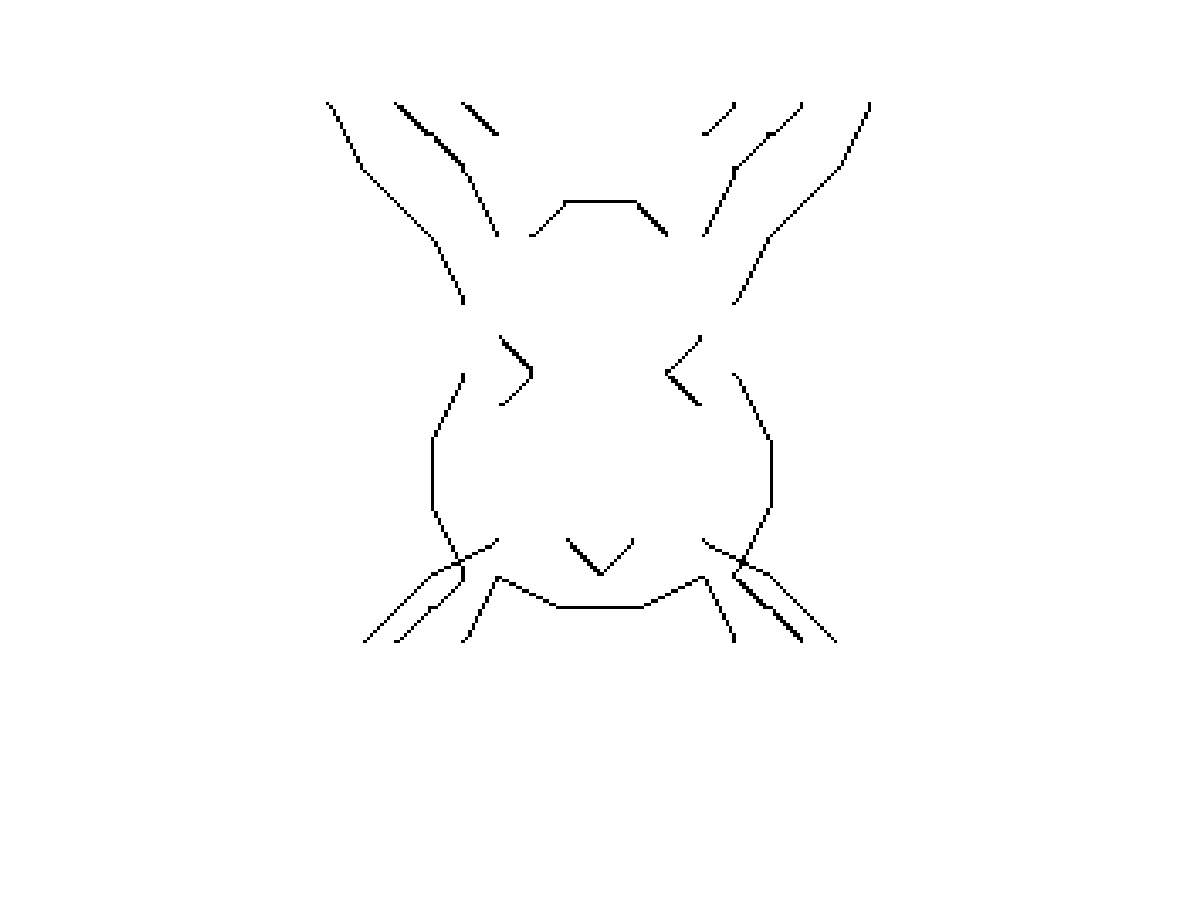}
    }
    \caption{Bernoulli templates for animal faces. These templates have low overlap and are well separable.} \label{fig:full_faces} 
\end{figure*}

The objective is to learn a number of templates ${\rm BT}_k, k=1,...,K$ for object recognition. Figure~\ref{fig:full_faces} illustrates $10$ templates of animal faces.  Each template consists of a number of sketches or edges in the image lattice, and is denoted by a Boolean vector ${\rm BT}_k =(s_{k1}, s_{k2}, \dots, s_{kn})$ with $n$ being the number of quantized positions and orientations of the lattice which is typically a large number $100\sim 1000$. $s_{kj}=1$ if there is a sketch at location $j$, and $s_{kj}=0$ otherwise. 
Images are generated from one of the $K$ templates with noise. Suppose $z_i =(r_{i1}, r_{12}, \dots, r_{in})$ is an image generated from template ${\rm BT}_k$, then $ r_{ij} = s_{kj} $ with probability $p$ and $ r_{ij} = 1- s_{kj}$ with probability $1-p$.
Thus we call ${\rm BT}_k, k=1,2...,K$ the Bernoulli templates. 
For simplicity we assume $p$ is fixed for all the templates and all the locations. 

The energy function that we use is the negative log of the posterior, given by $E(\Theta) = -\log P(\Theta |z_i : i=1\dots m)$ for $m$ examples $\{ z_i \}_{i=1}^m$. The model parameter $\Theta$ consists of the Boolean vectors ${\rm BT}_k =(s_{k1}, s_{k2}, \dots, s_{kn})$ and the mixture weights $\alpha_k$ for $k=1,...,K$.
By assuming a uniform prior we have
\begin{equation*}
P(\Theta | z_i: 1=1,...,m) =\prod_{i=1}^m \sum_{k=1}^K \alpha_k p^{\sum_{j=1}^n  1(r_{ij} = s_{kj})}(1-p)^{\sum_{j=1}^n 1(r_{ij} \neq s_{kj})},
\end{equation*}

In the following we present experiments on synthetic and real data.

\subsection{Experiments on Synthetic Data}

\cite{Barbu_2run_EM} proposes a Two-Round EM algorithm for learning Bernoulli templates with a performance bound that is dependent on the number of components $K$, the Beronouilli template dimension $n$, and noise level $p$. In this experiment, we examine how the ELM of the model space changes with these factors. We discretize the model space by allowing the weights to take values $\alpha_i \in \{0, 0.1, \dots, 1.0\}$. 
In order to adapt the GWL algorithm to the discrete space, we use coordinate descent in lieu of gradient descent. 

\begin{figure}
  \center
 \subfigure[Sketch dictionary]{
 \includegraphics[width=0.22\columnwidth]{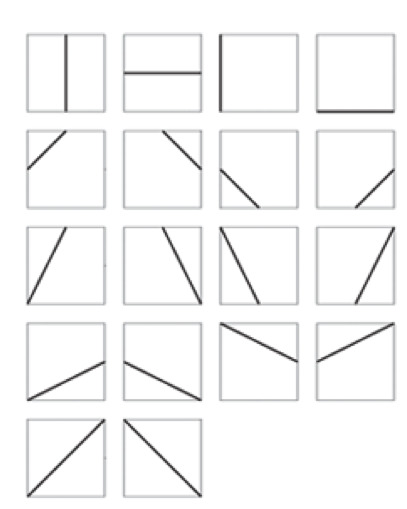}
 }
\hspace{10mm} \subfigure[Noisy dog]{
 \includegraphics[width=0.3\columnwidth]{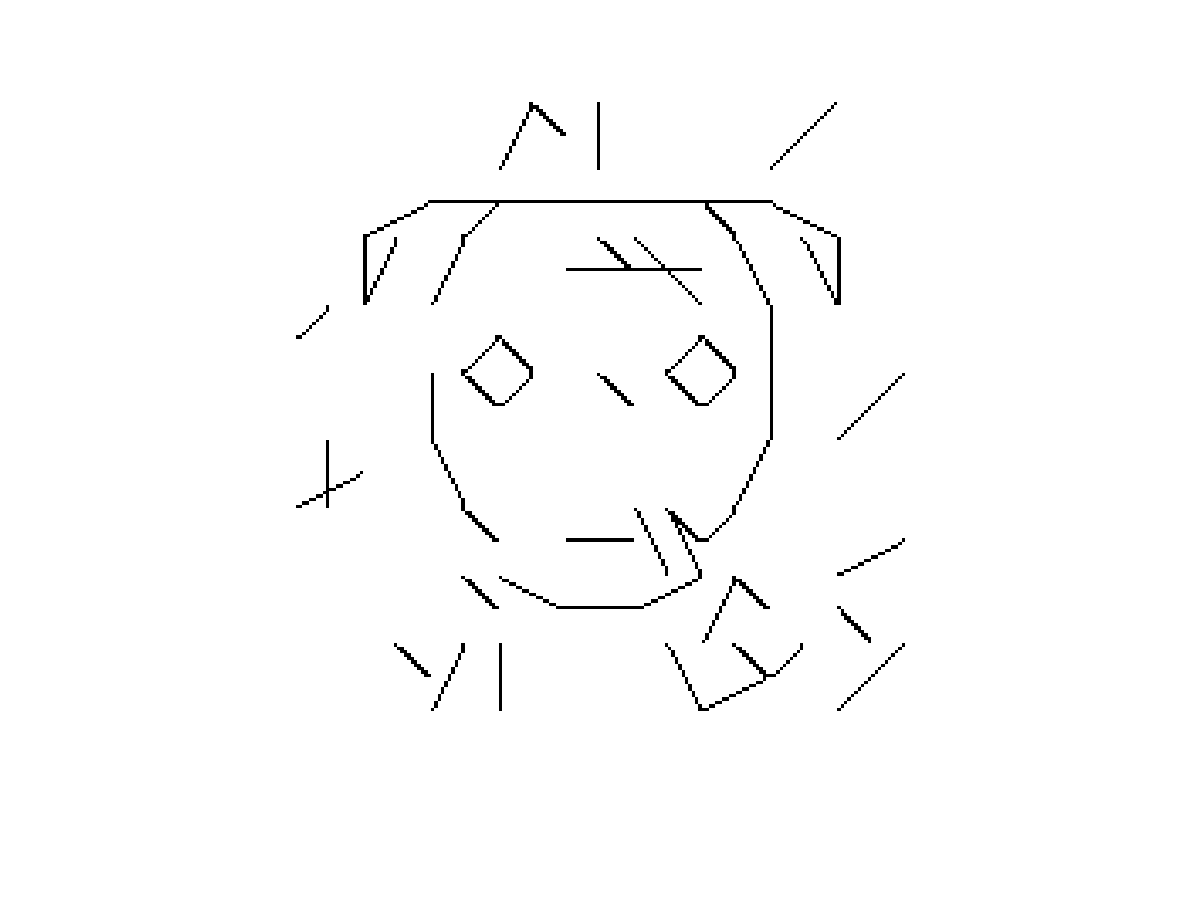}
 }
  \caption{(a) Quantized dictionary with 18 sketches for each cell in the image lattice. (b) Sample from the dog animal face template with noise level $p=0.1$} 
  \label{fig:sketch_dict} 
\end{figure}

We construct $10$ Bernouilli templates which represent animal faces in Figure \ref{fig:full_faces}. Each animal face is aligned to a grid of $9\times 9$ cells. Each cell may contain up to $3$ sketches. Within a cell, the sketches are quantized to $18$ discrete location and orientations. More specifically, each sketch is a straight line connecting the endpoints or midpoints of the edges of a square cell, and the $18$ possible sketches in a cell are shown  in Figure~\ref{fig:sketch_dict}.(a). They can well approximate the detected edges or Gabor sketches from real images. The Bernouilli template can therefore be represented as a $n=9\times 9 \times 18$ dimensional binary vector. Figure~\ref{fig:sketch_dict}.(b) shows a noisy dog face generated with noisy level $p=0.1$.   
 
\begin{figure}
    \center
  \subfigure[map]{
    \includegraphics[width=0.25 \columnwidth]{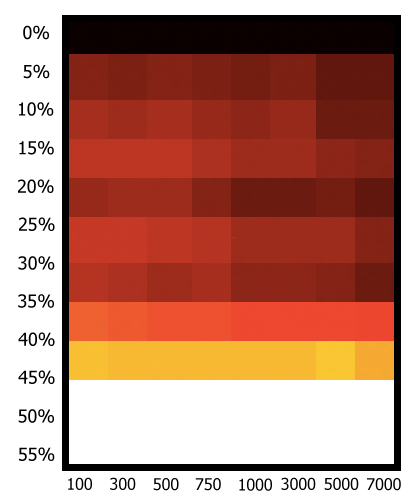}
     }
  \subfigure[number of local minima]{
    \includegraphics[width=0.7 \columnwidth]{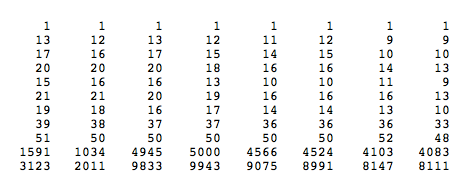}
     }
       \caption{ ELM complexity for varying values of $p$ and number of samples $m$ in learning the $10$  Bernouilli Templates. } \label{fig:full_faces_landscape} 
\end{figure}

We compute the ELMs of the Bernouilli mixture model for varying numbers of samples $m = 100, 300, \dots, 7000$ and varying noise level $p=0, 0.05, \dots, 0.5, 0.55$. The number of local minima in each energy landscape is tabulated in Figure \ref{fig:full_faces_landscape} (b) and drawn as a heat map in Figure \ref{fig:full_faces_landscape} (a). As expected, the number of local minima increases as the noise level $p$ increases, and decreases as the number of samples decreases. In particular, with no noise, the landscape is convex and with noise $p>0.45$, there are too many local minima and the algorithm does not converge.

\begin{figure*}
    \center
  \hspace{-4mm}   \subfigure[]{
        \includegraphics[width=0.14\textwidth]{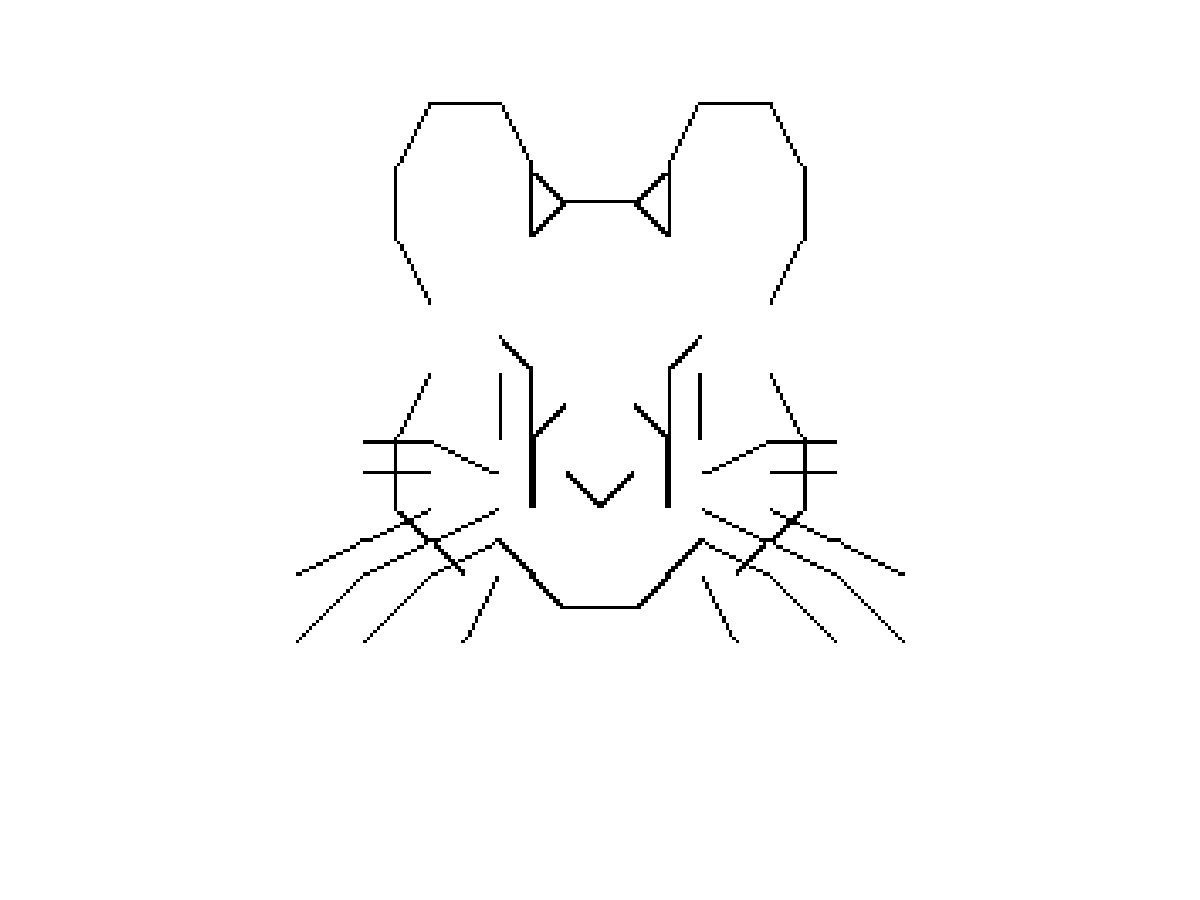}
    }
  \hspace{-4mm}   \subfigure[]{
        \includegraphics[width=0.14\textwidth]{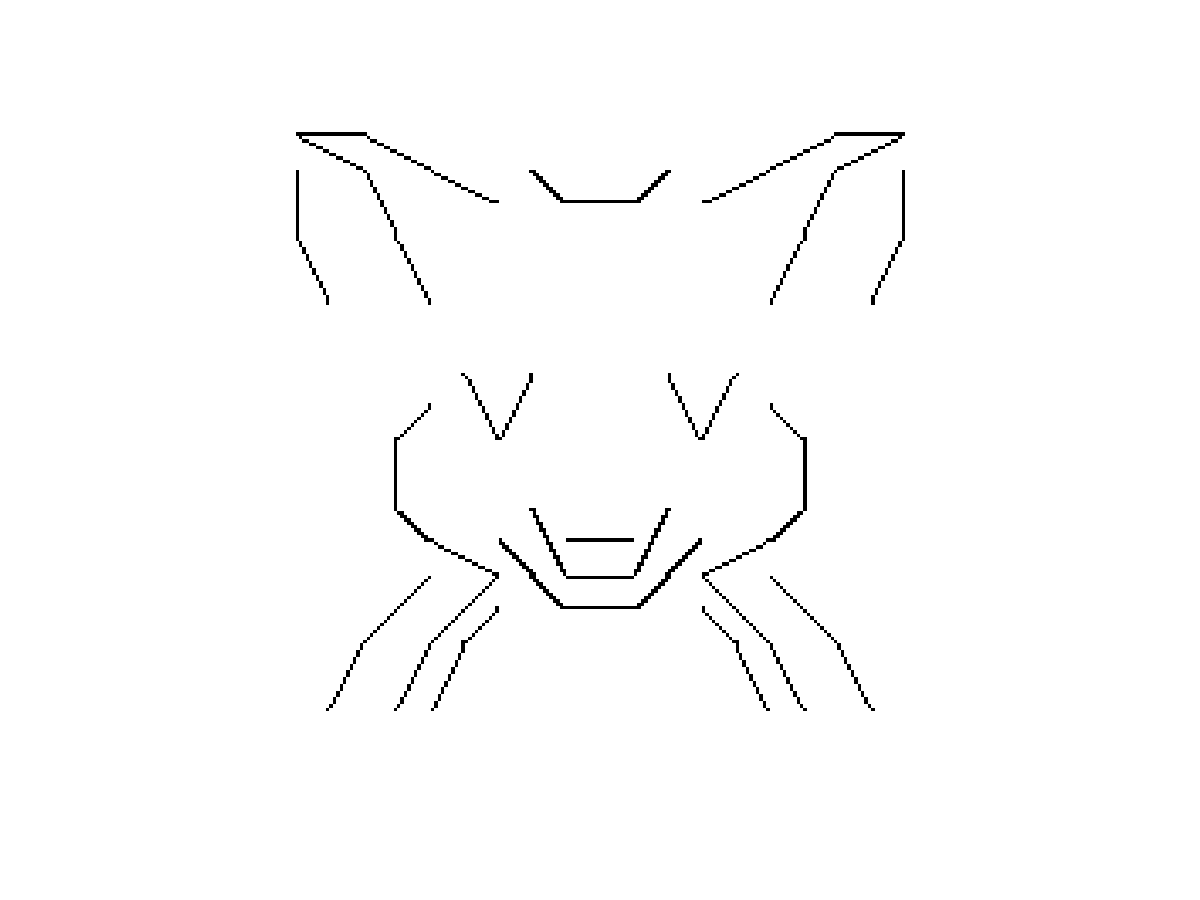}
    }
  \hspace{-4mm}      \subfigure[]{
        \includegraphics[width=0.14\textwidth]{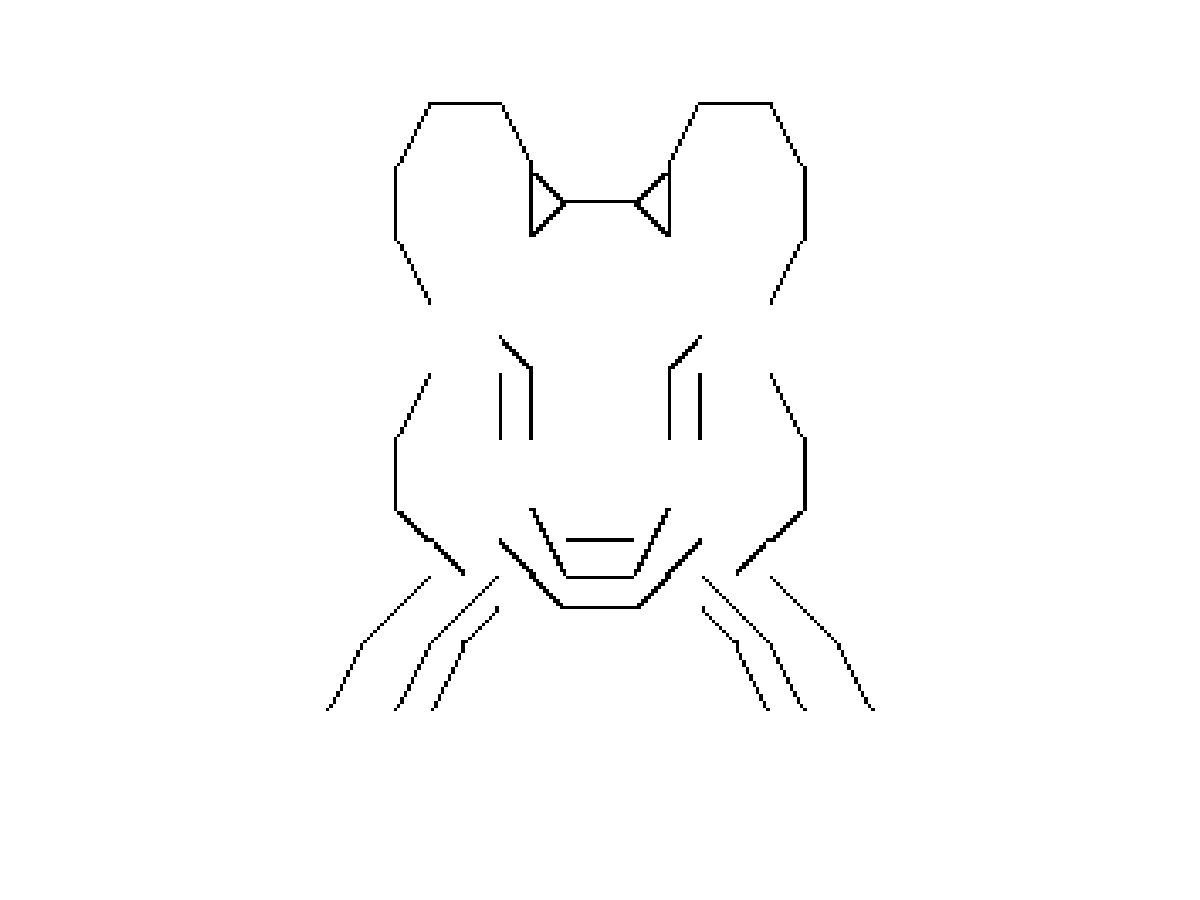}
    }
 \hspace{-4mm}   \subfigure[]{
        \includegraphics[width=0.14\textwidth]{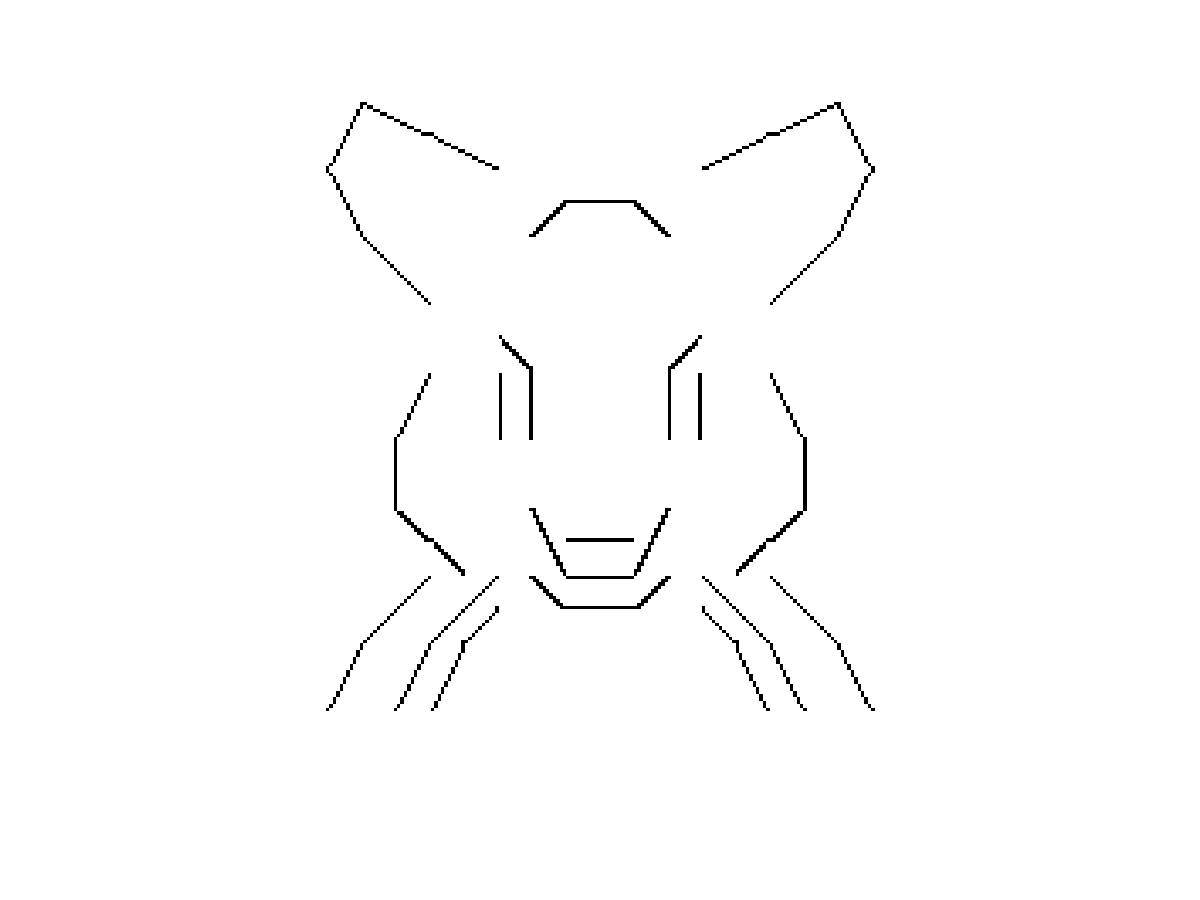}
    }
  \hspace{-4mm}      \subfigure[]{
        \includegraphics[width=0.14\textwidth]{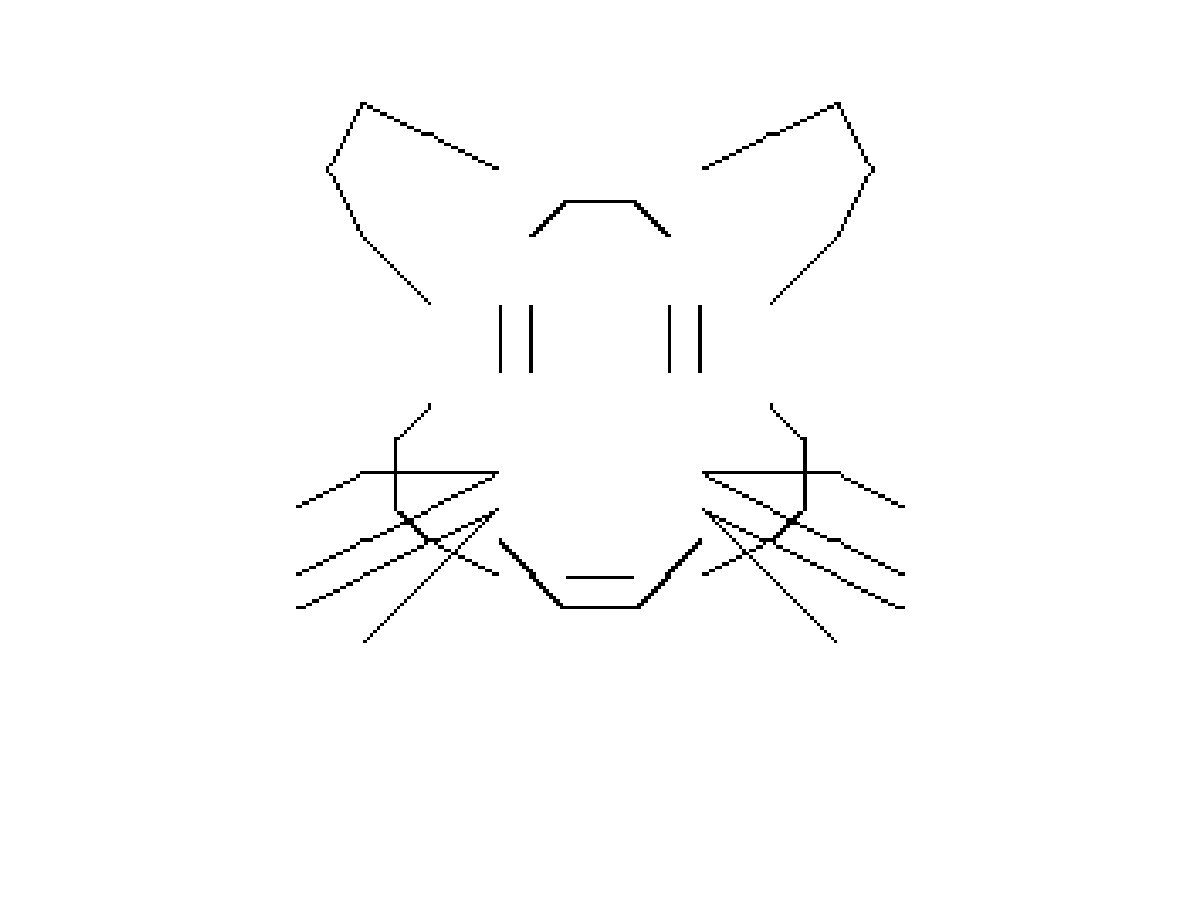}
    }
 \hspace{-4mm}    \subfigure[]{
        \includegraphics[width=0.14\textwidth]{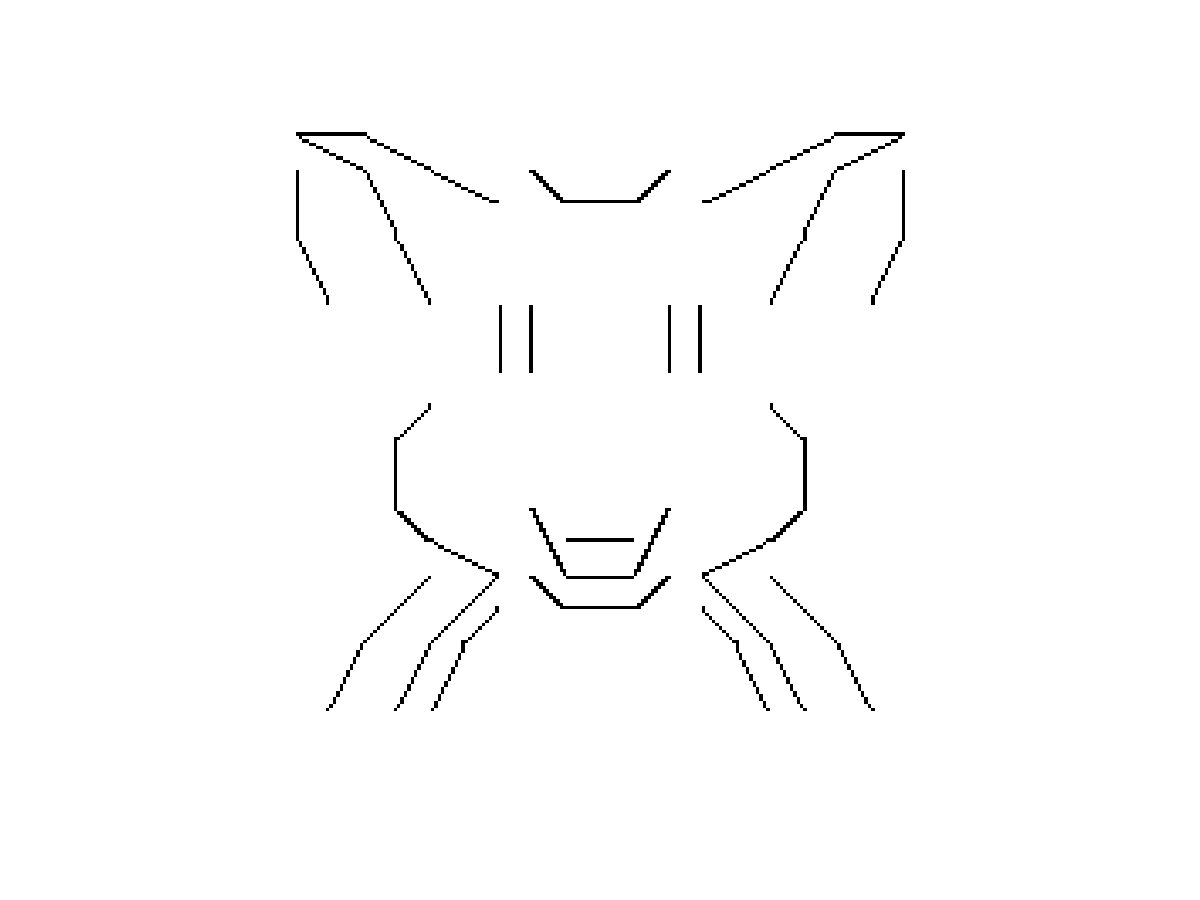}
    }
 \hspace{-4mm}       \subfigure[]{
        \includegraphics[width=0.14\textwidth]{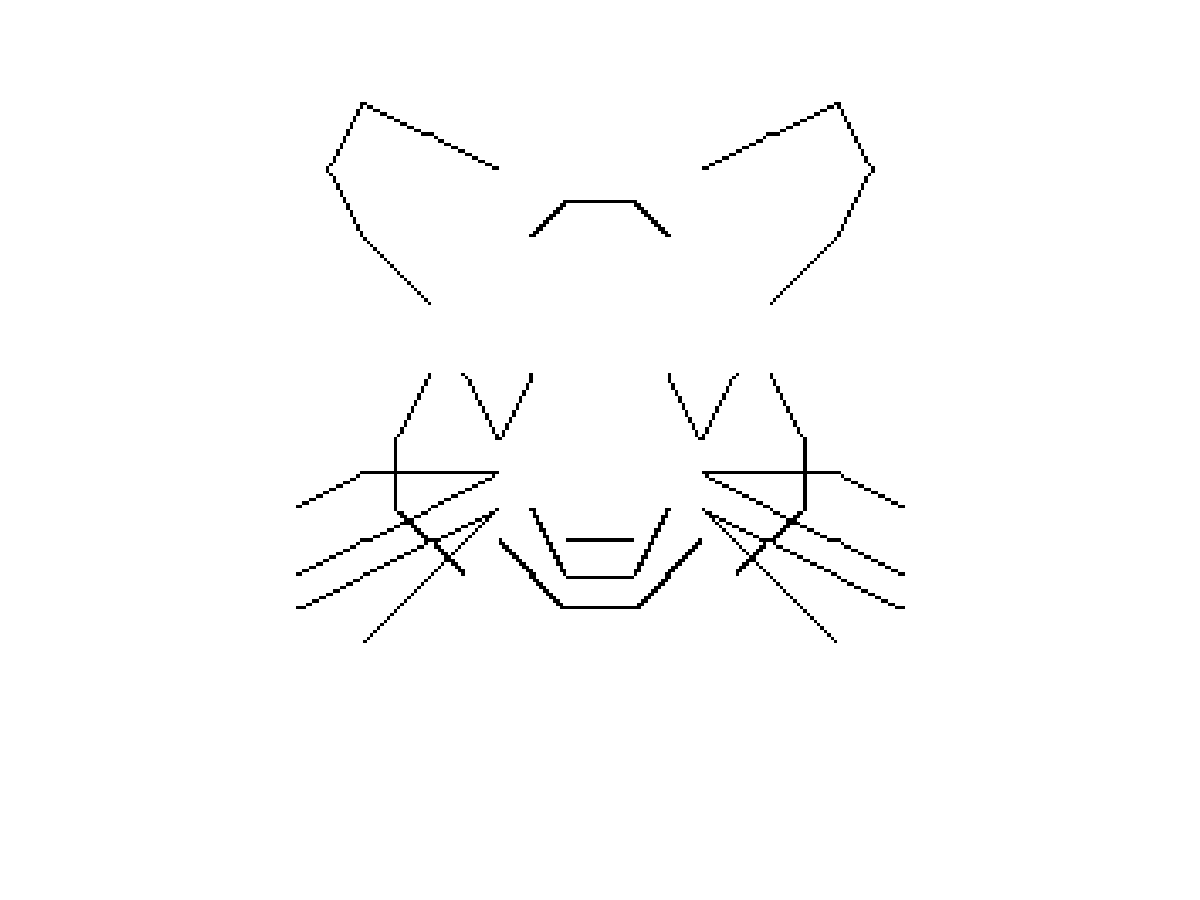}
    } \\
 \hspace{-4mm}    \subfigure[]{
        \includegraphics[width=0.14\textwidth]{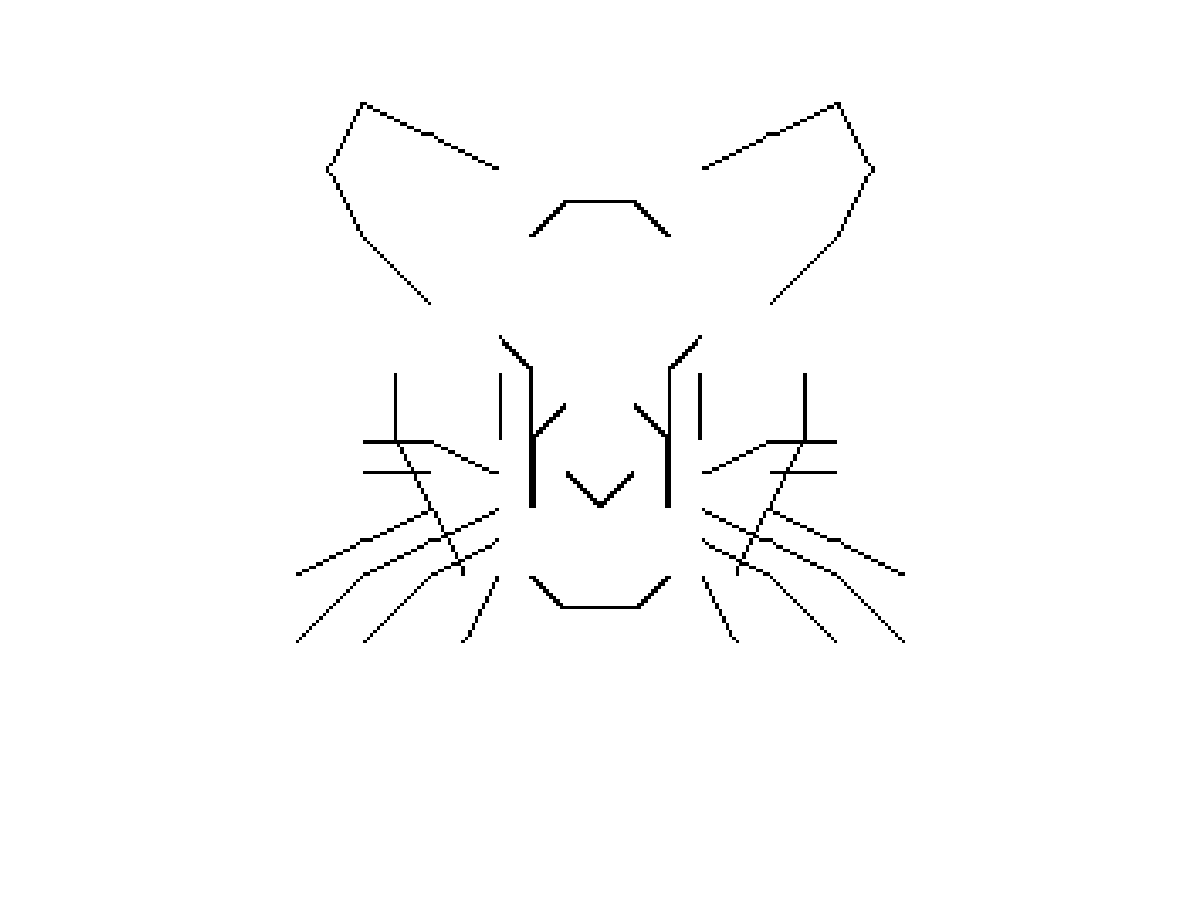}
    } 
 \hspace{-4mm}       \subfigure[]{
        \includegraphics[width=0.14\textwidth]{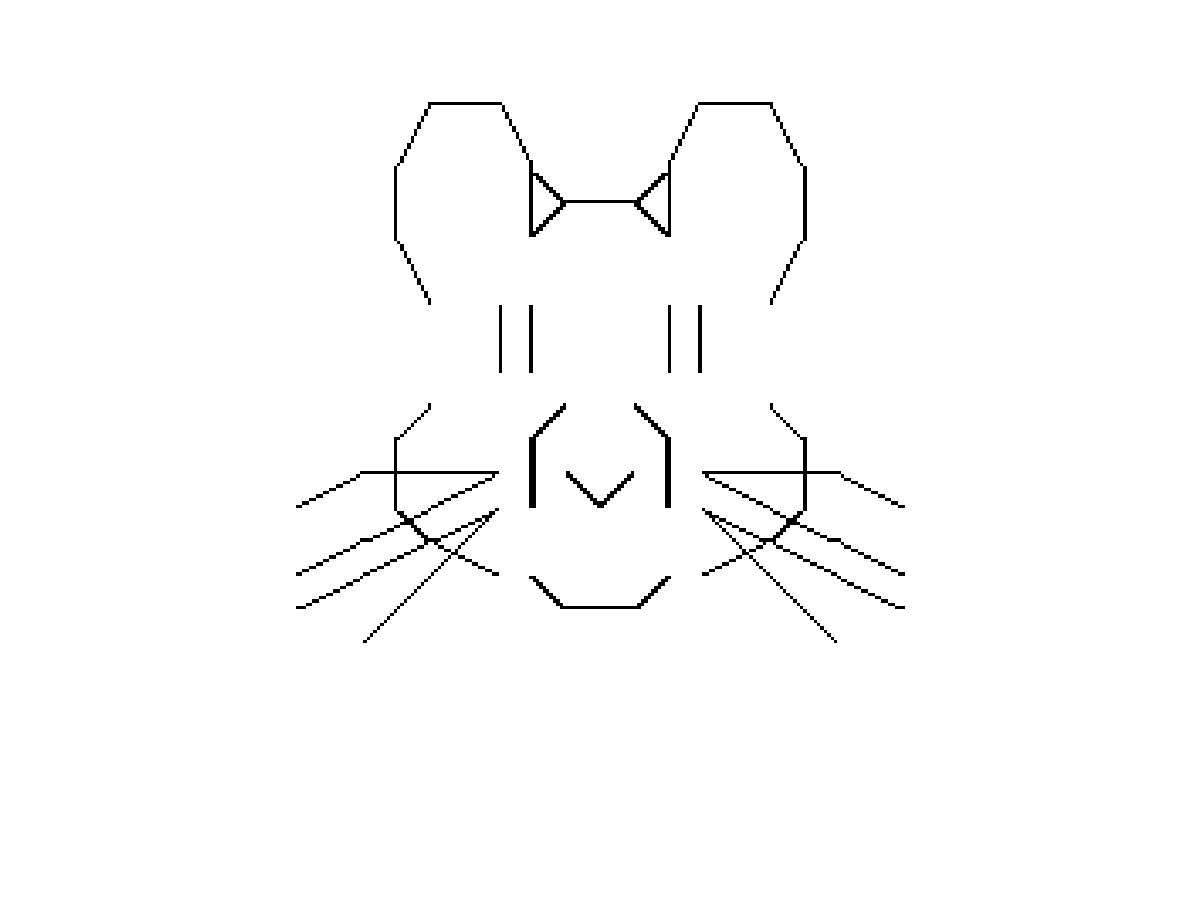}
    }
 \hspace{-4mm}    \subfigure[]{
        \includegraphics[width=0.14\textwidth]{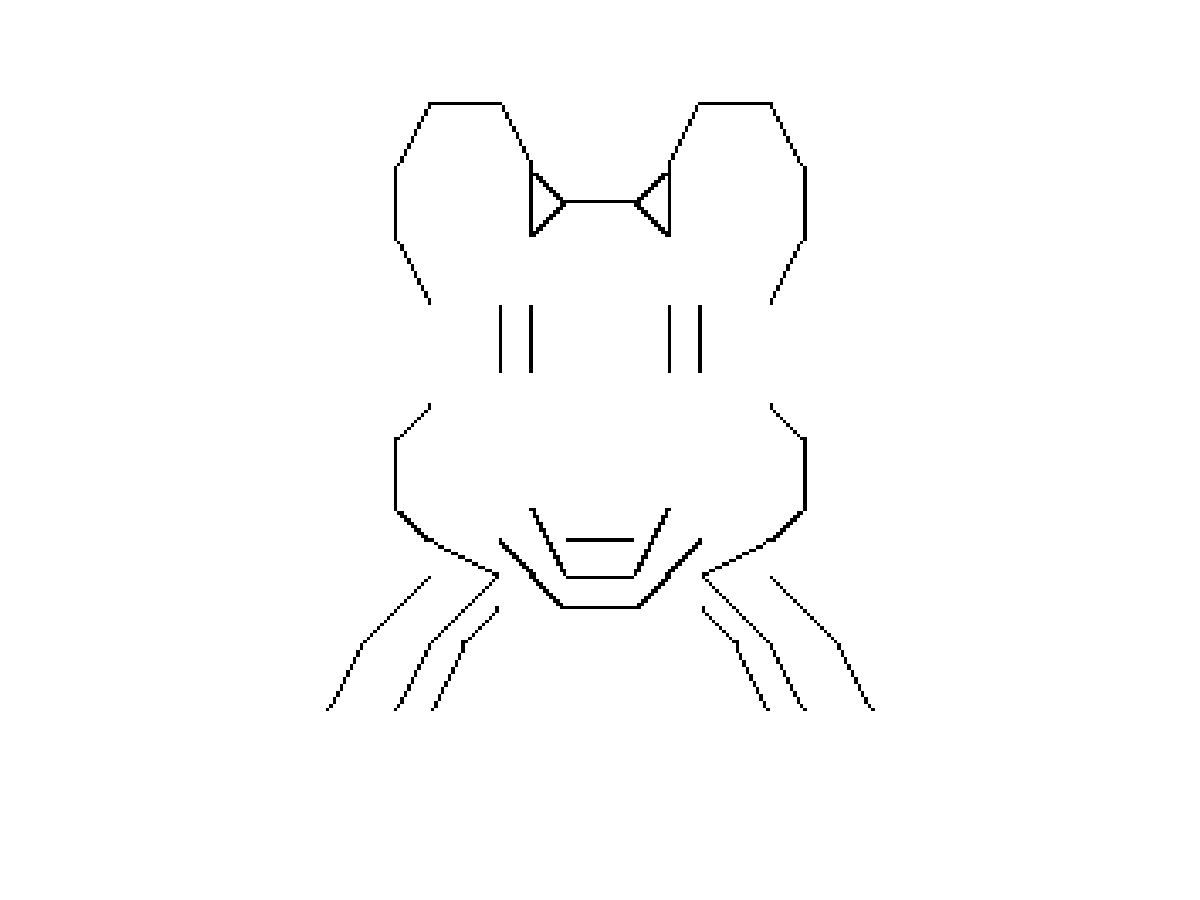}
    }
 \hspace{-4mm}       \subfigure[]{
        \includegraphics[width=0.14\textwidth]{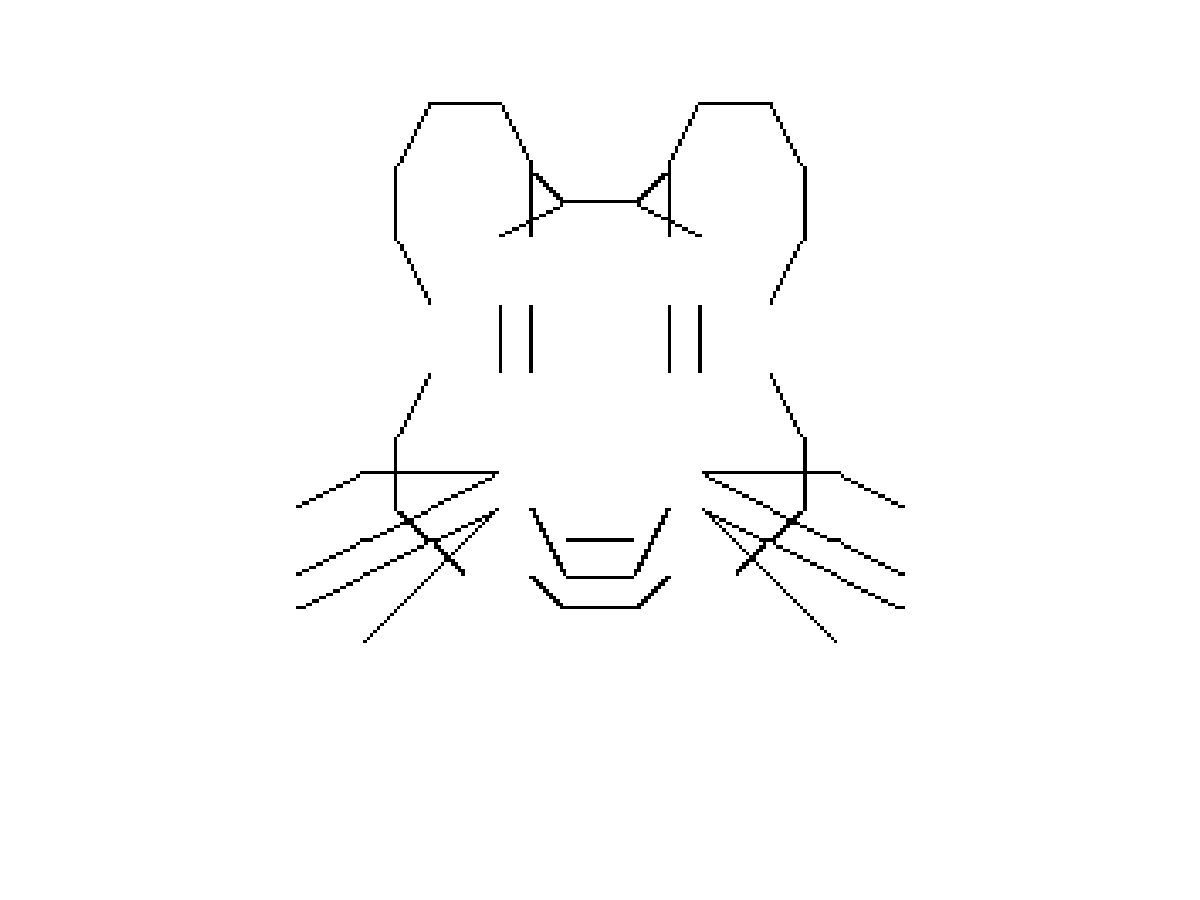}
    }
 \hspace{-4mm}    \subfigure[]{
        \includegraphics[width=0.14\textwidth]{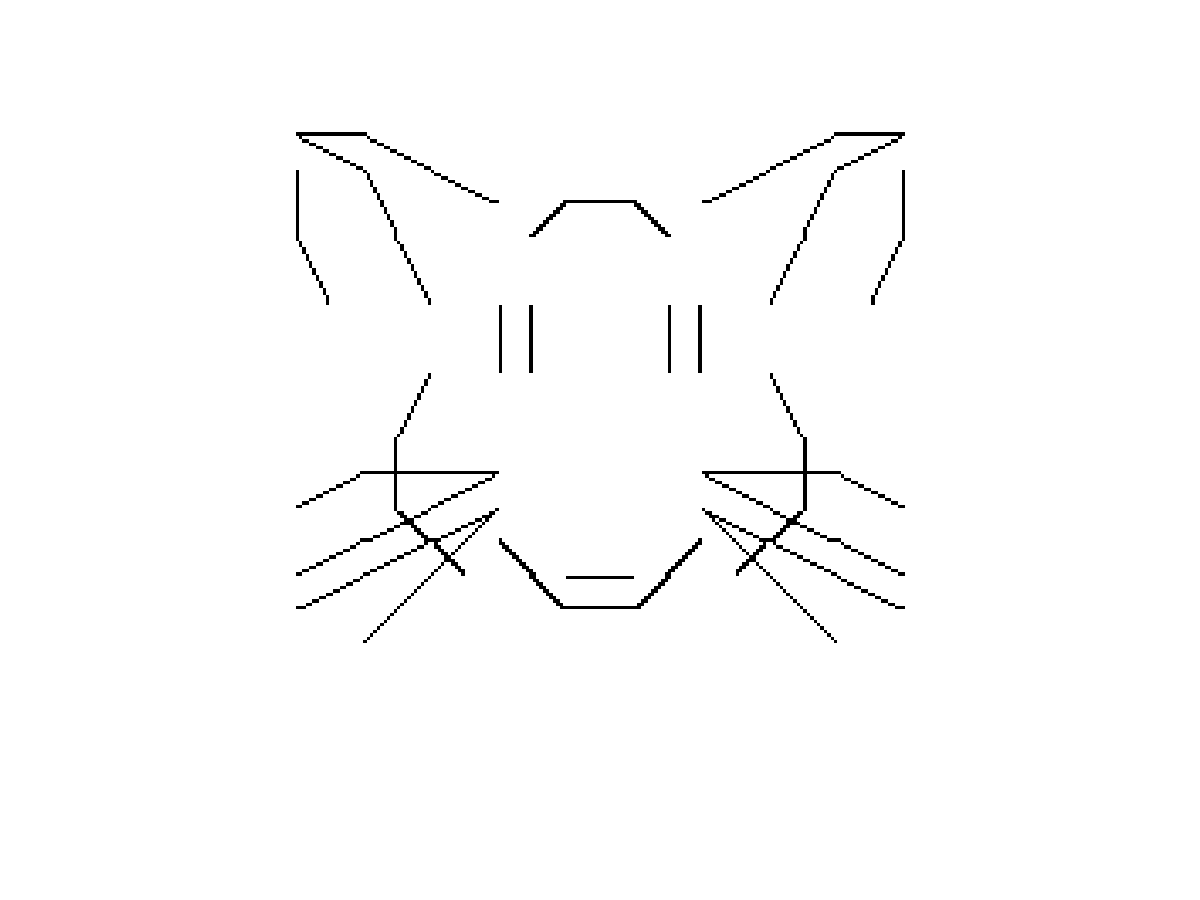}
    }
 \hspace{-4mm}       \subfigure[]{
        \includegraphics[width=0.14\textwidth]{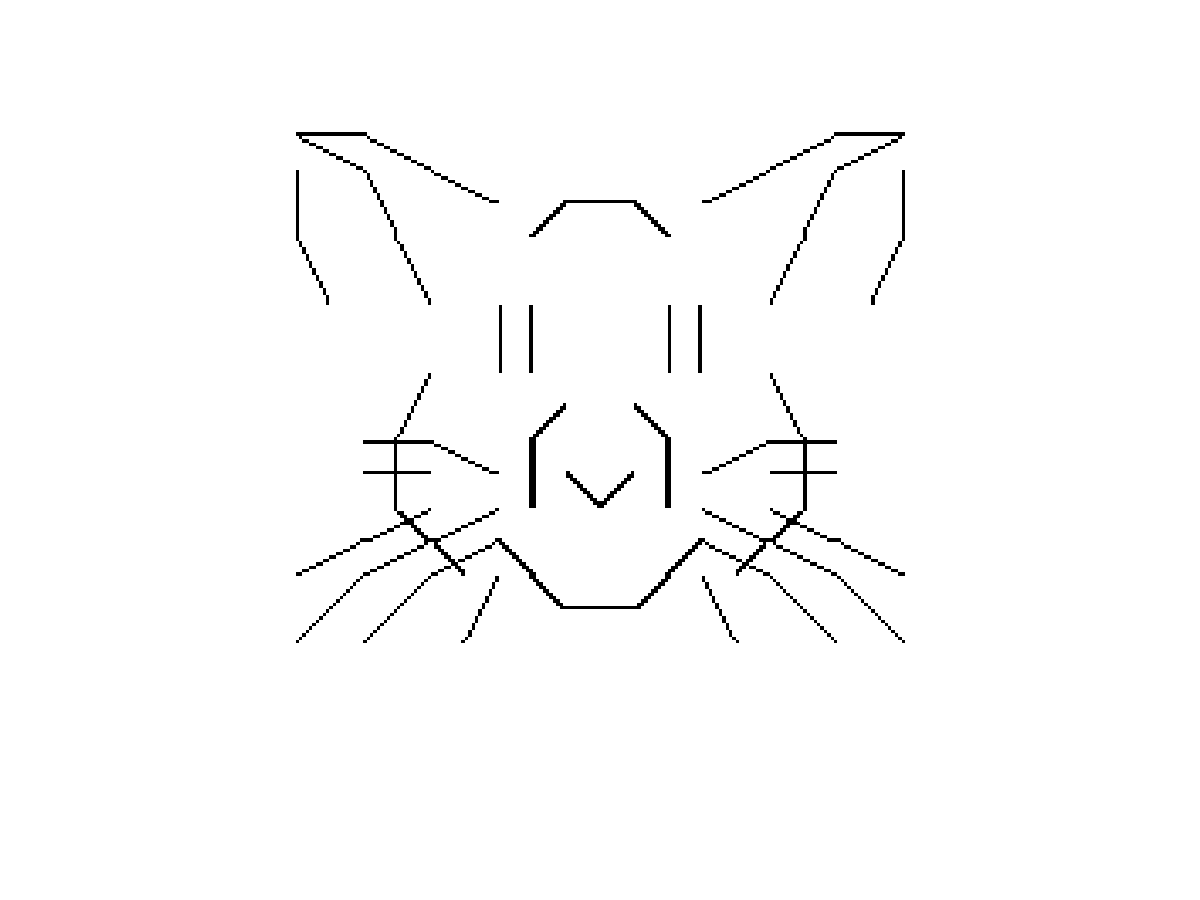}
    }
 \hspace{-4mm}    \subfigure[]{
        \includegraphics[width=0.14\textwidth]{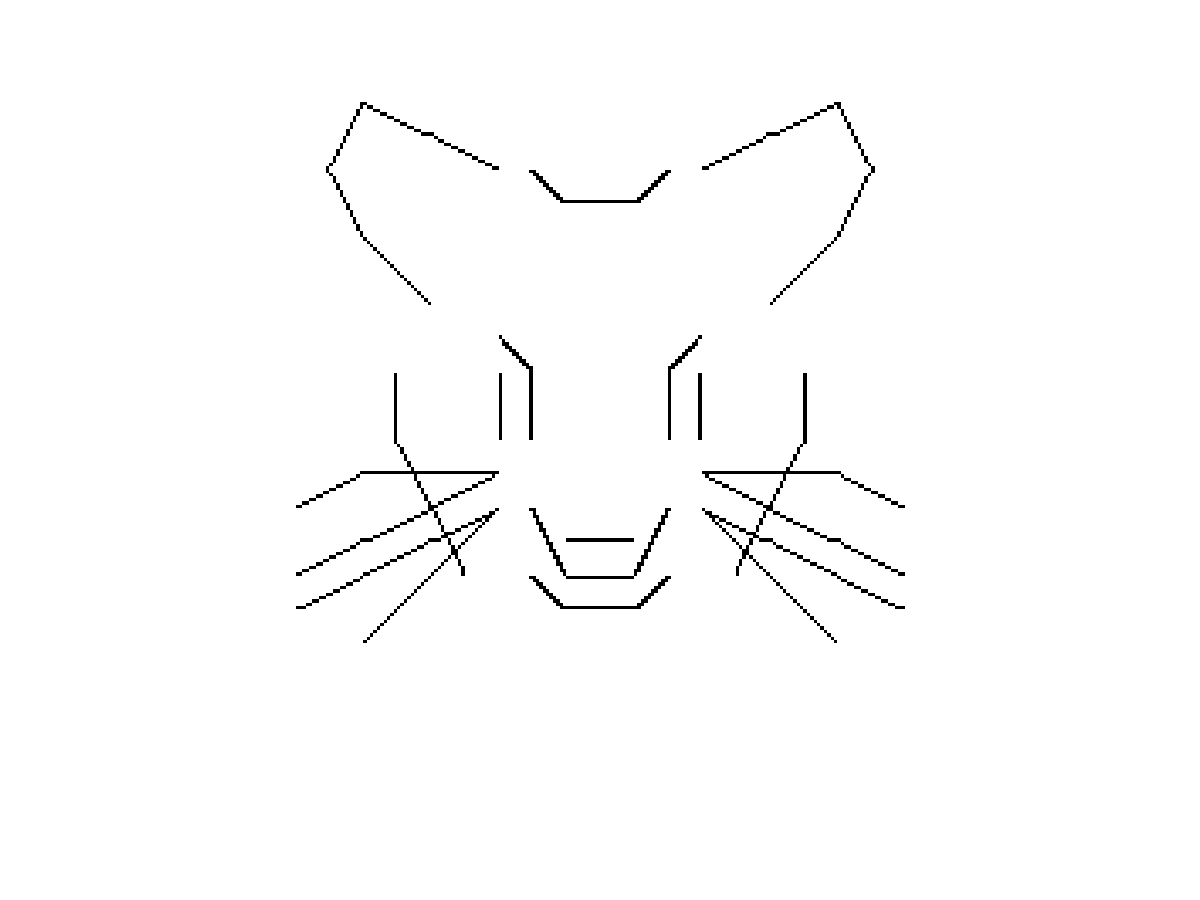}
    } \\
 \hspace{-4mm}       \subfigure[]{
        \includegraphics[width=0.14\textwidth]{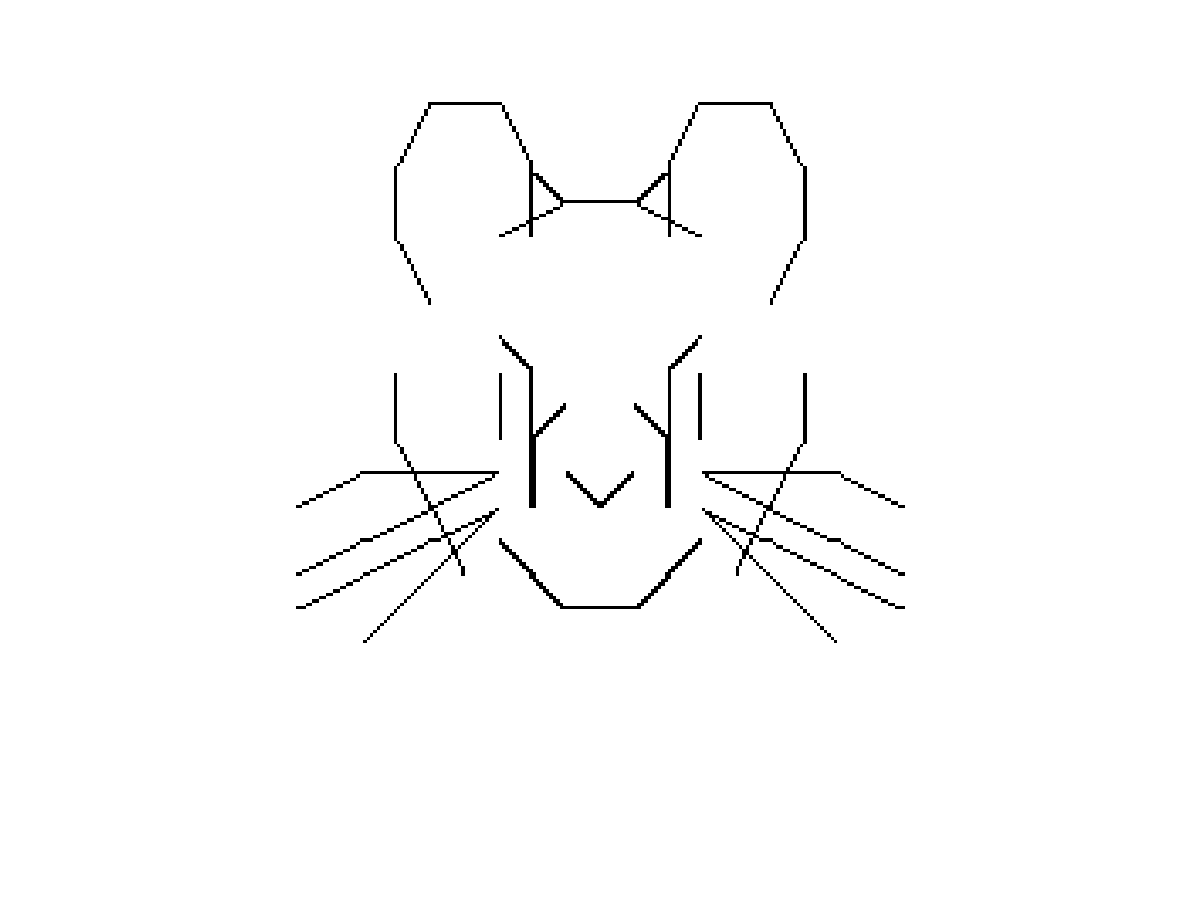}
    }
 \hspace{-4mm}    \subfigure[]{
        \includegraphics[width=0.14\textwidth]{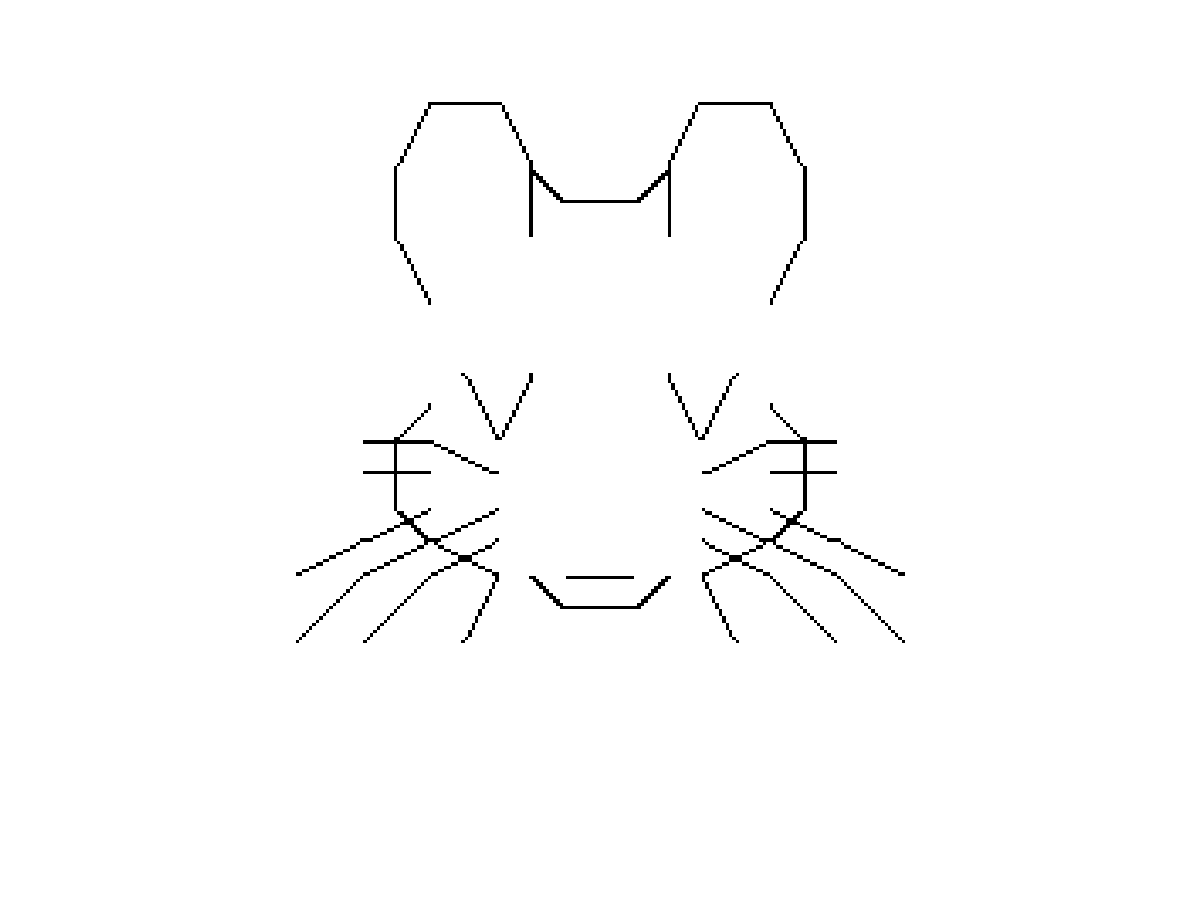}
    }
 \hspace{-4mm}    \subfigure[]{
        \includegraphics[width=0.14\textwidth]{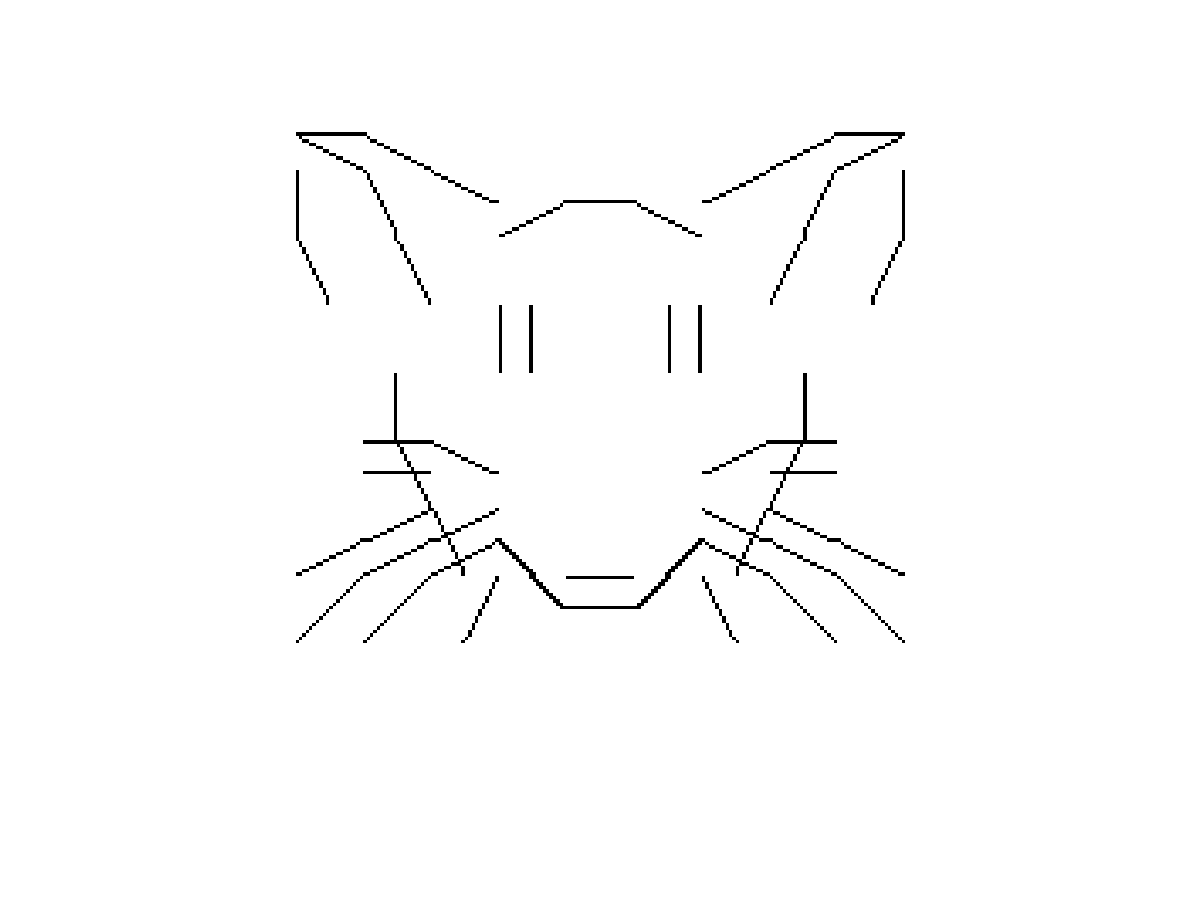}
    }
 \hspace{-4mm}    \subfigure[]{
        \includegraphics[width=0.14\textwidth]{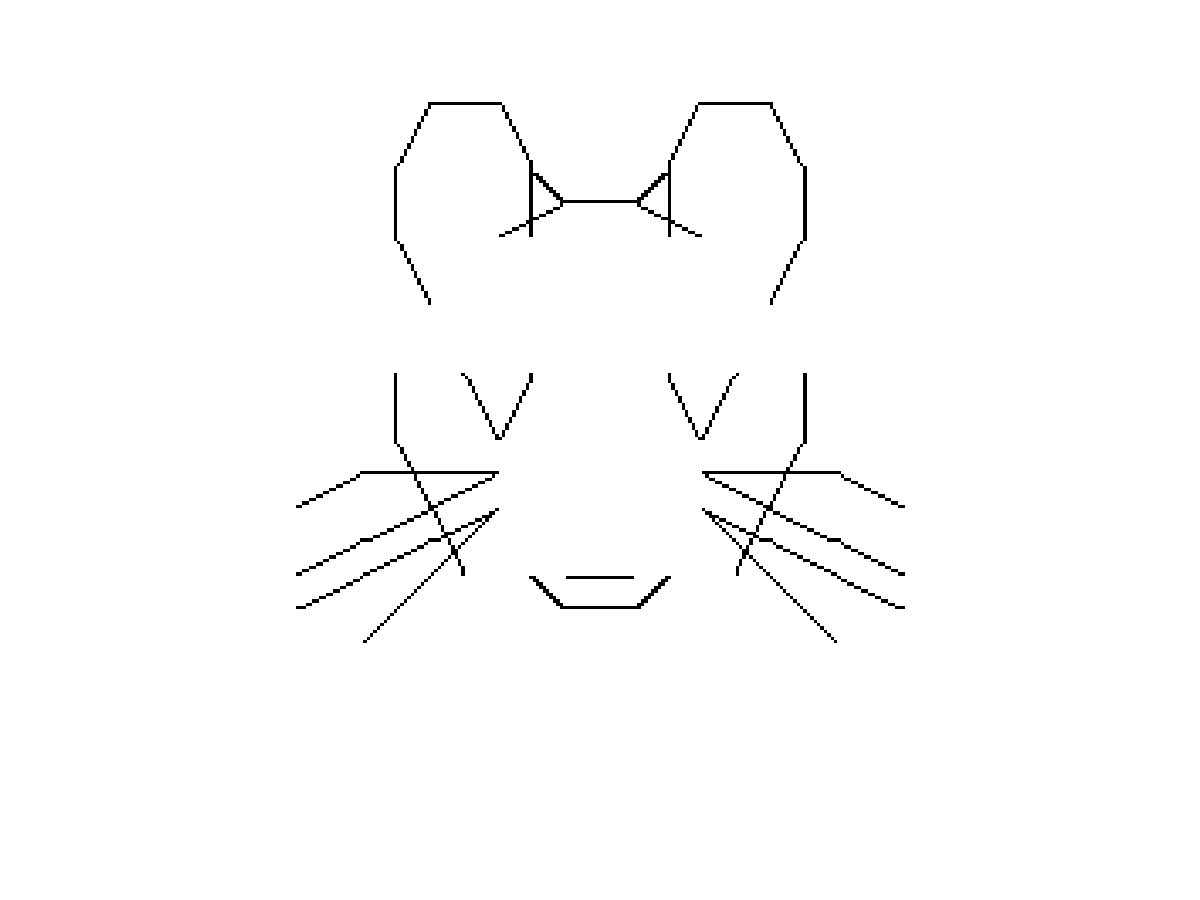}
    }
  \hspace{-4mm}  \subfigure[]{
        \includegraphics[width=0.14\textwidth]{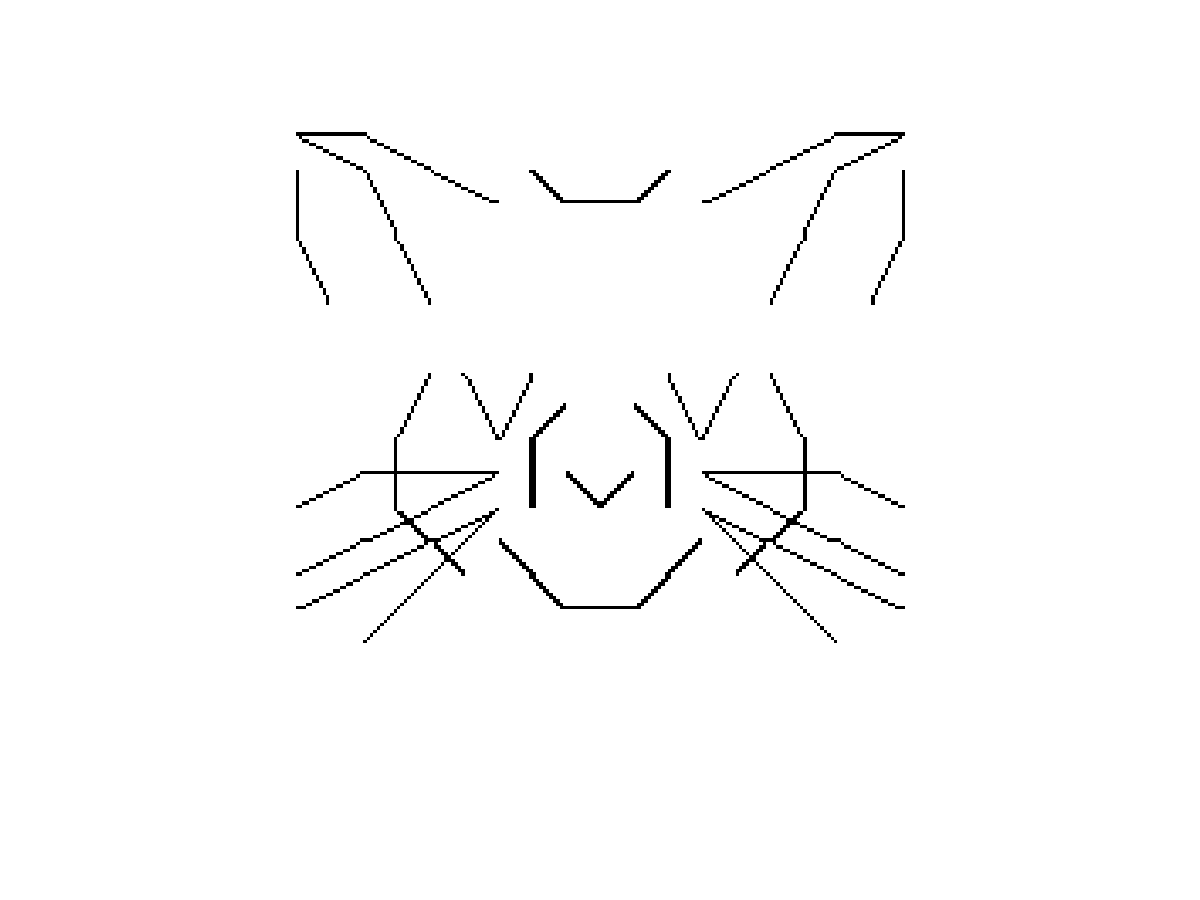}
    }
 \hspace{-4mm}  \subfigure[]{
        \includegraphics[width=0.14\textwidth]{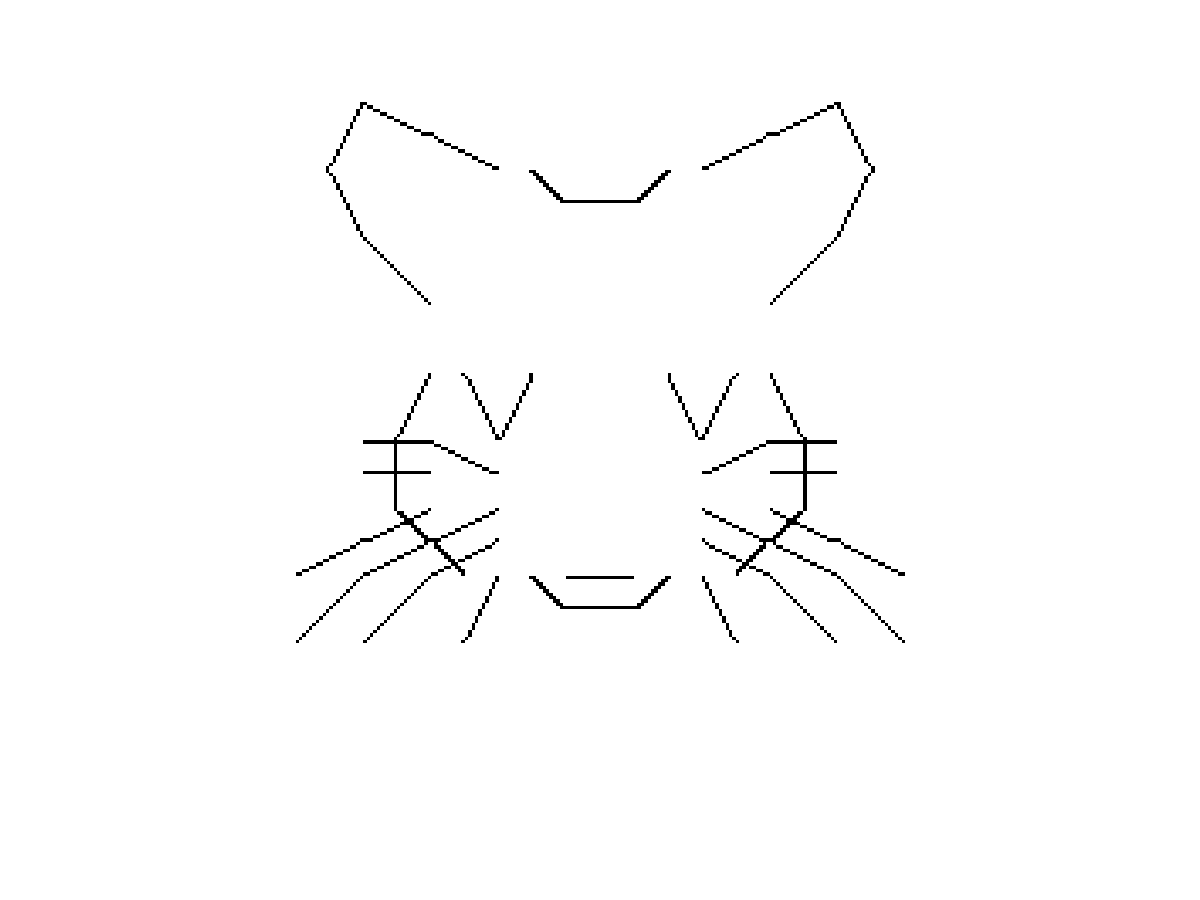}
    }
       \caption{Bernoulli templates for mouse faces with high overlap and low separability.} \label{fig:full_faces_overlap} 
\end{figure*}

\begin{figure}
    \center
        \includegraphics[width=0.7 \columnwidth]{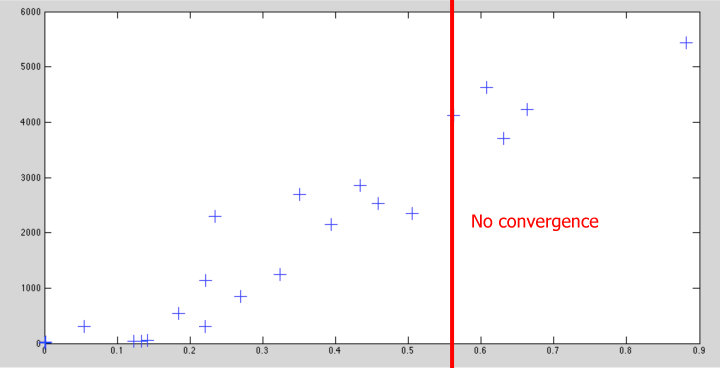}
        \caption{Number of local minima found for varying degrees of overlap in the Bernoulli templates. Each marker corresponds to a Bernoulli mixture model that consists of three of the 20 Bernoullie templates.} \label{fig:full_faces_convergence} 
\end{figure}

We repeat the same experiment using variants of a mouse face. 
We swap out each component of the mouse face (the eyes, ears, whiskers, nose, mouth, head top and head sides) with three different variants. We thereby generate $20$ Bernouilli templates in Figure~\ref{fig:full_faces_overlap}, which have relatively high degrees of overlap.
We generated the ELMs of various Bernouilli mixture models containing three of the $20$ templates and noise level $p=0$. In each Bernouilli mixture model, the three templates have different degrees of overlap.  
Hence we plot the number of local minima in the ELMs versus the degree of overlap as show in Figure \ref{fig:full_faces_convergence}. As expected, the number of local minima increases with the degree of overlap, and there are too many local minima for the algorithm to converge past overlap $c=0.5$.

\subsection{Experiments on Real Data}

\begin{figure}[t]
 \begin{minipage}[b]{0.45\linewidth}
     \center
 \includegraphics[width=0.7\columnwidth]{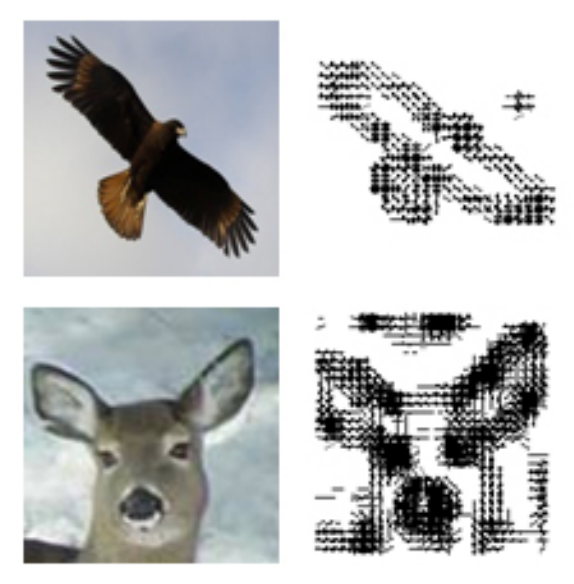}
    \caption{Animal face images and corresponding binary sketches indicates the existence of a Gabor filter response above a fixed threshold.
} \label{fig:gabor} 
 \end{minipage}
 \hspace{2em}
 \begin{minipage}[b]{0.45\linewidth}
    \center
 \includegraphics[trim=0cm 9cm 17cm 0cm, clip=true, width=1 \columnwidth]{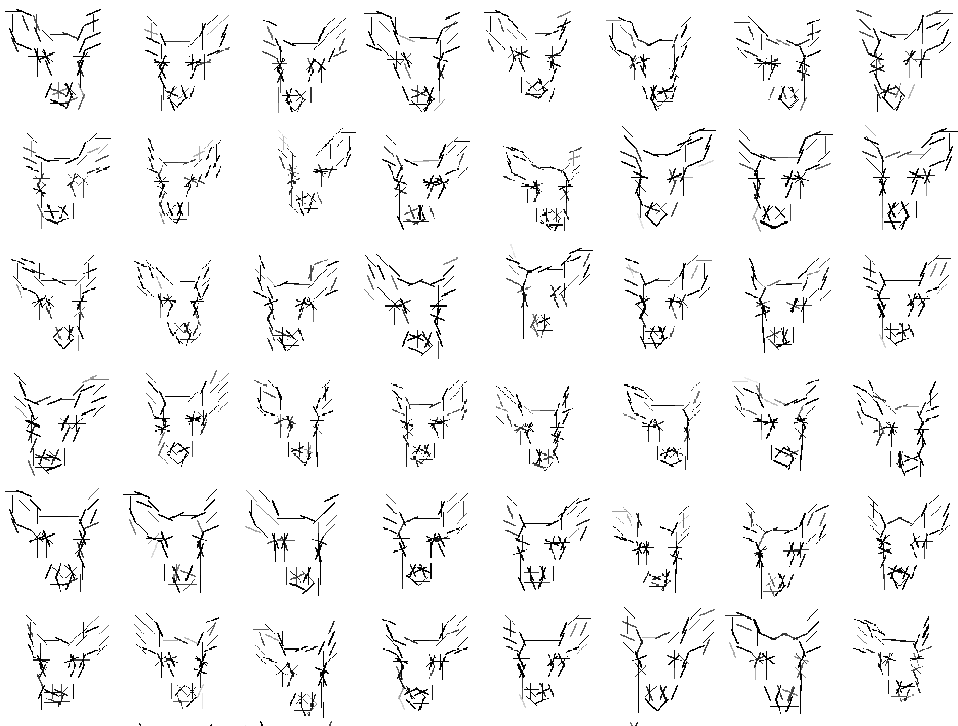}
    \caption{Deer face sketches binarized from real images.}
 \label{fig:deer-sketches} 
 \end{minipage}
\end{figure}

We perform the Bernouilli templates experiment on a set of real images of animal faces. We binarize the images by extracting the prominent sketches on a 9x9 grid. Eight Gabor filters with eight different orientations centered in the centers and corners of each cell are applied to the image. The filters with a strong response above a fixed threshold correspond to edges detected in the image; these are mapped to the dictionary of $18$ elements. Thus each animal face is represented as a $18\times9\times9$ dimensional binary vector. The Gabor filter responses on animal face pictures are shown in Figure \ref{fig:gabor}. The binarized animal faces are shown in Figure \ref{fig:deer-sketches}.

\begin{figure}
    \center
 \includegraphics[width=0.9 \columnwidth]{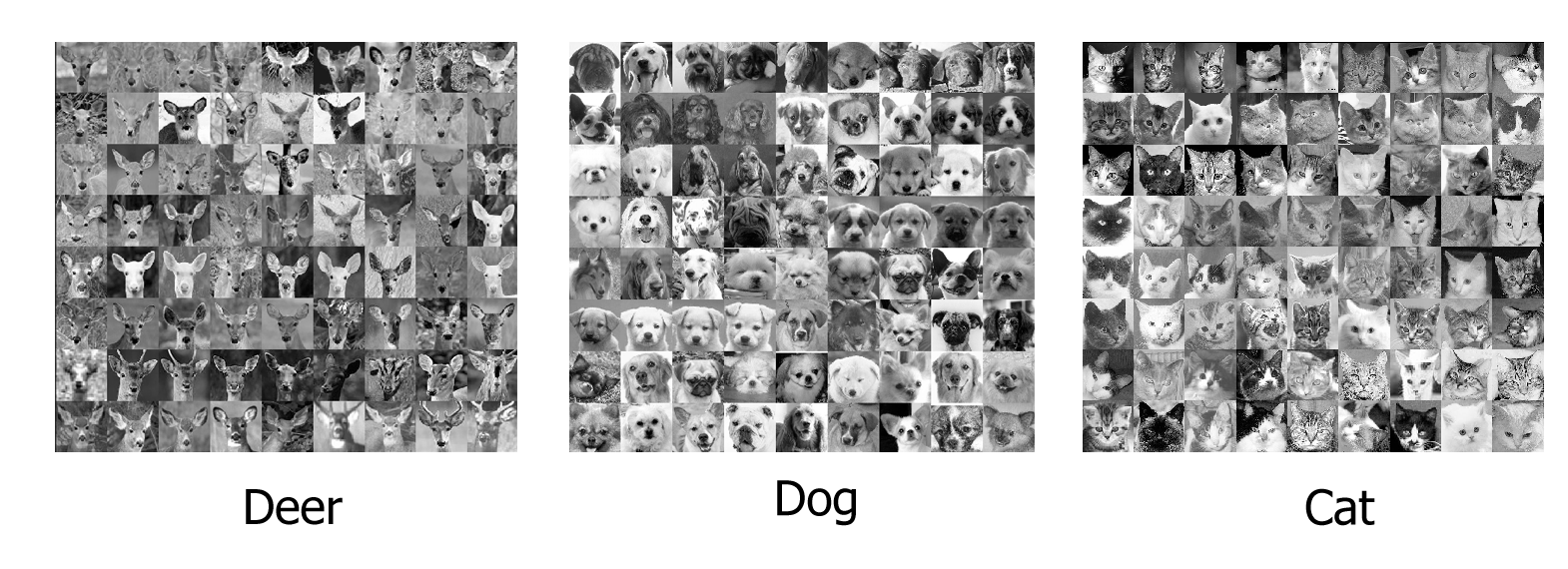}
    \caption{Animal face images of three categories.}
 \label{fig:all-faces} 

\end{figure}

We chose 3 different animal types -- deer, cat and mouse, with an equal number of images chosen from each category (Figure \ref{fig:all-faces}). The binarized versions of these images can be modeled as a mixture of 3 Bernouilli templates - each template corresponding to one animal face type.

\begin{figure}
    \center
 \includegraphics[width= 0.5\columnwidth]{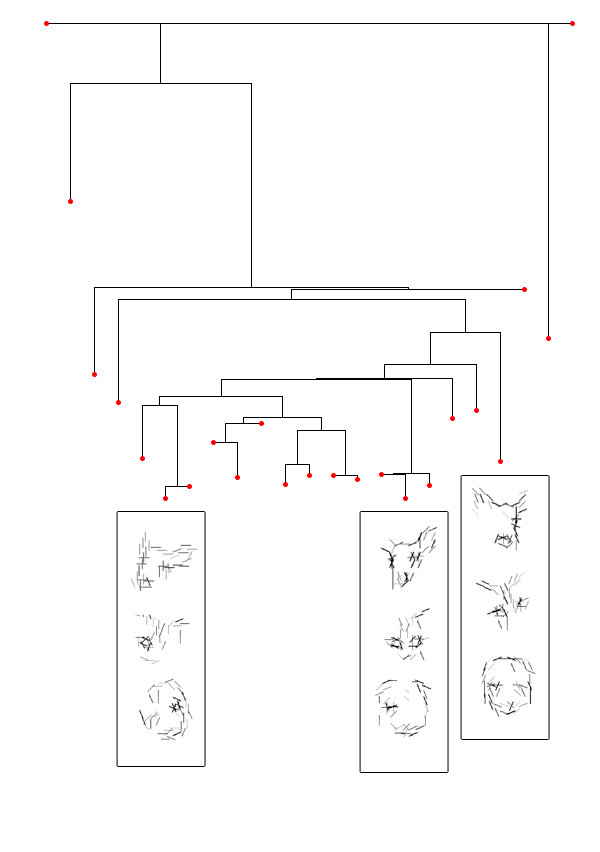}
    \caption{ELM of the three animal faces dataset (dog, cat, and deer). We show the Bernouilli templates corresponding to three local minima with large energy barriers. }
 \label{fig:all-faces-ELM} 
\end{figure}

The ELM is shown in Figure \ref{fig:all-faces-ELM} along with the Bernouilli templates corresponding to three local minima separated by large energy barriers. We make two observations: 1. The templates corresponding to each animal type are clearly identifiable, and therefore the algorithm has converged on reasonable local minima. 2. The animal faces have differing orientations across the local minima (the deer face on in the left-most local minimum is rotated and tilted to the right and the dog face in the same local minimum is rotated and lilted to the left), which explains the energy barriers between them. 

\begin{figure}
    \center
    \subfigure[SW-cut] {
        \includegraphics[width=0.3\columnwidth]{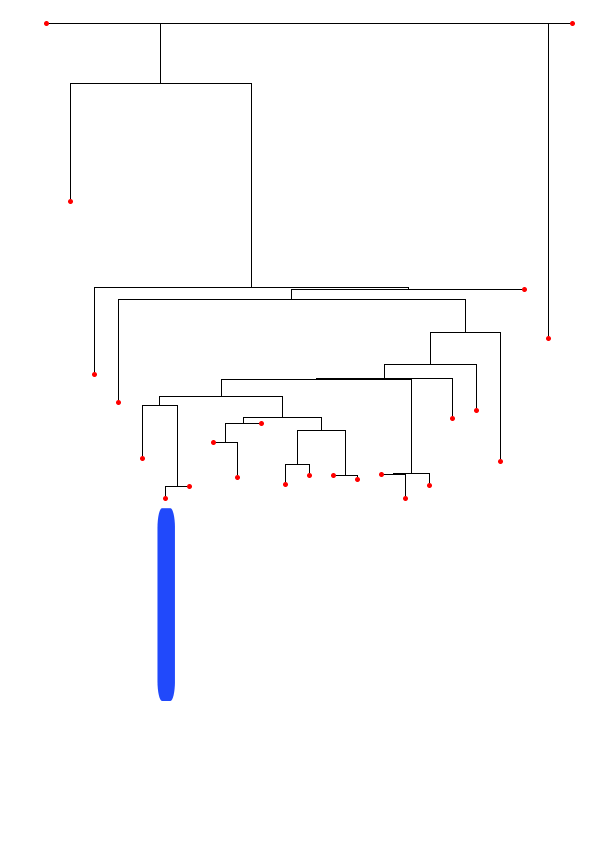}
    }
    \subfigure[EM ]{
        \includegraphics[width=0.3\columnwidth]{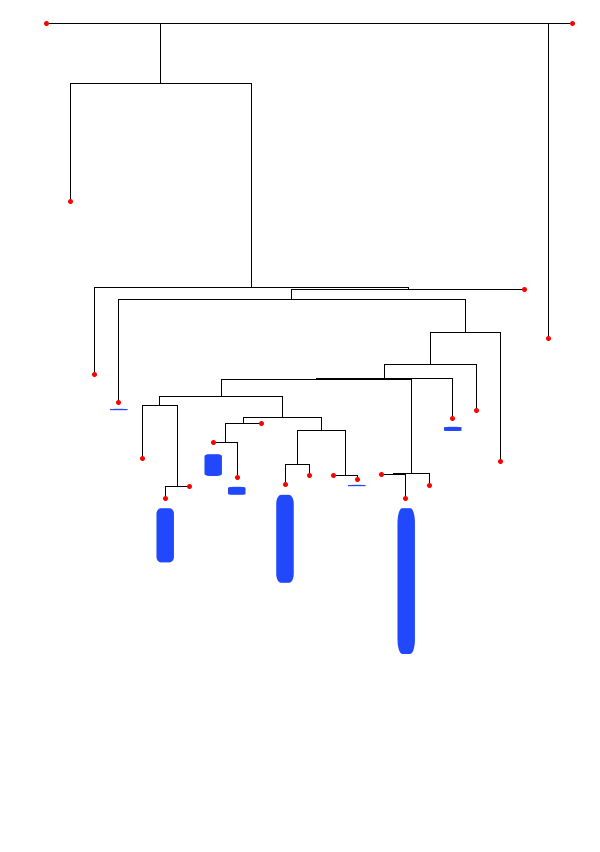}
    }
     \subfigure[k-means ]{
        \includegraphics[width=0.3\columnwidth]{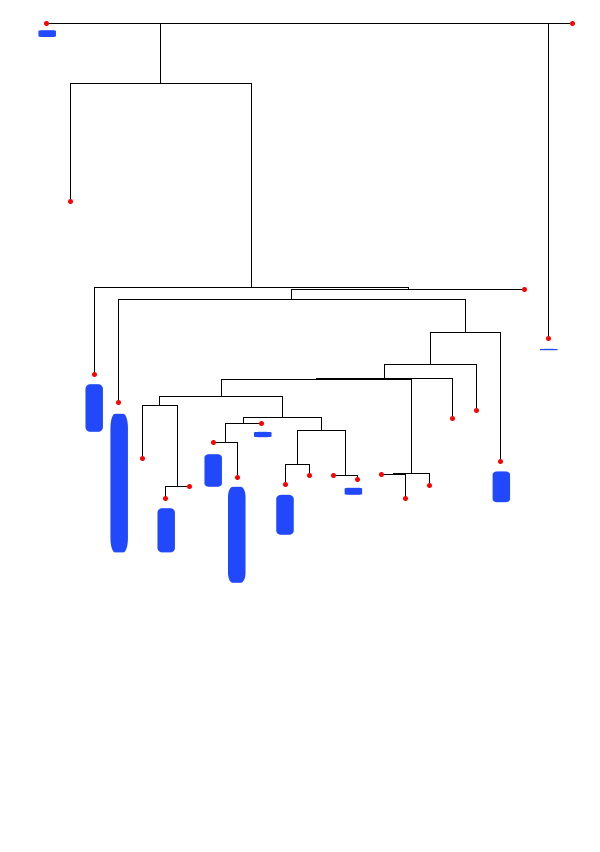}
    }
    \caption{ Comparison of SW-cut, k-means, and EM algorithm performance on the ELM of animal face Bernouilli mixture model.} \label{fig:algs-comp-animals}
\end{figure}

Figure \ref{fig:algs-comp-animals} shows a comparison of the SW-cut, k-means, and EM algorithm performance as a histogram on the ELM of animal face Bernouilli mixture model. The histogram is obtained by running each algorithm 200 times with a random initialization, then finding the closest local minimum in the ELM to the output of the algorithm. The counts of the closest local minima are then displayed as a bar plot next to each local minimum. 
It can be seen that SW-cut always finds the global minimum, while k-means performs the worst probably because of the high degree of overlap between the sketches of the three types of animal faces.

\section{Experiment III: ELM of bi-clustering}\label{sec:bc}

bi-clustering is a learning process (see a survey by \cite{Madeira04survey}) which has been widely used in bioinformatics, e.g., finding genes with similar expression patterns under subsets of conditions (\cite{Cheng00, Getz00, Cho04}). It is also used in data mining, e.g., finding people who enjoy similar movies (\cite{Yang02}), and in learning language models by finding co-occurring words and phrases in grammar rules (\cite{Tu08}). 

\begin{figure*}
    \center
        \includegraphics[width=\textwidth]{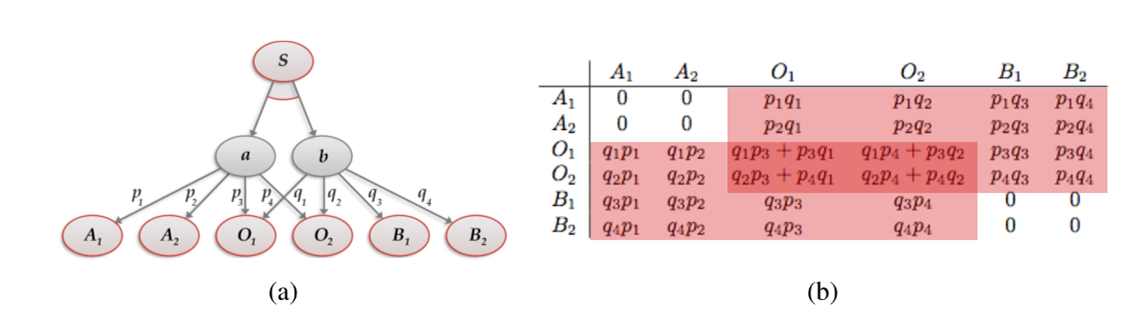}
    \caption{ (a) A bi-clustering model. (b) The co-occurrence matrix with the theoretical frequencies of its elements. } \label{fig:bi-cluster} 
\end{figure*}

Figure~\ref{fig:bi-cluster}.(a) shows a bi-clustering model (with multiplicative coherence) in the form of a three layer And-Or graph. The underlying pattern $S$ has two conjunction factors $a$ and $b$. $a$ can choose from a number of alternative elements $A_1, A_2, O_1, O_2$ at probability $p_1, ..., p_4$ respectively. Similarly, $b$ can choose from elements $O_1, O_2, B_1, B_2$ with probability $q_1, ..., q_4$ respectively.
It can be seen that $a$ and $b$ have shared elements $O_1, O_2$.
For comparison, we note that the clustering models in experiments I and II can be seen as three-layer Or-And graphs with a mixture (Or-node) on the top and each component is a conjunction of multiple variables.

From data sampled from this model, one can compute a co-occurrence matrix for the two elements chosen by $a$ and $b$, and the theoretical co-occurring frequency is shown in Figure~\ref{fig:bi-cluster}.(b). When only a small number of observations are available, this matrix may have significant fluctuations. There may also be unwanted background elements in the matrix. The goal of bi-clustering is to identify the bi-cluster (one of the two submatrixes in Figure~\ref{fig:bi-cluster}.(b)) from the noisy matrix. Note that this is a simple special case of the bi-clustering problem and in general the matrix may contain many bi-clusters that are not necessarily symmetrical.
 
We denote the bi-cluster to be identified by $\Theta = \langle A,B \rangle$ where $A$ is the set of rows and $B$ is the set of columns of the bi-cluster. Note that the goal of bi-clustering is not to explain all the data but to identify a subset of the data that exhibit certain properties (e.g., coherence). Therefore, instead of using likelihood or posterior probability to define the energy function, we use the following energy function adapted from \cite{Tu08}.
\begin{align*}
\label{eqn:bi-clusterLG}
E(\Theta) = &\left( s\log s + \sum_{x\in A, y\in B}a_{x,y} \log a_{x,y} - \sum_{x\in A} r_x \log r_x  - \sum_{y\in B} c_y\log c_y\right) \\
& - \alpha \left(2\sum_{x\in A, y\in B}a_{x,y} - |A| - |B|\right).
\end{align*}
In the above formula, $a_{x,y}$ is the element at row $x$ and column $y$, $r_x$ is the sum of row $x$, $c_y$ is the sum of column $y$, and $s$ is the total sum of the bi-cluster.
The first term in the energy function measures the coherence of the bi-cluster, which reaches its minimal value of 0 if the bi-cluster is perfectly multiplicatively coherent (i.e., the elements are perfectly proportional).
The second term corresponds to the prior, which favors bi-clusters that cover more data; the $-|A|-|B|$ term is added to exclude rows and columns that are entirely zero from the bi-cluster.

We experimented with synthetic bi-clustering models in which $a$ and $b$ each have $10$ alternative elements. We varied the following factors to generate a set of different models: (i) the levels of overlaps between $a$ and $b$: $0, 1, ..., 10$; and (ii) random background noises at level $p$. We generated $1000$ data points from each model and constructed the matrix.  For each data matrix, we ran our algorithm to plot the ELMs with different values of $\alpha$, the strength of the prior.

\begin{figure*}
    \center
    \subfigure[] {
        \includegraphics[width=\textwidth]{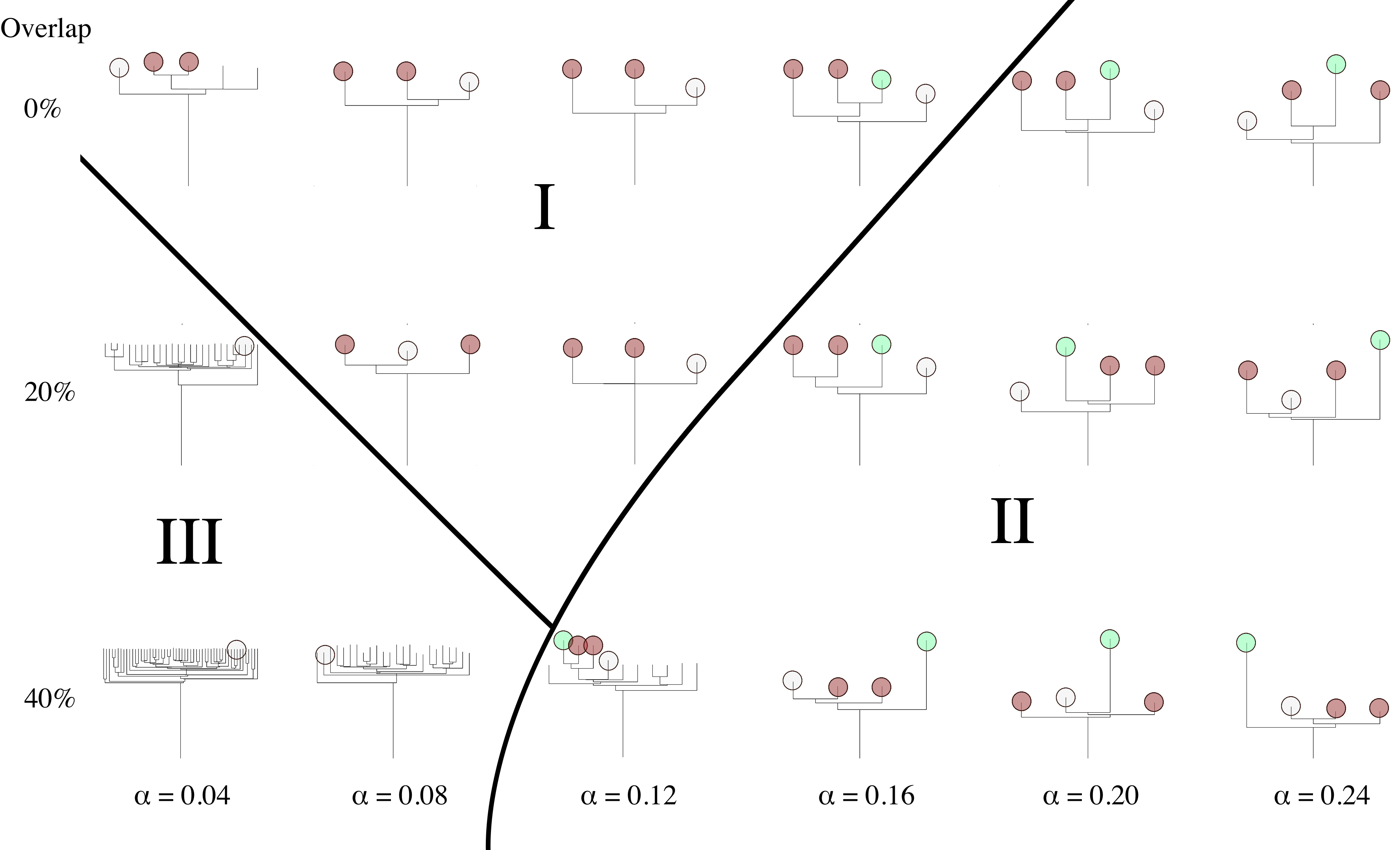}
    } 
    \caption{ Energy Landscape Maps for learning two bi-clusters with 0\%,20\%, 40\% overlap and  hyperparameter $\alpha$.  Red:  correct bi-cluster; Grey: empty bi-cluster; Green: maximal bi-cluster. } \label{fig:maps1} 
\end{figure*}

Figure~\ref{fig:maps1} shows some of the ELMs with the overlap being $0\%, 20\%, 40\%$ respectively, the prior strength being $\alpha = 0.02, 0.06, \dots, 0.24$, and the noise level $p=0.00$. 
The local maxima corresponding to the correct bi-clusters (either the target bi-cluster or its transposition) are marked with solid red circles; the empty bi-cluster is marked with a gray circle; and the maximal bi-cluster containing the whole data matrix is marked with a solid green circle. 


 These ELMs can be divided into three regimes. 
\begin{itemize}
\item Regime I: the true model is easily learnable; the global maxima correspond to the correct bi-clusters and there are fewer than 6 local minima.

\item Regime II: the prior is too strong, the ELM has a dominating minimum which is the maximal bi-cluster. Thus the model is biased and cannot recover the underlying bi-cluster. 

\item Regime III: the prior is too weak, resulting in too many local minima at similar energy levels. The true model may not be easily learned, although it is possible to obtain approximately correct solutions.  
\end{itemize}

\begin{figure*} 
    \center
    \subfigure[Noise $p=0.00$]{
        \includegraphics[width=0.4 \textwidth]{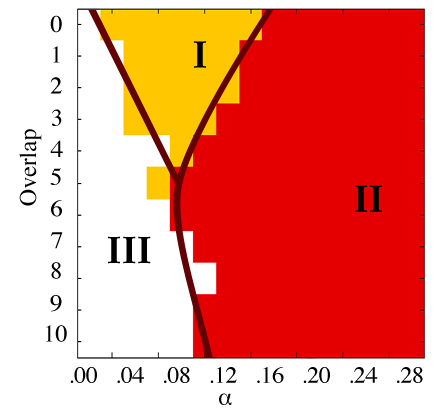}
    }
    \subfigure[Noise $p=0.02$]{
        \includegraphics[width=0.4 \textwidth]{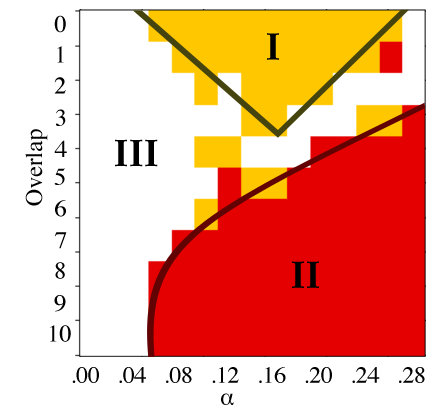}
    }
    \caption{Difficulty map for bi-clustering (a) without noise (b) with noise. Region I: the true model is easily learnable. Region II: the true model cannot be learned. Region III: approximations to the true model may be learned with some difficulty.} \label{fig:maps2} 
\end{figure*}

Thus we transfer the table in Figure~\ref{fig:maps1} into a ``difficulty map''.
Figure \ref{fig:maps2}(a) shows the difficulty map with three regimes with a noise level $p=0.00$; Figure \ref{fig:maps2}(b) shows the difficulty map with $p=0.02$. 
Such difficulty maps visualize the effects of various conditions and parameters and thus can be useful in choosing and configuring the biclustering algorithms.

\section{Conclusion and Discussion}

We present a method for computing the energy landscape maps (ELMs) in model spaces for cluster and bi-cluster learning, and thus visualize for the first time the non-convex energy minimization problems in statistical learning. By plotting the ELMs, we have shown how different problem settings, such as separability, levels of supervision, levels of noise, and  strength of prior  impact the complexity of the energy landscape. We have also compared the behaviors of different learning algorithms in the ELMs. 

Our study leads to the following problems which are worth exploring in future work.
 \begin{enumerate}
 \item If we repeatedly smooth  the energy function, adjacent branches in the ELM will gradually be merged, and this produces a series of ELMs
 representing the coarser structures of the landscape. These ELMs construct the scale space of the landscape.  From this scale space ELM, we shall be able to study the difficulty of the underlying learning problem.

 \item One way to control the scale space of ELM is to design a learning strategy. It starts with lower-dimensional space, simple examples, and proper amount of supervision, and thus the ELM is almost convex.  Once the learning algorithm reaches the global minimum of this ELM, we increase the number of hard examples and dimensions and the ELM becomes increasingly complex. Hopefully, the minimum reached in the previous ELM will be close to the global minimum of ELM at the next stage.  We have studied a case of such \emph{curriculum learning} on dependency grammars in \cite{Maria_curriculum}.

 \item The clustering models are defined on Or-And graph structure (here, 'or' means mixture and 'and' means conjunction of dimensions, and the bi-clustering models are defined on And-Or graph.  In general, it was shown that many advanced learning problems are defined on hierarchical and compositional graphs, which is summarized as multi-layers of And-Or graphs in~ \cite{Zhu06as}. Studying the ELMs for such models will be more challenging but crucial for many  learning tasks of practical importance.
 
 \end{enumerate}

\bibliography{bib}
\bibliographystyle{imsart-nameyear}

%
%
%
%
%
%

\end{document}